\documentclass[11pt]{article}

\usepackage[preprint]{acl}

\usepackage{times}
\usepackage{latexsym}

\usepackage[T1]{fontenc}

\usepackage[utf8]{inputenc}

\usepackage{microtype}

\usepackage{inconsolata}

\usepackage{graphicx}

\usepackage{hyperref}
\usepackage{url}

\usepackage[ruled,vlined,linesnumbered]{algorithm2e}

\usepackage{tcolorbox}
\usepackage{xcolor}

\usepackage{booktabs}
\usepackage[table]{xcolor}
\usepackage{xspace}

\usepackage{multirow}
\usepackage{subcaption}

\usepackage{float}
\usepackage{wrapfig}

\usepackage[table]{xcolor}
\definecolor{SpecRowBlue}{HTML}{E6F4FF}
\newcommand{\specrow}{\rowcolor{SpecRowBlue}}
\definecolor{ForestGreen}{rgb}{0.13,0.55,0.13}
\definecolor{BrickRed}{RGB}{203,65,84}

\definecolor{SpecRowGray}{HTML}{F2F3F5}

\newcommand{\ourmethod}{\textsc{SPEC-RL}}

\usepackage{amsmath}
\usepackage{amsfonts}
\usepackage{arydshln}

\usepackage{nicefrac}
\usepackage{enumitem}

\title{SPEC-RL: Accelerating On-Policy Reinforcement Learning with\\Speculative Rollouts}

\author{
\textbf{Bingshuai Liu\textsuperscript{1}}\thanks{Equal contribution.},
\textbf{Ante Wang\textsuperscript{1,3}}\footnotemark[1],
\textbf{Zijun Min\textsuperscript{1}}\footnotemark[1],
\textbf{Liang Yao\textsuperscript{2}},
\\
\textbf{Haibo Zhang\textsuperscript{2}},
\textbf{Yang Liu\textsuperscript{3}},
\textbf{Xu Han\textsuperscript{3}},
\textbf{Peng Li\textsuperscript{3}},
\textbf{Anxiang Zeng\textsuperscript{2}},
\textbf{Jinsong Su\textsuperscript{1}}\thanks{Corresponding author.}
\\
\textsuperscript{1}School of Informatics, Xiamen University
\\
\textsuperscript{2}LLM Team, Shopee Pte. Ltd.
\quad
\textsuperscript{3}Tsinghua University
\\
\texttt{\{bsliu,wangante,minzijun\}@stu.xmu.edu.cn},
\texttt{\{leon.yao,peter.wu\}@shopee.com},
\\
\texttt{liuyang2011@tsinghua.edu.cn},
\texttt{lipeng@air.tsinghua.edu.cn},
\\
\texttt{thu.hanxu13@gmail.com},
\texttt{zeng0118@ntu.edu.sg},
\texttt{jssu@xmu.edu.cn}
}

\begin{document}
\maketitle
\begin{abstract}
Large Language Models (LLMs) increasingly rely on reinforcement learning with verifiable rewards (RLVR) to elicit reliable chain-of-thought reasoning.
However, the training process remains bottlenecked by the computationally expensive rollout stage. 
Existing acceleration methods—such as parallelization, objective- and data-driven modifications, and replay buffers—either incur diminishing returns, introduce bias, or overlook redundancy across iterations.
We identify that rollouts from consecutive training epochs frequently share a large portion of overlapping segments, wasting computation. 
To address this, we propose \textbf{SPEC-RL}, a novel framework that integrates \textbf{SPEC}ulative decoding with the \textbf{RL} rollout process.
SPEC-RL reuses prior trajectory segments as speculative prefixes and extends them via a draft-and-verify mechanism, avoiding redundant generation while ensuring policy consistency. 
Experiments on diverse math reasoning and generalization benchmarks, including AIME24, MATH-500, OlympiadBench, MMLU-STEM, and others, demonstrate that SPEC-RL reduces rollout time by 2–3$\times$ without compromising policy quality. 
As a purely rollout-stage enhancement, SPEC-RL integrates seamlessly with mainstream algorithms (e.g., PPO, GRPO, DAPO), offering a general and practical path to scale RLVR for large reasoning models. 
Our code is available at: 
\url{https://github.com/ShopeeLLM/Spec-RL}.

\end{abstract}

\section{Introduction}

\begin{figure}
\hspace{-0.5em}
    \includegraphics[width=0.5\textwidth]{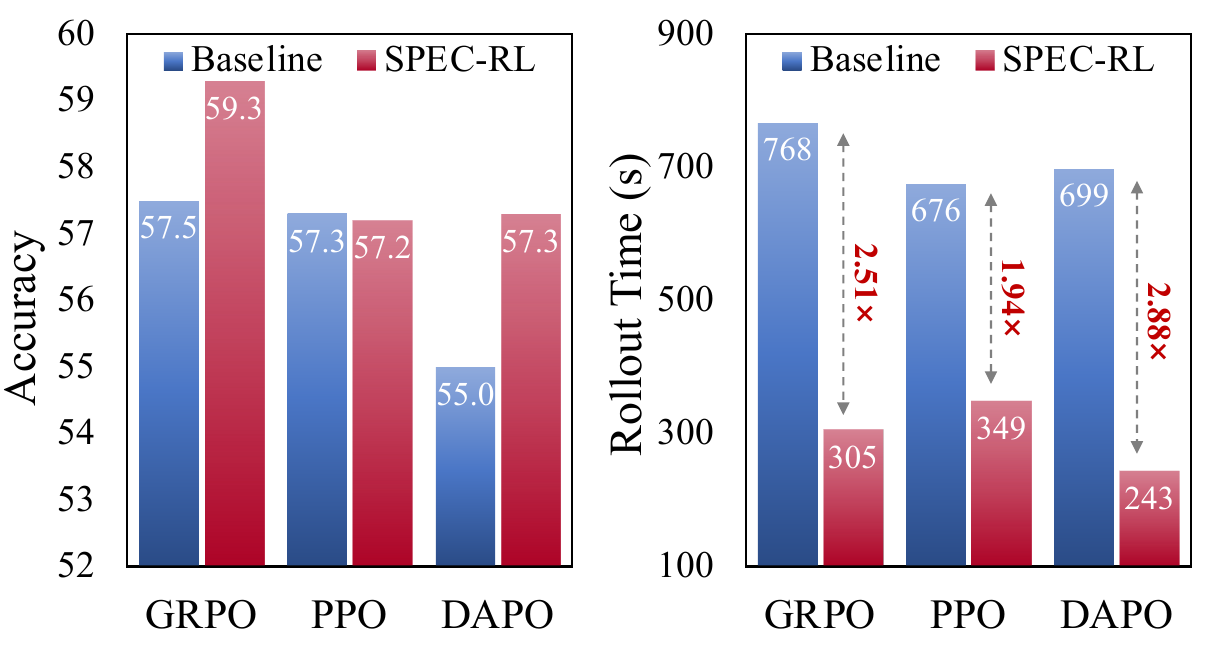}
    \caption{SPEC-RL achieves a 2–3$\times$ speedup in per-step rollout time without compromising average performance on Qwen3-8B-Base across various algorithms.}
    \label{fig:overview}
\end{figure}

Large Language Models (LLMs) have recently achieved substantial progress on challenging reasoning-intensive tasks, such as mathematical problem-solving~\citep{lewkowycz2022solving}, program synthesis~\citep{chen2021evaluating,li2022competition}, and multi-step agentic planning~\citep{yao2023react,yao2023tree}.
A key enabler of these advances is reinforcement learning with verifiable rewards (RLVR)~\citep{lambert2024tulu3,guo2025deepseekr1,yue2025does}, which has emerged as a widely adopted paradigm for incentivizing models to produce faithful and reliable chain-of-thought (CoT) reasoning~\citep{wei2022chain-cot}. 
However, RLVR training pipelines remain constrained by the rollout stage, a fundamental efficiency bottleneck, despite its demonstrated efficacy~\citep{zheng2025greso}.
During this stage, the model must generate large quantities of trajectories through interaction with the environment, a process that is computationally expensive and scales poorly with model size. As a result, the cost and latency of trajectory generation dominate overall training time, severely limiting the practicality of scaling RLVR to increasingly capable LLMs.

To mitigate rollout inefficiency, prior work has explored three directions.
First, parallelized rollout generation increases throughput by producing many trajectories per iteration~\citep{xu2025not}, but its benefits fade as computational and synchronization costs rise.
Second, model-based accelerations reduce environment interaction through modified objectives~\citep{brantley2025astarpo,lin2025cppo}, data restructuring~\citep{liu2025prefixgrouper,zhang2025sortedrl}, or sample selection heuristics~\citep{yu2025dapo,zheng2025greso}, though these approaches often introduce bias and added complexity.
Third, caching methods such as replay buffers reuse prior trajectories~\citep{zhang2025rlep}, thereby improving data utilization, but still require fresh on-policy rollouts and struggle when policies shift significantly.

\begin{figure}
  \centering
  \includegraphics[width=0.42\textwidth]{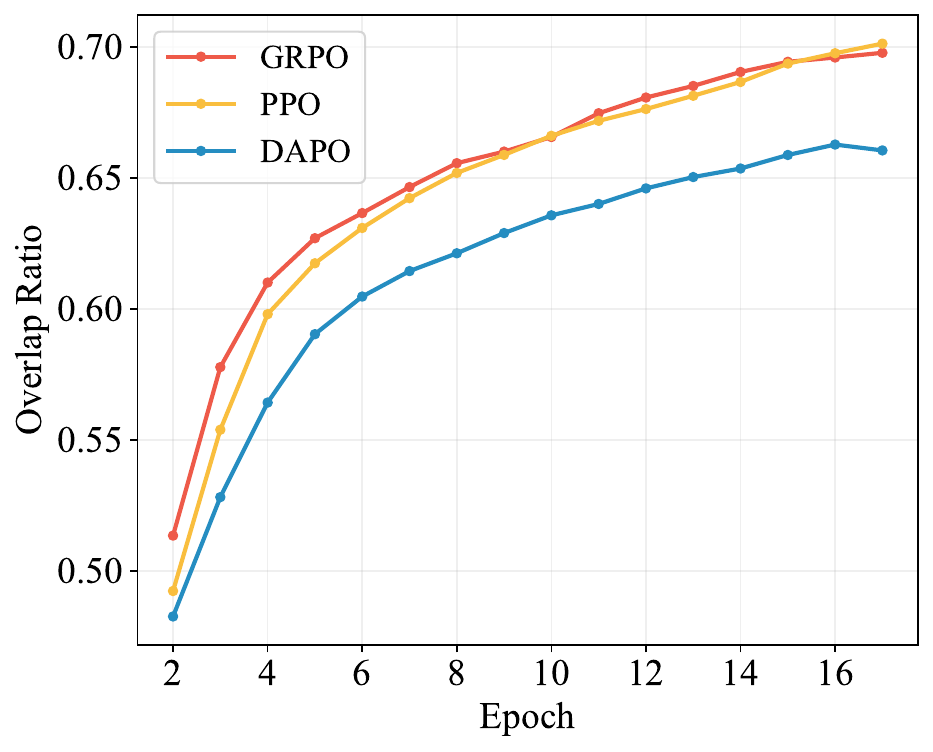}
  \caption{Token overlap ratios across different RL algorithms, computed with ROUGE-1~\citep{lin2004rouge} by comparing rollout responses from consecutive training epochs.}
  
  \label{fig:motivation}
\end{figure}

Existing research~\cite{yang2025qwen3,wang2025rlvr} has found that the quality of the RL training set is more critical than increasing its quantity. Consequently, these studies often adopt a rigorous approach to curating a small but high-quality dataset. This practice typically leads to training for tens or even hundreds of epochs.
The extensive exploration over this same set of instances inevitably results in high redundancy between rollouts from consecutive training epochs, as illustrated in Figure~\ref{fig:motivation}.
Such redundancy naturally arises due to incremental policy updates, with the current policy often behaving similarly to the previous one.
Moreover, in environments with fixed initial states or tasks (e.g., repeated prompts in an LLM reasoning task), the early parts of trajectories tend to overlap across iterations. 
As a result, significant computation is wasted regenerating these overlapping segments. This motivates the central question of our work: \textbf{\emph{can such redundancy be systematically exploited to accelerate rollouts?}}

\begin{figure*}
    \centering
    \includegraphics[width=1.0\textwidth]{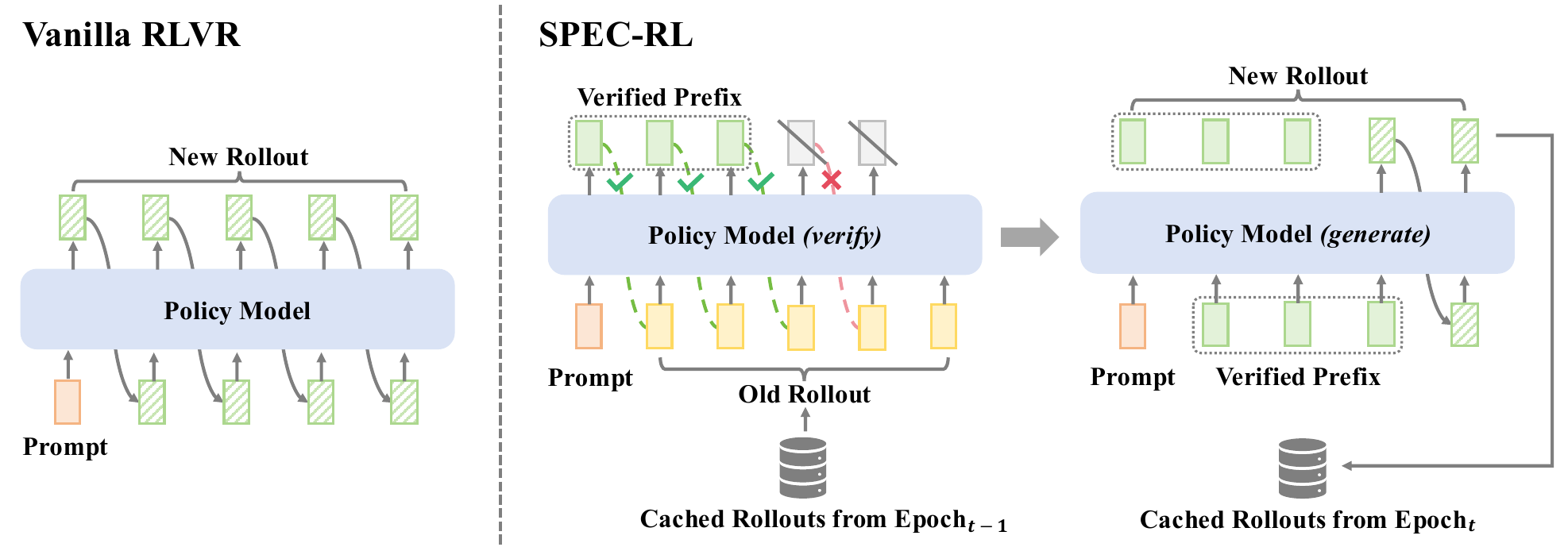}
    \caption{A comparison of the rollout processes in vanilla RLVR and our method. At each training step, vanilla RLVR regenerates full responses. In contrast, \ourmethod, retrieves cached rollouts from the previous epoch. It then verifies these rollouts in parallel, keeps the verified prefixes, and resumes generation with the current policy to produce the final response.
    }
    \label{fig:method}
\end{figure*}

We answer this question by proposing \textbf{SPEC-RL}, a novel framework that integrates \textbf{SPEC}ulative decoding with the \textbf{RL} rollout process. 
Rather than regenerating full trajectories from scratch, \ourmethod\ treats old rollouts from the previous epoch as implicit drafts: following the speculative decoding paradigm, old rollout tokens are verified under the current policy to form a verified prefix.
When the first rejection position is reached, the current policy continues generation from that point onward, as illustrated in Figure~\ref{fig:method}.
This approach is directly analogous to draft-and-verify methods in text generation, where a draft sequence is proposed and then validated in parallel by the target model~\citep{leviathan2023speculativedecoding}. 
By incorporating the same mechanism into RL rollouts, SPEC-RL leverages cached rollouts to skip redundant computation while ensuring that the final outputs remain faithful to the current policy.
The verified prefix is quickly extended by the latest policy, ensuring that the final trajectory remains consistent with the current policy’s behavior.

Our experiments demonstrate that SPEC-RL substantially improves training efficiency across diverse tasks and model scales. 
Concretely, \ourmethod\ consistently reduces rollout generation time by 2--3$\times$ on average, while maintaining or even improving final policy performance across a wide range of math reasoning benchmarks (AIME24~\citep{aime-24-AIMEProblemsSolutions}, MATH-500~\citep{hendrycks2021-math-500-measuring}, Minerva Math~\citep{lewkowycz2022-minervamath-solving}, OlympiadBench~\citep{he2024olympiadbench}, AMC 2023~\citep{amc-23-AMCProblemsSolutions}) and out-of-distribution benchmarks (MMLU-STEM~\citep{hendrycks2020-mmlu-stem-measuring}, IFEval~\citep{zhou2023-ifeval-instruction}).
Importantly, SPEC-RL is designed as a modular enhancement to the data collection phase, making it readily applicable to a wide range of mainstream RLVR algorithms, such as GRPO, DAPO, and PPO.

In summary, we identify substantial rollout redundancy in RLVR and show that reusing overlapping trajectory segments can greatly reduce sampling cost. Building on this insight, we introduce \ourmethod, the first framework to incorporate speculative decoding into RL rollouts by treating previous-epoch trajectories as implicit drafts and verifying reusable prefixes. This design integrates smoothly with mainstream RL algorithms and yields significant rollout acceleration while maintaining policy performance.

\section{Preliminaries}
\label{sec:prelim}

\subsection{On-Policy Reinforcement Learning}
Reinforcement learning with verifiable rewards (RLVR) is a training paradigm that couples RL with objective, externally verifiable signals. It typically employs on-policy methods like Proximal Policy Optimization (PPO), where the agent learns exclusively from rollouts generated by its current, cautious policy.
Formally, for an instance $(\mathbf{x}, \mathbf{y}^\ast)$ from the training set $\mathcal{D}$—where $\mathbf{x}$ is a prompt and $\mathbf{y}^\ast$ is the ground-truth answer—a rollout is sampled from the policy as $\mathbf{y} \sim \pi(\cdot \mid \mathbf{x})$. A reward function $R(\mathbf{y}, \mathbf{y}^\ast)$ then evaluates whether $\mathbf{y}$ matches $\mathbf{y}^\ast$. This feedback is typically binary and tamper-resistant to ensure precision, as in the case of numerical comparisons or unit test results.
The on-policy RL objective is to maximize the expected reward of these generated rollouts:
\begin{equation}
J(\theta) = \mathbb{E}_{(\mathbf{x}, \mathbf{y}^\ast)\sim \mathcal{D}, \mathbf{y} \sim \pi(\cdot \mid \mathbf{x})} \big[ R(\mathbf{y}, \mathbf{y}^\ast) \big].
\end{equation}

This approach ensures the training data remain consistent with the current policy distribution, carefully balancing exploration with stable, incremental updates to guarantee reliable performance. In this work, we retain the standard RL objective and policy update methods, focusing instead on improving the efficiency of the rollout stage.

\subsection{Speculative Decoding}
Speculative decoding follows a draft-and-verify paradigm: an efficient draft model \( p \) proposes multiple future tokens, which are then verified in parallel by the target model \( q \). A drafted token \( z_i \sim p(\cdot \mid \mathbf{x}, \mathbf{z}_{<i}) \) is accepted with probability
\begin{equation}
\label{eq:spec-accept}
\alpha_i
 \;=\; \min\!\Bigl(1,\; \frac{q(z_i \mid \mathbf{x}, \mathbf{z}_{<i})}{p(z_i \mid \mathbf{x}, \mathbf{z}_{<i})} \Bigr).
\end{equation}

This ensures the process samples exactly from \( q \), preserving fidelity while accelerating generation by reducing costly target computations. The speedup is governed by the acceptance rate and the cost gap between the draft and target models.

\section{Method}
\label{sec:method}

The goal of \ourmethod\ is to accelerate RL rollouts by avoiding redundant regeneration. 
Instead of sampling complete trajectories from scratch at every step, we leverage cached rollouts from the previous epoch and reuse as much of them as possible, only generating the minimal continuation that is inconsistent with the current policy (Figure~\ref{fig:method}). 
This reduces the number of decoded tokens and directly cuts rollout latency. 

\begin{algorithm}[t]
\caption{Speculative Rollout}
\label{alg:training-algo}
\KwIn{
    Current policy $\pi_{\theta}$;
    Prompt $\mathbf{x}$;
    Previous rollout $\mathbf{y}^{\text{prev}}$ with token probabilities $\mathbf{p}^{\text{prev}}$;
    Lenience parameter $\ell \geq 1$.
}
Compute current token probabilities:
$\mathbf{p}^{\text{curr}} \gets \big[ \pi_{\theta}(y^{\text{prev}}_i \mid \mathbf{x}, \mathbf{y}^{\text{prev}}_{<i}) \big]_{i=1}^{|\mathbf{y}^{\text{prev}}|}$\;

Compute per-token acceptance probability:
$\widetilde{\boldsymbol{\alpha}} \gets \min\Big( \mathbf{1},\; \ell \cdot \dfrac{\mathbf{p}^{\text{curr}}}{\mathbf{p}^{\text{prev}}} \Big)$\;

Initialize rejection position $n \gets |\mathbf{y}^{\text{prev}}| + 1$\;
\For{$i = 1$ \KwTo $|\mathbf{y}^{\text{prev}}|$}{
    Sample $u \sim \mathcal{U}(0,1)$\;
    \If{$u > \widetilde{\alpha}_i$}{
        $n \gets i$\;
        \textbf{break}\;
    }
}
Generate continuation from position $n$:
$\mathbf{y}^{\text{curr}}_{\geq n} \sim \pi_{\theta}(\cdot \mid \mathbf{x}, \mathbf{y}^{\text{prev}}_{<n})$\;

Assemble final response:
$\mathbf{y}^{\text{curr}} \gets \mathbf{y}^{\text{prev}}_{<n} \oplus \mathbf{y}^{\text{curr}}_{\geq n}$\;

\Return $\mathbf{y}^{\text{curr}}$
\end{algorithm}

\subsection{Speculative Decoding over Cached Rollouts with Lenience}
The core of \ourmethod\ is adapting speculative decoding to RL by treating previous rollouts as draft sequences.
Formally, when sampling a rollout for a prompt \(\mathbf{x}\) during epoch \(t\), let \(\mathbf{y}^{\text{prev}}\) denote a rollout generated for the same prompt in the previous epoch \(t-1\) under the policy \(\pi_{\text{prev}}\).
Following the standard draft--and--verify formulation in Equation~\ref{eq:spec-accept}, we verify each token \(y^{\text{prev}}_i\) under the current policy \(\pi_{\text{curr}}\) and decide whether it can be reused. The acceptance probability for token \(i\) is
\(
\min\!\Bigl(1, \nicefrac{\pi_{\text{curr}}(y^{\text{prev}}_{i} \mid \mathbf{x}, \mathbf{y}^{\text{prev}}_{<i})}
                     {\pi_{\text{prev}}(y^{\text{prev}}_{i} \mid \mathbf{x}, \mathbf{y}^{\text{prev}}_{<i})}\Bigr).
\)
This ensures exact consistency with the current policy; however, it can be overly strict in practice, limiting the amount of reusable tokens.

To address this, we introduce a lenience parameter $\ell$, following prior work~\citep{chen2024cascade}. This relaxes the acceptance condition by shifting the decision boundary. Formally, the acceptance probability becomes
\begin{equation}
\label{eq:lenience-speculative-decoding}
\tilde{\alpha}_{i} = \min\!\Bigl(1,\;\ell \cdot 
        \frac{\pi_{\text{curr}}(y^{prev}_{i}\mid \mathbf{x}, \mathbf{y}^{prev}_{<i})}
             {\pi_{\text{prev}}(y^{prev}_{i}\mid \mathbf{x}, \mathbf{y}^{prev}_{<i})}
        \Bigr).
\end{equation}

A token is accepted if a random sample \(u \sim \mathcal{U}(0,1)\) satisfies \(u \le \tilde{\alpha}_i\), and rejected otherwise. The verified prefix \(\mathbf{y}^{\text{prev}}_{<n}\) is determined by identifying the first rejected token at position \(n\).
The final rollout, \(\mathbf{y}^{\text{curr}}=\mathbf{y}^{\text{prev}}_{<n} \oplus \mathbf{y}^{\text{curr}}_{\geq n}\), is formed by concatenating this verified prefix with a newly generated continuation \(\mathbf{y}^{\text{curr}}_{\geq n} \sim \pi_{\text{curr}}(\cdot \mid \mathbf{x}, \mathbf{y}^{\text{prev}}_{<n})\).
The overall process of verification, generation, and assembly is summarized in Algorithm~\ref{alg:training-algo}.

\paragraph{Difference from speculative decoding}
\ourmethod\ follows the draft--and--verify paradigm of speculative decoding, but in a simplified, single-round form. 
Vanilla speculative decoding typically requires a separate draft model, loading extra parameters, scheduling overhead, and multiple verification rounds. 
In contrast, \ourmethod\ reuses the previous policy as the draft, with cached rollouts available ``for free''. 
This eliminates the need for auxiliary models while preserving the fidelity guarantees of speculative decoding.

\paragraph{The mechanism of lenience}
The parameter $\ell$ provides a flexible mechanism to balance rollout efficiency against on-policy fidelity: setting \(\ell = 1\) recovers the original speculative decoding rule; \(\ell > 1\) increases the acceptance rate, leading to longer reused prefixes; \(\ell \to \infty\) corresponds to full prefix reuse; and \(\ell \to 0\) reverts to standard RLVR without reuse.
Although this introduces a parameter that requires manual tuning, we demonstrate empirically that a moderate value of \(\ell\) yields significant speedup without degrading performance.

\subsection{Implementing \ourmethod\ in RL Training}
To enable practical use in RL training pipelines, \ourmethod\ introduces a lightweight cache module that stores rollouts from the previous epoch and continuously refreshes them during training.
When the same prompt reappears, its cached response is retrieved to serve as a draft sequence for reuse.
To ensure efficacy, all draft verification requests within a training batch are packed into a single call to the rollout engine.
Verified prefixes and prompts are aligned via left padding, enabling parallel processing of further continuations without fragmentation.
Once new rollouts are generated, the cache is updated for the corresponding prompts for later use.
This ensures retrieved rollouts are always generated by the most recent policy, naturally increasing token reuse probability.

This design allows \ourmethod\ to operate as a drop-in module: it modifies only the rollout stage, requires no changes to reward computation or policy updates, and is compatible with mainstream algorithms such as GRPO and PPO.

\section{Experiments}
\label{sec:experiments}


\begin{table*}[!t]
\centering
\resizebox{\textwidth}{!}{%
\begin{tabular}{l r r r r r r r r r r}
\toprule
\multirow{3}{*}{\textbf{Algorithm}} &
\multicolumn{2}{c}{\textbf{Rollout Efficiency}} &
\multicolumn{5}{c}{\textbf{Math Reasoning}} &
\multicolumn{2}{c}{\textbf{OOD}} &
\multirow{3}{*}{\textbf{AVG}} \\
\cmidrule(lr){2-3}\cmidrule(lr){4-8}\cmidrule(lr){9-10}
& \textbf{Tokens (M)} & \textbf{Speedup} &
\textbf{AMC23} & \textbf{AIME24} &
\textbf{\begin{tabular}[c]{@{}c@{}}MATH\\ 500\end{tabular}} &
\textbf{\begin{tabular}[c]{@{}c@{}}Minerva\\ Math\end{tabular}} &
\textbf{\begin{tabular}[c]{@{}c@{}}Olympiad\\ Bench\end{tabular}} &
\textbf{\begin{tabular}[c]{@{}c@{}}MMLU\\ STEM\end{tabular}} &
\textbf{IFEval} & \\
\midrule

\rowcolor{SpecRowGray}
\multicolumn{11}{c}{\textbf{\textit{Qwen3-1.7B-Base}}} \\
Base Model & - & - & 21.9 & 2.5 & 45.0 & 12.5 & 16.7 & 39.3 & 17.9 & 22.3 \\
\hdashline
GRPO & 554.8 & 1.00$\times$ & 35.0 & 7.8 & 64.4 & 26.5 & 25.5 & \bf 60.7 & 24.4 & 34.9 \\
\specrow
\hspace{2mm}$\hookrightarrow$~+~\ourmethod
& \bf 182.7 & \bf 2.29$\times$ & \bf 38.3 & \bf 9.1 & \bf 68.0 & \bf 29.4 & \bf 29.3 & 58.3 & \bf 28.8 & \bf 37.3 \\
PPO & 565.1 & 1.00$\times$ & 35.3 & 8.3 & 63.0 & \bf 26.8 & 25.3 & \bf 59.4 & 25.5 & 34.8 \\
\specrow
\hspace{2mm}$\hookrightarrow$~+~\ourmethod
& \bf 230.8 & \bf 1.94$\times$ & \bf 36.9 & 7.7 & \bf 64.8 & 25.4 & \bf 25.9 & 58.6 & \bf 25.9 & \bf 35.0 \\
DAPO & 543.1 & 1.00$\times$ & 33.1 & 4.9 & \bf 60.8 & 24.6 & 23.0 & 52.2 & 24.8 & 31.9 \\
\specrow
\hspace{2mm}$\hookrightarrow$~+~\ourmethod
& \bf 171.6 & \bf 2.17$\times$ & 33.1 & 4.9 & 60.0 & \bf 25.7 & \bf 25.5 & \bf 53.5 & \bf 27.0 & \bf 32.8 \\

\midrule
\rowcolor{SpecRowGray}
\multicolumn{11}{c}{\textbf{\textit{Qwen3-8B-Base}}} \\
Base Model & - & - & 40.2 & 11.5 & 67.4 & 27.2 & 34.1 & 60.4 & 29.9 & 38.7 \\
\hdashline
GRPO & 1033.1 & 1.00$\times$ & 69.2 & 24.2 & 86.4 & 43.8 & \bf 53.0 & \bf 84.6 & 41.2 & 57.5 \\
\specrow
\hspace{2mm}$\hookrightarrow$~+~\ourmethod
& \bf 336.6 & \bf 2.51$\times$ & \bf 72.8 & \bf 26.9 & \bf 87.8 & \bf 44.1 & 51.0 & 84.5 & \bf 47.7 & \bf 59.3 \\
PPO & 984.0 & 1.00$\times$ & 68.9 & \bf 26.4 & \bf 85.8 & 43.0 & \bf 51.6 & 83.8 & \bf 41.6 & \bf 57.3 \\
\specrow
\hspace{2mm}$\hookrightarrow$~+~\ourmethod
& \bf 400.1 & \bf 1.94$\times$ & \bf 70.5 & 25.1 & 85.2 & \bf 43.4 & 50.8 & \bf 84.4 & 41.0 & 57.2 \\
DAPO & 1052.2 & 1.00$\times$ & 65.2 & 24.0 & \bf 84.8 & 40.1 & 48.6 & \bf 82.4 & 39.6 & 55.0 \\
\specrow
\hspace{2mm}$\hookrightarrow$~+~\ourmethod
& \bf 326.2 & \bf 2.88$\times$ & \bf 69.7 & \bf 26.4 & 84.4 & \bf 43.8 & \bf 50.4 & 82.2 & \bf 44.4 & \bf 57.3 \\

\midrule
\rowcolor{SpecRowGray}
\multicolumn{11}{c}{\textbf{\textit{LLaMA-3.2-1B-Instruct}}} \\
Base Model & - & - & 7.5 & 0.6 & 14.2 & 4.0 & 2.8 & 32.6 & 37.0 & 14.1 \\
\hdashline
GRPO & 553.9 & 1.00$\times$ & 7.5 & 1.4 & 19.2 & \bf 3.3 & 4.9 & 33.1 & 37.0 & 15.2 \\
\specrow
\hspace{2mm}$\hookrightarrow$~+~\ourmethod
& \bf 162.5 & \bf 2.60$\times$ & \bf 8.8 & \bf 1.7 & \bf 19.4 & 1.8 & \bf 5.0 & \bf 34.5 & \bf 37.2 & \bf 15.5 \\
PPO & 521.5 & 1.00$\times$ & 7.8 & 0.8 & \bf 20.8 & 4.0 & \bf 6.4 & 34.3 & \bf 42.7 & 16.7 \\
\specrow
\hspace{2mm}$\hookrightarrow$~+~\ourmethod
& \bf 210.6 & \bf 2.01$\times$ & \bf 9.2 & \bf 1.4 & 20.2 & \bf 5.5 & 5.0 & \bf 35.3 & 40.7 & \bf 16.8 \\
DAPO & 482.6 & 1.00$\times$ & 9.2 & 0.9 & 19.2 & 4.0 & 5.5 & 33.0 & \bf 38.6 & 15.8 \\
\specrow
\hspace{2mm}$\hookrightarrow$~+~\ourmethod
& \bf 123.1 & \bf 2.48$\times$ & \bf 11.6 & \bf 2.0 & \bf 20.2 & 4.0 & 5.5 & \bf 35.5 & 38.4 & \bf 16.7 \\
\bottomrule
\end{tabular}
}
\caption{Main results across various models and RL algorithms on DeepMath-6K.}
\label{tab:math_reasoning}
\end{table*}

\subsection{Experimental Setup}
\paragraph{Models and hyperparameters}
We evaluate \ourmethod\ on Qwen3-1.7B-Base, Qwen3-8B-Base, and LLaMA-3.2-1B-Instruct across various RL algorithms, including GRPO, PPO, and DAPO.
All models are trained using the veRL~\citep{sheng2025verl-hybridflow} framework with vLLM~\citep{kwon2023-vllm-efficient} as the rollout engine, on data sampled from DeepMath (6,144 examples, denoted as DeepMath-6K)~\citep{he2025deepmath} and SimpleRL (8,192 examples, denoted as SimpleRL-8K)~\citep{zeng2025simplerl}. 
All experiments use a rollout batch size of 1,024 and a maximum response length of 4,096 tokens, conducted on a single node with $8\times$ NVIDIA H100 GPUs. 
Rollout is performed at a temperature of $ 1.0$.
The actor learning rate is fixed at $5\times10^{-7}$, and for PPO we set the critic learning rate to $1\times10^{-5}$. 
Further hyperparameters are detailed in Appendix~\ref{appendix:hyper-params}.
\paragraph{Benchmarks and metrics}
We evaluate rollout efficiency and accuracy on a broad suite of benchmarks. 
\textbf{Rollout efficiency} is reported as the number of generated tokens and the relative speedup (baseline time divided by method time). 
\textbf{Math reasoning benchmarks} include AMC 2023~\citep{amc-23-AMCProblemsSolutions}, AIME24~\citep{aime-24-AIMEProblemsSolutions}, MATH-500~\citep{hendrycks2021-math-500-measuring}, Minerva Math~\citep{lewkowycz2022-minervamath-solving}, and OlympiadBench~\citep{he2024olympiadbench}. 
\textbf{Out-of-distribution (OOD) benchmarks} include MMLU-STEM~\citep{hendrycks2020-mmlu-stem-measuring} and IFEval~\citep{zhou2023-ifeval-instruction}, which evaluate the generalization capability of the model. 
Full evaluation details are provided in Appendix~\ref{appendix:eval-setup}.

\begin{table*}[!t]
\centering
\resizebox{\textwidth}{!}{%
\begin{tabular}{l r r r r r r r r r r}
\toprule
\multirow{3}{*}{\textbf{Algorithm}} &
\multicolumn{2}{c}{\textbf{Rollout Efficiency}} &
\multicolumn{5}{c}{\textbf{Math Reasoning}} &
\multicolumn{2}{c}{\textbf{OOD}} &
\multirow{3}{*}{\textbf{AVG}} \\
\cmidrule(lr){2-3}\cmidrule(lr){4-8}\cmidrule(lr){9-10}
& \textbf{Tokens (M)} & \textbf{Speedup} &
\textbf{AMC23} & \textbf{AIME24} &
\textbf{\begin{tabular}[c]{@{}c@{}}MATH\\ 500\end{tabular}} &
\textbf{\begin{tabular}[c]{@{}c@{}}Minerva\\ Math\end{tabular}} &
\textbf{\begin{tabular}[c]{@{}c@{}}Olympiad\\ Bench\end{tabular}} &
\textbf{\begin{tabular}[c]{@{}c@{}}MMLU\\ STEM\end{tabular}} &
\textbf{IFEval} & \\
\midrule
\ourmethod
& \bf 182.7 & 2.29$\times$ & \bf 38.3 & \bf 9.1 & \bf 68.0 & \bf 29.4 & \bf 29.3 & \bf 58.3 & \bf 28.8 & \bf 37.3 \\
\hspace{4mm}$\Rightarrow$~Random Reuse & 304.5 & \bf 2.35$\times$ & 30.3 & 5.0 & 60.4 & 21.7 & 25.3 & 53.1 & 24.0 & 31.4 \\
\hspace{4mm}$\Rightarrow$~Delayed Reuse & 308.8 & 1.44$\times$ & 35.6 & 9.0 & 66.4 & 29.0 & 28.6 & 57.3 & 28.3 & 36.3 \\
\bottomrule
\end{tabular}
}
\caption{Comparison between \ourmethod\ and its variants when training Qwen3-1.7B-Base on DeepMath-6K with GRPO. 
}
\label{tab:specdec_vs_random}
\end{table*}



\begin{table*}[!t]
\centering
\resizebox{\textwidth}{!}{%
\begin{tabular}{l r r r r r r r r r r}
\toprule
\multirow{3}{*}{\textbf{Algorithm}} &
\multicolumn{2}{c}{\textbf{Rollout Efficiency}} &
\multicolumn{5}{c}{\textbf{Math Reasoning}} &
\multicolumn{1}{c}{\textbf{OOD}} &
\multirow{3}{*}{\textbf{AVG}} \\
\cmidrule(lr){2-3}\cmidrule(lr){4-8}\cmidrule(lr){9-9}
& \textbf{Tokens (M)} & \textbf{Speedup} &
\textbf{AMC23} &
\textbf{AIME24} &
\textbf{\begin{tabular}[c]{@{}c@{}}MATH\\ 500\end{tabular}} &
\textbf{\begin{tabular}[c]{@{}c@{}}Minerva\\ Math\end{tabular}} &
\textbf{\begin{tabular}[c]{@{}c@{}}Olympiad\\ Bench\end{tabular}} &
\textbf{\begin{tabular}[c]{@{}c@{}}MMLU\\ STEM\end{tabular}} &
\textbf{IFEval} & \\
\midrule
GRPO & 554.8 & 1.00$\times$ & 35.0 & 7.8 & 64.4 & 26.5 & 25.5 & 60.7 & 24.4 & 34.9 \\
~~~~$\hookrightarrow$~+~\ourmethod\ $\ell=1$  
& 419.1 & 1.22$\times$ & 37.8 & 8.0 & 63.8 & 28.7 & 26.5 & 59.6 & 25.9 & 35.8 \\
~~~~$\hookrightarrow$~+~\ourmethod\ $\ell=e^{0.2}$ 
& 246.7 & 1.86$\times$ & 35.8 & 7.8 & 66.4 & \bf29.8 & \bf29.6 & 58.5 & 25.9 & 36.3 \\
~~~~$\hookrightarrow$~+~\ourmethod\ $\ell=e^{0.5}$ 
& 182.7 & 2.29$\times$ & \bf38.3 & \bf9.1 & \bf68.0 & 29.4 & 29.3 & 58.3 & 28.8 & \bf37.3 \\
~~~~$\hookrightarrow$~+~\ourmethod\ $\ell=e^{0.8}$ 
& 144.8 & 2.64$\times$ & 28.6 & 5.0 & 63.6 & 27.2 & 25.0 & \bf61.7 & 26.2 & 33.9 \\
~~~~$\hookrightarrow$~+~\ourmethod\ $\ell=e^{2.0}$ 
& 114.4 & 3.05$\times$ & 26.2 & 5.2 & 55.0 & 21.0 & 21.9 & 53.5 & \bf29.0 & 30.3 \\
~~~~$\hookrightarrow$~+~\ourmethod\ $\ell=\infty$ 
& \bf40.0 & \bf14.86$\times$ & 30.2 & 4.4 & 60.4 & 19.9 & 23.7 & 44.1 & 22.0 & 29.2 \\
\bottomrule
\end{tabular}
}
\caption{Ablation on lenience parameter $\ell$ on DeepMath-6K. Here $\ell=1$ corresponds to vanilla speculative decoding, while $\ell=\infty$ corresponds to full reuse.}
\label{tab:ablation_lenience}
\end{table*}

\subsection{Main Results}
As shown in Table~\ref{tab:math_reasoning}, across nine model--algorithm settings, \ourmethod\ yields an average speedup of 2.31$\times$ by reducing generated tokens by 66\%.
The largest gain is with Qwen-3-8B-Base under DAPO (1{,}052.2M → 326.2M tokens; 2.88$\times$), while even the smallest case (Qwen-3-8B-Base with PPO) achieves 1.94$\times$ without accuracy loss.
These improvements closely track the reduction in generated tokens, confirming that token-level savings drive the acceleration.

On math benchmarks, accuracy remains broadly stable: larger models are highly robust, while smaller models show only small fluctuations.
For OOD tasks, MMLU-STEM stays nearly unchanged, and IFEval improves in several cases—for example, +6.5 points on Qwen-3-8B-Base with GRPO.
Overall, \ourmethod\ accelerates rollout generation substantially without degrading reasoning quality, and sometimes even improves out-of-distribution generalization.

Additional analyses, including wall-clock breakdowns, performance on larger model scale, and the effect of training-set size, comparisons on DeepMath-6K and SimpleRL-8K, per-step training curves, are provided in Appendices \ref{appendix:e2e_breakdown}, \ref{appendix:larger-model-14b}, \ref{appendix:data-size}, \ref{appendix:extra-datasets}, \ref{appendix:details-steps-algo-models}.

\subsection{Ablation Studies and Analyses}
\label{section:lenience-effect}
The draft--and--verify strategy and the lenience parameter $\ell$ play critical roles in \ourmethod. In this section, we conduct ablation experiments on different rollout reuse methods and various values of $\ell$. We consistently use Qwen3-1.7B-Base with GRPO, trained on the DeepMath-6K dataset. Our analysis proceeds as follows.

\paragraph{Comparison with different reuse variants}
To better understand our method, we design two variants:
\begin{itemize}[leftmargin=*]
    \item \textbf{Random Reuse:} Instead of using Equation~\ref{eq:lenience-speculative-decoding} to compute the acceptance criterion, rejection positions are sampled uniformly at random. This results in roughly half of the tokens being reused on expectation.
    \item \textbf{Delayed Reuse:} Instead of retrieving rollouts from the most recent epoch as draft sequences, this variant uses rollouts from two epochs ago to validate the necessity of using the most recent rollouts.
\end{itemize}
As shown in Table~\ref{tab:specdec_vs_random}, Random Reuse enjoys higher efficiency against \ourmethod\ (2.35$\times$ vs.\ 2.29$\times$) without verification cost, but it causes a substantial drop in accuracy (31.4 vs.\ 37.3).
This highlights that naive reuse introduces harmful noise, whereas \ourmethod\ leverages speculative verification to retain policy fidelity while accelerating training.
Delayed Reuse shows competitive performance compared to the GRPO baseline (36.3 vs.\ 37.3); however, its efficiency is significantly reduced (1.44$\times$ vs.\ 2.29$\times$). This occurs because the larger policy gap leads to fewer reused tokens, demonstrating the effectiveness of our immediate cache-updating strategy.
The detailed results are provided in Appendix \ref{appendix:random-reuse}.

\begin{figure*}[!t]
    \centering
    \includegraphics[width=1.0\textwidth]{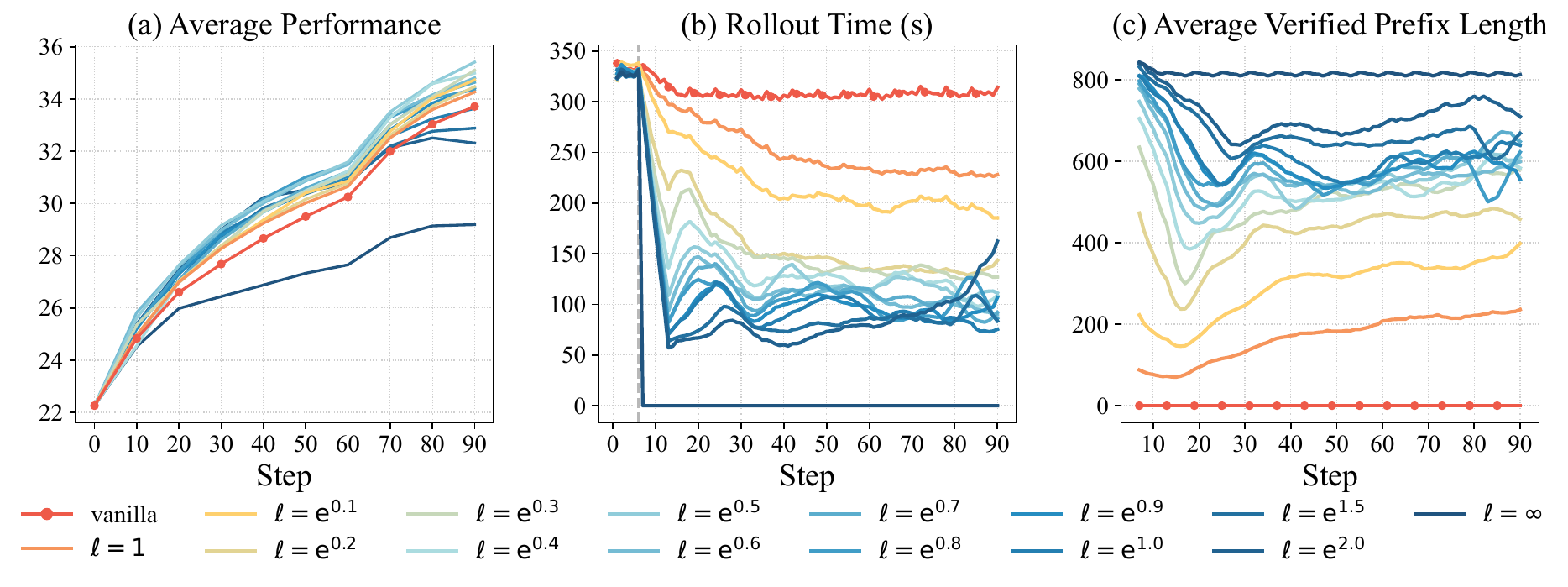}
    \caption{The effect of lenience $\ell$ on model performance and efficiency: (a) averaged test performance, (b) rollout time, and (c) averaged verified prefix length at different training steps.}
    \label{fig:lenience-reward-prefix-skip}
\end{figure*}

\begin{figure*}[!t]
    \centering
    \includegraphics[width=1.0\textwidth]{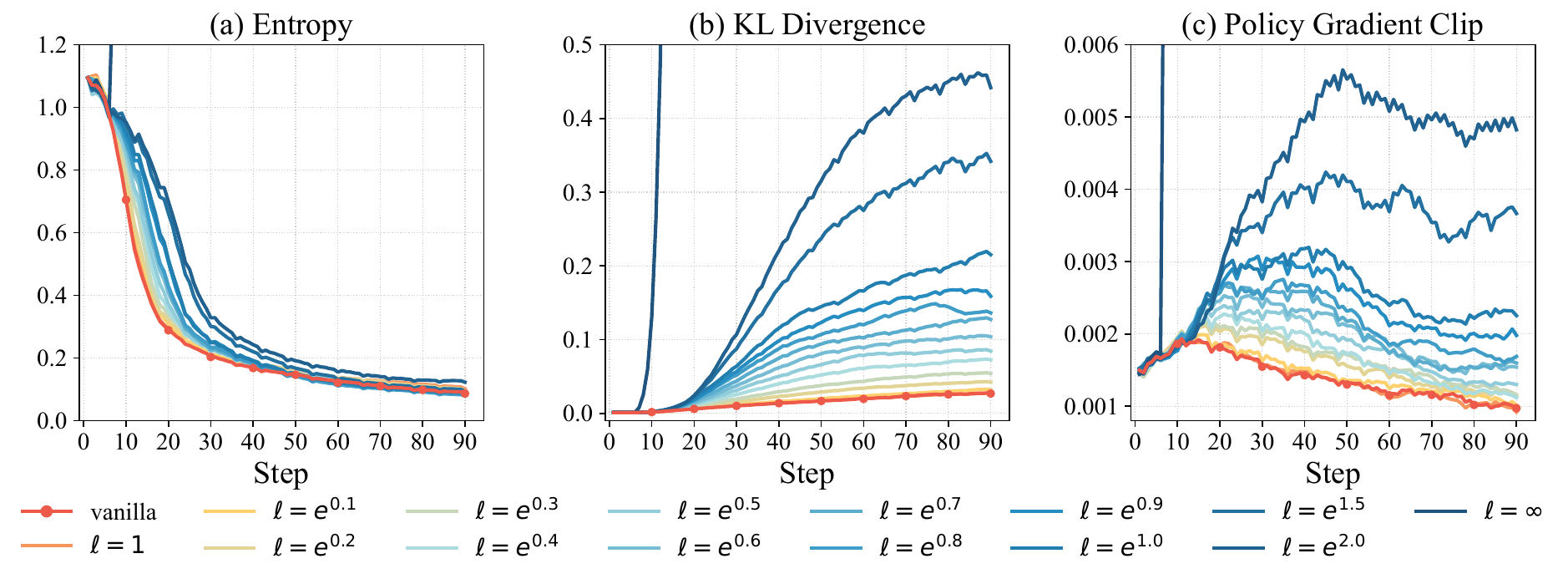}
    \caption{Training dynamics of \ourmethod\ under different $\ell$ for three metrics: (a) entropy, (b) KL divergence, and (c) policy gradient clip ratio.}
    \label{fig:actor}
\end{figure*}

\paragraph{Impact of lenience $\ell$ on efficacy}
As shown in Table~\ref{tab:ablation_lenience} and Figure~\ref{fig:lenience-reward-prefix-skip}, increasing $\ell$ consistently improves rollout efficiency: starting from vanilla speculative decoding at $\ell=1$ with a speedup of only 1.22$\times$, the acceleration rises steadily and reaches 14.86$\times$ when $\ell \to \infty$.
This improvement stems from the greater prefix length reused when $\ell$ increases. Interestingly, as training progresses, the reused prefix length exhibits a trend of first decreasing and then rising. This occurs because policy parameters are updated more aggressively in the early stages of training, widening the policy gap and reducing reuse. As training continues, however, the policy stabilizes, updates become less frequent, and reuse increases accordingly.
Accuracy, however, does not follow the same trend—performance peaks at $\ell=e^{0.5}$ with 37.3, but declines when reuse becomes overly aggressive, dropping to 29.2 at $\ell \to \infty$. 
Overall, moderate lenience values strike the best balance, yielding 2–3$\times$ rollout speedups while preserving or slightly improving accuracy, whereas extreme reuse sacrifices performance despite dramatic acceleration.
The detailed intermediate results throughout training are provided in Appendix~\ref{appendix:lenience-ablation}.

\paragraph{Impact of Lenience $\ell$ on diagnostic metrics}
Due to the complexity of modern RL algorithms, several metrics, such as policy entropy, KL divergence, and the clip fraction of rollouts, are commonly used to diagnose training health. We investigate how the lenience parameter $\ell$ influences these metrics; the corresponding results are plotted in Figure~\ref{fig:actor}.
For $\ell = 1$, the metrics closely match those of the GRPO baseline. This is expected because rollouts are fully on-policy in this setting. As $\ell$ increases, all three metrics rise consistently, reflecting greater deviation of the rollout data from the current policy. In the extreme case where $\ell \to \infty$, rollouts are reused completely, causing values to increase sharply and exceed the plotted range, clearly indicating the training instability.
Nevertheless, with a moderate choice of $\ell$, the metrics remain within a stable region. For example, when $\ell < e^{0.5}$, the clip fraction stays below 0.0025 and KL divergence remains under 0.1. Additional case studies in Appendix~\ref{appendix:case-study} demonstrate how \ourmethod\ reuses validated prefixes while preserving the integrity of the reasoning chain. This illustrates why moderate reuse strikes an optimal balance, thereby allowing \ourmethod\ to maintain strong performance without sacrificing training stability.

\begin{figure}[t]
    \centering
    \includegraphics[width=0.49\textwidth]{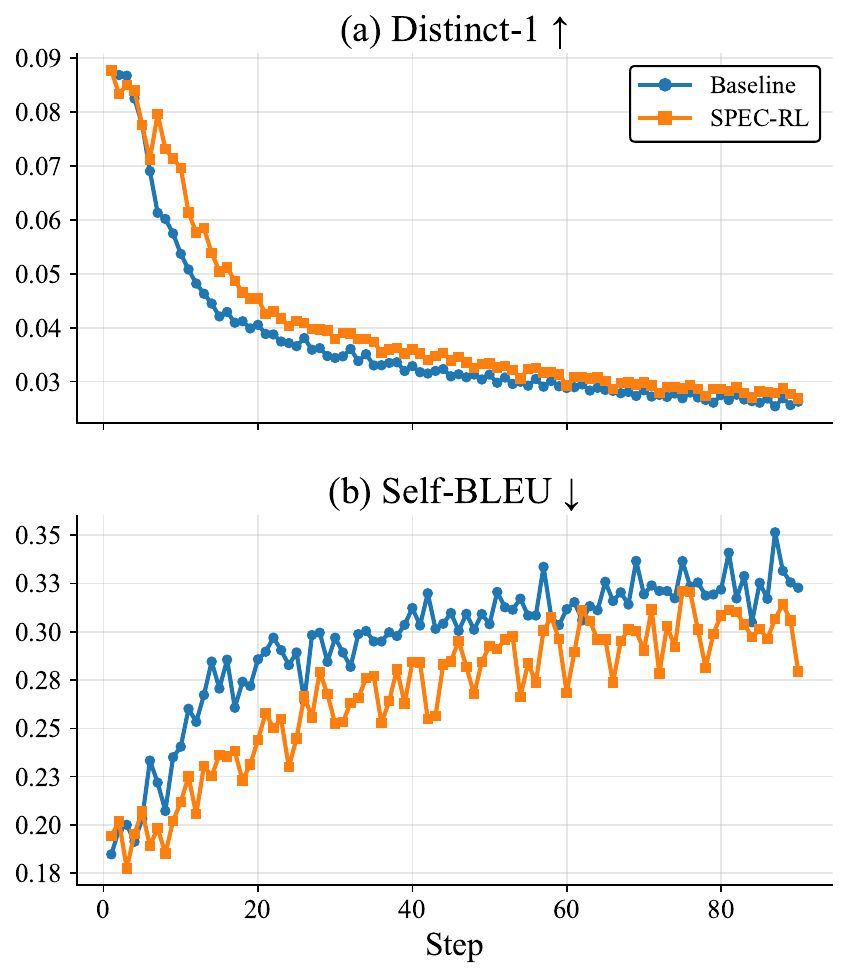}
    \caption{Effects of SPEC-RL on rollout diversity measured by Distinct-1 and Self-BLEU.}
    \label{fig:diversity}
\end{figure}

\paragraph{Analysis of rollout diversity}
We further examine whether speculative reuse affects rollout diversity by comparing SPEC-RL with the GRPO baseline at identical training steps. Diversity is measured using two standard metrics: \textbf{Distinct-1}~\citep{li2016diversity-distinct-1}, which captures unigram variability, and \textbf{Self-BLEU}~\citep{zhu2018texygen-self-bleu}, which measures similarity among samples within a batch. As shown in Figure~\ref{fig:diversity}(a)--(b), SPEC-RL consistently achieves \emph{equal or higher} diversity than the GRPO baseline throughout training.
Although SPEC-RL reuses prefixes generated by a previous policy, this reuse affects the similarity of the prefix region to the old policy’s outputs. However, it does \emph{not} constrain the diversity of the rollouts in a training batch at each step when compared to standard GRPO. This preservation of diversity can be attributed to the fact that, as RL training progresses, generated rollouts typically become less diverse. In contrast, the verified prefixes reused by SPEC-RL are drawn from earlier epochs and often retain greater diversity. These diverse prefixes naturally lead to varied subsequent rollouts.
In practice, the trajectory variation at each training step, which is essential for exploration and effective learning, remains fully intact. The observed trends confirm that SPEC-RL maintains, and in some cases even enhances, final task performance.

\section{Related Work}
\paragraph{Efficiency in RLVR.}
Recent studies explore RLVR efficiency from multiple perspectives.
Some parallelize rollout generation to increase throughput~\citep{xu2025not}, though the gains are often limited by computation and synchronization overhead.
Others improve efficiency by modifying optimization objectives or introducing additional constraints~\citep{brantley2025astarpo,lin2025cppo}, or by reorganizing training data to increase the proportion of informative samples~\citep{liu2025prefixgrouper,zhang2025sortedrl}.
Heuristic and system-level designs further reduce wasted rollouts and stabilize training~\citep{yu2025dapo,zheng2025greso}.
More recently, RhymeRL~\citep{he2025history} shows that suffix-tree based speculative decoding can reduce rollout latency by reusing draft tokens from historical responses.
In contrast, our method adopts a lightweight and plug-and-play speculative reuse mechanism tailored for RLVR.
Instead of maintaining complex suffix-tree structures, we directly reuse verified prefixes through speculative decoding with a lenience mechanism.
This design significantly lowers system complexity and integration cost, while enabling more effective reuse and yielding stronger acceleration in practice, without modifying RLVR objectives or update rules.

\paragraph{Speculative decoding.}
Speculative decoding has become a widely studied technique for accelerating the generation process.
\cite{leviathan2023speculativedecoding} first proposed using a lightweight draft model to propose tokens, which are then verified by a stronger target model.
Subsequent works extended this principle in several directions.
Medusa~\citep{cai2024medusa} attaches multiple proposal heads to increase parallelism, while Cascade Speculative Drafting~\citep{chen2024cascade} chains multiple draft–target stages for additional speedups. 
Other variants study accepting multiple tokens at once~\citep{qin2024optimized-multi-token-accept} or verifying blocks of tokens instead of single steps~\citep{sun2024block-verify}. 
We adapt speculative decoding to RLVR training by reusing outputs from the prior policy as drafts and verifying them under the current policy, enabling prefix reuse during rollout without additional models and remaining compatible with high-throughput rollout engines.

\section{Conclusion}
\label{sec:conclusion}
We address the rollout bottleneck in reinforcement learning with verifiable rewards (RLVR) by introducing \ourmethod, which integrates speculative decoding into rollout generation. Instead of regenerating trajectories from scratch, \ourmethod\ treats previous-epoch rollouts as implicit drafts: tokens are verified under the current policy to form a verified prefix, then generation resumes from the first rejection position. A lenience parameter $\ell$ modulates the acceptance rule, trading off reuse and exploration.
Experiments on Qwen and LLaMA models with GRPO, PPO, and DAPO show consistent 2--3$\times$ rollout speedups with largely preserved, and sometimes improved, reasoning and OOD performance. This demonstrates that rollout redundancy can be systematically exploited without modifying objectives, rewards, or update rules.

\section*{Limitations}

The proposed approach relies on speculative reuse of cached responses generated by earlier policies, and its acceleration benefit therefore depends on the quality and diversity of the cache. 
When the cached pool becomes stale or overly homogeneous, speculative reuse may provide limited exploratory signal and reduce its effectiveness. 
While fixed lenience reuse settings are sufficient in our experiments, more principled adaptive lenience scheduling strategies remain to be explored. 
In addition, SPEC-RL does not provide acceleration in the very early stage of training, since speculative reuse requires a non-empty cache, leading to a cold-start phase in the first epoch.
Finally, SPEC-RL focuses on single-turn reasoning tasks, and extending speculative reuse to more complex settings, such as multi-turn interactions, multimodal large language models, and agentic RL scenarios, remains an important direction for future work.



\bibliography{custom}

\clearpage
\appendix
\section{Experimental Details}
\label{appendix:app}

\begin{table*}
\centering

\resizebox{\textwidth}{!}{%
\setlength{\tabcolsep}{4.5pt}
\begin{tabular}{l r r r r r r r r r r r r r r}
\toprule
& \multicolumn{1}{c}{\textbf{End-to-end (h)}} & \multicolumn{13}{c}{\textbf{Average step time (s)}} \\
\cmidrule(lr){2-2} \cmidrule(lr){3-15}
\textbf{Algorithm} &
\textbf{Total} &
\textbf{Total} & $\boldsymbol{\Delta}$ vs. base &
\textbf{verification} & \textbf{rollout} & \textbf{assembly} &
\textbf{reward} & \textbf{old-log-probs} & \textbf{ref} & \textbf{values} &
\textbf{adv} & \textbf{update-critic} & \textbf{update-actor} & \textbf{others} \\
\midrule
\rowcolor{SpecRowGray}
\multicolumn{15}{c}{\textbf{\textit{Qwen-3-1.7B-Base}}} \\
GRPO     & 12.63 & 505.1 & -- & --    & 309.9 & --   & 91.0 & 17.2 & 15.8 & --    & 0.4 & --    & 56.0 & 14.9 \\
\hspace{2mm}$\hookrightarrow$~+~\ourmethod  
         & 8.65  & 346.0 & \textcolor{ForestGreen}{\(\downarrow\,159.1\)} & \textcolor{BrickRed}{22.1} & 135.2 \textcolor{ForestGreen}{(2.29$\times$)} & \textcolor{BrickRed}{1.5} & 81.0 & 17.1 & 16.3 & -- & 0.5 & -- & 56.2 & 16.2 \\
PPO      & 14.10 & 563.9 & -- & --    & 308.1 & --   & 100.5 & 17.2 & --    & 14.0 & 4.7 & 46.0 & 56.5 & 16.9 \\
\hspace{2mm}$\hookrightarrow$~+~\ourmethod  
         & 10.78 & 431.2 & \textcolor{ForestGreen}{\(\downarrow\,132.7\)} & \textcolor{BrickRed}{22.7} & 158.6 \textcolor{ForestGreen}{(1.94$\times$)} & \textcolor{BrickRed}{1.4} & 94.1 & 17.3 & -- & 13.8 & 4.6 & 45.0 & 55.5 & 18.1 \\
DAPO     & 11.10 & 443.8 & -- & --    & 301.3 & --   & 93.1 &  8.6 & --    & -- & 0.3 & -- & 25.9 & 14.6 \\
\hspace{2mm}$\hookrightarrow$~+~\ourmethod  
         & 7.90  & 316.0 & \textcolor{ForestGreen}{\(\downarrow\,127.9\)} & \textcolor{BrickRed}{21.0} & 139.0 \textcolor{ForestGreen}{(2.17$\times$)} & \textcolor{BrickRed}{1.4} & 97.9 & 18.1 & -- & -- & 0.2 & -- & 25.9 & 12.7 \\
\midrule
\rowcolor{SpecRowGray}
\multicolumn{15}{c}{\textbf{\textit{Qwen-3-8B-Base}}} \\
GRPO     & 31.66 & 1266.4 & -- & --    & 768.2 & --   & 73.2 & 66.8 & 66.9 & --    & 4.2 & --     & 263.8 & 23.4 \\
\hspace{2mm}$\hookrightarrow$~+~\ourmethod  
         & 21.03 & 841.0  & \textcolor{ForestGreen}{\(\downarrow\,425.4\)} & \textcolor{BrickRed}{74.7} & 305.8 \textcolor{ForestGreen}{(2.51$\times$)} & \textcolor{BrickRed}{1.3} & 61.4 & 63.8 & 62.4 & -- & 4.9 & -- & 248.8 & 18.0 \\
PPO      & 34.85 & 1393.9 & -- & --    & 676.7 & --   & 70.5 & 65.4 & --    & 57.4 & 4.2 & 224.1 & 260.4 & 35.3 \\
\hspace{2mm}$\hookrightarrow$~+~\ourmethod  
         & 26.97 & 1078.8 & \textcolor{ForestGreen}{\(\downarrow\,315.1\)} & \textcolor{BrickRed}{71.5} & 349.3 \textcolor{ForestGreen}{(1.94$\times$)} & \textcolor{BrickRed}{1.4} & 64.9 & 59.6 & -- & 52.1 & 4.9 & 205.9 & 236.9 & 32.5 \\
DAPO     & 24.29 & 971.8  & -- & --    & 699.2 & --   & 64.4 & 66.3 & --    & -- & 0.1 & -- & 121.1 & 20.7 \\
\hspace{2mm}$\hookrightarrow$~+~\ourmethod  
         & 12.90 & 515.9  & \textcolor{ForestGreen}{\(\downarrow\,455.9\)} & \textcolor{BrickRed}{51.0} & 243.0 \textcolor{ForestGreen}{(2.88$\times$)} & \textcolor{BrickRed}{1.1} & 54.0 & 51.2 & -- & -- & 0.1 & -- & 97.5 & 18.0 \\
\midrule
\rowcolor{SpecRowGray}
\multicolumn{15}{c}{\textbf{\textit{LLaMA-3.2-1B-Instruct}}} \\
GRPO     & 10.20 & 408.0 & -- & --    & 229.7 & --   & 105.8 & 12.6 & 11.5 & --    & 0.4 & --    & 34.7 & 13.2 \\
\hspace{2mm}$\hookrightarrow$~+~\ourmethod  
         & 7.28  & 291.3 & \textcolor{ForestGreen}{\(\downarrow\,116.7\)} & \textcolor{BrickRed}{17.2} & 88.3 \textcolor{ForestGreen}{(2.60$\times$)} & \textcolor{BrickRed}{1.4} & 110.4 & 13.0 & 11.9 & -- & 0.5 & -- & 34.4 & 14.4 \\
PPO      & 10.94 & 437.6 & -- & --    & 218.9 & --   & 117.6 & 12.5 & --    & 10.0 & 4.8 & 10.0 & 32.6 & 31.3 \\
\hspace{2mm}$\hookrightarrow$~+~\ourmethod  
         & 8.60  & 344.0 & \textcolor{ForestGreen}{\(\downarrow\,93.6\)}  & \textcolor{BrickRed}{17.5} & 108.9 \textcolor{ForestGreen}{(2.01$\times$)} & \textcolor{BrickRed}{1.3} & 110.9 & 12.4 & -- & 10.1 & 4.6 & 10.1 & 34.3 & 33.8 \\
DAPO     & 9.77  & 328.4 & -- & --    & 198.4 & --   & 100.8 & 11.2 & --    & -- & 0.1 & -- & 9.6  & 8.5 \\
\hspace{2mm}$\hookrightarrow$~+~\ourmethod  
         & 6.97  & 238.4 & \textcolor{ForestGreen}{\(\downarrow\,90.0\)} & \textcolor{BrickRed}{13.4} & 80.0 \textcolor{ForestGreen}{(2.48$\times$)} & \textcolor{BrickRed}{1.1} & 110.5 & 11.5 & -- & -- & 0.1 & -- & 9.9 & 12.0 \\
\bottomrule
\end{tabular}
}
\caption{End-to-end training time comparison across models and algorithms. 
We report both the \textbf{wall-clock training hours} (``End-to-end (h)'') and the \textbf{average step time} (``Total (s)'') with a detailed breakdown.
\textit{validation} refers to our newly introduced speculative decoding process that verifies old-policy rollouts in parallel; 
\textit{assemble} denotes combining verified prefixes with newly generated continuations to form complete rollouts; 
the remaining parts (\textit{reward}, \textit{old-log-probs}, \textit{ref}, \textit{values}, \textit{adv}, \textit{update-critic}, \textit{update-actor}, \textit{others}) follow the standard pipeline of the verl ramework in execution order.
}

\label{tab:e2e_breakdown}
\end{table*}

This section provides additional details on experimental settings, shared training configurations, hyperparameters, and reward design.
We describe the common rollout and optimization setups across models and algorithms, as well as the evaluation protocols used for all benchmarks.

\subsection{Hyperparameters}
\label{appendix:hyper-params}
We report the shared training settings (model families, rollout engine, batch size, sequence lengths, training steps, and optimizer details), as well as the algorithm-specific configurations. 
All experiments use Qwen-3-1.7B-Base, Qwen-3-8B-Base, Qwen-3-14B-Base, and LLaMA-3.2-1B as backbone models. 
Rollouts are generated using vLLM (rollout $N=8$) with a global batch size of 1024. 
The maximum prompt length is 1,024 tokens, and the maximum response length is 4,096 tokens. 
For optimization, the actor is trained using AdamW (learning rate $5\times 10^{-7}$, weight decay 0.01, and gradient clipping of 1.0). 
For PPO, the critic is additionally optimized with AdamW (learning rate $1\times 10^{-5}$, weight decay 0.01, clipping 1.0). 
Algorithm-specific differences are as follows. 
GRPO enables KL regularization with a coefficient of $0.0001$, whereas PPO and DAPO disable KL regularization. 
DAPO further adopts a wider clipping range (high = 0.28, $c=10$) compared to GRPO and PPO (high = 0.2, $c=3$). 
Additionally, DAPO utilizes dynamic sampling. 
To ensure fair comparison with GRPO and PPO, we control for the total amount of rollout data: each training step in DAPO corresponds to multiple generation steps, and the evaluation interval is reduced from every 10 steps to every 5 steps.
\ourmethod\ uses default lenience values of $e^{0.5}$ for GRPO, $e^{0.3}$ for PPO, and $e^{0.15}$ for DAPO, chosen via grid search to balance rollout efficiency and stability. 
All methods employ the \texttt{math-verify} reward, which assigns $+1$ if the final boxed or numeric answer matches the ground truth and $0$ otherwise. 
This simple, deterministic design ensures that the reward is aligned with evaluation metrics across benchmarks.

We use a rule-based reward function that depends solely on the correctness of the final answer. 
Specifically, we utilize the \texttt{math-verify} library to verify each generated solution: if the predicted answer matches the reference, the model receives a reward of $+1$, and otherwise, $0$. 
The \texttt{math-verify} library is responsible for parsing the model output, extracting the final boxed or numeric answer, and checking it against the ground truth. 
No format-based shaping or auxiliary heuristics are used. 
This choice maintains a simple, deterministic, and aligned reward signal across all benchmarks, aligning with the evaluation objective.

\subsection{Evaluation Setups}
\label{appendix:eval-setup}

Our evaluation setup largely follows prior work~\citep{zeng2025simplerl, yang2024qwen2-eval}, ensuring consistency and comparability with established baselines. 
For all math reasoning benchmarks, including AMC23, AIME24, MATH-500, Minerva Math, OlympiadBench, and MMLU-STEM, we use a maximum generation length of 16{,}000 tokens, with nucleus sampling ($p=0.95$) and temperature set to 1.0.
For the more challenging benchmarks, we increase the number of evaluation samples and report Pass@1 averaged over 16 samplings for AMC23 and over 32 samplings for AIME24.
For IFEval, we employ the lighteval~\citep{lighteval} framework for evaluation, maintaining the same decoding parameters as those used in the math reasoning benchmarks. 
This uniform setup ensures that all comparisons focus on the effects of \ourmethod, rather than variations in decoding configurations.
For experiments on DeepMath-6K, we report the performance at step 90 (corresponding to 15 epochs with 6,144 examples and a batch size of 1,024). For SimpleRL-8K, we report the performance at step 100.

\section{Additional Results and Analyses}
This section reports additional experimental results and analyses that complement the main findings.
We present extended ablations across datasets (DeepMath-6K vs.\ SimpleRL-8K) and training-set sizes (2K–6K), detailed efficiency analyses, and end-to-end time breakdowns.
We further include full step-level training trajectories, additional baseline comparisons, and case studies to provide a more comprehensive view of the behavior of \ourmethod\ throughout training.

\begin{table*}[!t]
\centering
\resizebox{\linewidth}{!}{%
\begin{tabular}{l r r r r r r r r r r}
\toprule
\multirow{3}{*}{\textbf{Algorithm}} &
\multicolumn{2}{c}{\textbf{Rollout Efficiency}} &
\multicolumn{5}{c}{\textbf{Math Reasoning}} &
\multicolumn{2}{c}{\textbf{OOD}} &
\multirow{3}{*}{\textbf{AVG}} \\
\cmidrule(lr){2-3}\cmidrule(lr){4-8}\cmidrule(lr){9-10}
& \textbf{Tokens (M)} & \textbf{Speedup} &
\textbf{AMC23} & \textbf{AIME24} &
\textbf{\begin{tabular}[c]{@{}c@{}}MATH\\ 500\end{tabular}} &
\textbf{\begin{tabular}[c]{@{}c@{}}Minerva\\ Math\end{tabular}} &
\textbf{\begin{tabular}[c]{@{}c@{}}Olympiad\\ Bench\end{tabular}} &
\textbf{\begin{tabular}[c]{@{}c@{}}MMLU\\ STEM\end{tabular}} &
\textbf{IFEval} & \\
\midrule

\rowcolor{SpecRowGray}
\multicolumn{11}{c}{\textbf{\textit{Qwen3-14B-Base}}} \\
Base Model & - & - & 45.2 & 9.1 & 68.6 & 28.3 & 33.5 & 68.3 & 43.6 & 42.4 \\
\hdashline
GRPO & 587.6 & 1.00$\times$ & 66.2 & 21.1 & \bf 87.2 & 43.4 & 49.5 & 84.7 & 49.7 & 57.4 \\
\specrow
\hspace{2mm}$\hookrightarrow$~+~\ourmethod
& \bf 234.4 & \bf 2.09$\times$ & \bf 67.8 & \bf 26.2 & 86.8 & \bf 46.3 & \bf 50.7 & \bf 84.9 & \bf 54.9 & \bf 59.7 \\
PPO & 579.2 & 1.00$\times$ & 67.5 & 22.1 & \bf 86.6 & 41.9 & 50.2 & \bf 87.1 & \bf 49.5 & 57.8 \\
\specrow
\hspace{2mm}$\hookrightarrow$~+~\ourmethod
& \bf 310.1 & \bf 1.66$\times$ & \bf 75.5 & \bf 29.1 & 86.2 & 41.9 & \bf 53.6 & 82.3 & 47.1 & \bf 59.4 \\
DAPO & 641.1 & 1.00$\times$ & 62.7 & 16.4 & 84.6 & 41.9 & 45.2 & \bf 72.9 & 49.9 & 53.4 \\
\specrow
\hspace{2mm}$\hookrightarrow$~+~\ourmethod
& \bf 242.8 & \bf 2.46$\times$ & \bf 65.6 & \bf 19.9 & \bf 85.2 & \bf 42.6 & \bf 48.4 & 72.4 & \bf 51.4 & \bf 55.1 \\

\bottomrule
\end{tabular}
}
\caption{Results on \textit{Qwen3-14B-Base} across different RLVR algorithms on DeepMath-6K.}
\label{tab:qwen3_14b_results}
\end{table*}

\subsection{End-to-End Time Breakdown}
\label{appendix:e2e_breakdown}
Table~\ref{tab:e2e_breakdown} reports the per-stage breakdown of training time. 
In the vanilla baseline, rollout generation dominates the runtime, often accounting for more than 60\% of the total.
With \ourmethod, this cost is largely shifted into a lightweight verification stage, where cached rollouts are first verified in parallel under the current policy and then evaluated by the speculative decoding rule to determine the rejection position, and a minimal assembly stage, where verified prefixes and regenerated suffixes are merged into complete responses.
Both stages add only minor overhead (on Qwen-3-1.7B-Base, verification $\sim$20s and assembly $\sim$1–2s), while the total step time is reduced by about 129–161s, making the extra cost negligible compared to the savings from reduced rollout.
For instance, on Qwen-3-8B-Base/GRPO, the rollout time decreases from 768.2s to 305.8s, while all other stages, such as reward computation and policy updates, remain nearly unchanged. 
Overall, although these new stages slightly increase non-rollout costs, the dominant effect is the 2–3 times reduction in rollout tokens, yielding substantially faster end-to-end training.

\begin{table*}[!t]
\centering
\resizebox{\textwidth}{!}{%
\begin{tabular}{l r r r r r r r r r r}
\toprule
\multirow{3}{*}{\textbf{Algorithm}} &
\multicolumn{2}{c}{\textbf{Rollout Efficiency}} &
\multicolumn{5}{c}{\textbf{Math Reasoning}} &
\multicolumn{2}{c}{\textbf{OOD}} &
\multirow{3}{*}{\textbf{AVG}} \\
\cmidrule(lr){2-3}\cmidrule(lr){4-8}\cmidrule(lr){9-10}
& \textbf{Tokens (M)} & \textbf{Speedup} &
\textbf{AMC23} & \textbf{AIME24} &
\textbf{\begin{tabular}[c]{@{}c@{}}MATH\\ 500\end{tabular}} &
\textbf{\begin{tabular}[c]{@{}c@{}}Minerva\\ Math\end{tabular}} &
\textbf{\begin{tabular}[c]{@{}c@{}}Olympiad\\ Bench\end{tabular}} &
\textbf{\begin{tabular}[c]{@{}c@{}}MMLU\\ STEM\end{tabular}} &
\textbf{IFEval} & \\
\midrule
\rowcolor{SpecRowGray}
\multicolumn{11}{c}{\textbf{\textit{Deepmath-6K (Qwen-3-1.7B-Base)}}} \\
GRPO               & 554.8 & 1.00$\times$  & 35.0 & 7.8 & 64.4 & 26.5 & 25.5 & \bf 60.7 & 24.4 & 34.9 \\
\specrow
\hspace{2mm}$\hookrightarrow$~+~\ourmethod
& \bf 182.7 & \bf 2.29$\times$ & \bf 38.3 & \bf 9.1 & \bf 68.0 & \bf 29.4 & \bf 29.3 & 58.3 & \bf 28.8 & \bf 37.3 \\
\midrule
\rowcolor{SpecRowGray}
\multicolumn{11}{c}{\textbf{\textit{SimpleRL-8K (Qwen-3-1.7B-Base)}}} \\
GRPO               & 639.4 & 1.00$\times$ & 40.5 & 8.6 & 68.2 & 27.2 & 30.5 & 49.4 & 24.0 & 35.5 \\
\specrow
\hspace{2mm}$\hookrightarrow$~+~\ourmethod & \bf 354.0 & \bf 1.53$\times$ & \bf 41.2 & \bf 12.9 & \bf 72.2 & 27.2 & \bf 32.1 & \bf 57.4 & \bf 27.7 & \bf 38.7 \\
\bottomrule
\end{tabular}
}
\caption{Ablation study on different training datasets. Results show that our method maintains improvements in rollout efficiency and accuracy across both Deepmath-6K and SimpleRL-8K settings.}
\label{tab:ablation_datasets}
\end{table*}

\subsection{Scalability to Larger Models}
\label{appendix:larger-model-14b}
To complement the main results in Table ~\ref{tab:math_reasoning}, we further evaluate \ourmethod\ on a larger backbone, \textit{Qwen3-14B-Base}, to examine whether the efficiency gains and performance trends observed on smaller models persist at increased scale.
The main results are summarized in Table~\ref{tab:qwen3_14b_results}, and the complete training trajectories tables are reported below in Section ~\ref{appendix:details-steps-algo-models}.

Across all three RLVR algorithms (GRPO, PPO, and DAPO), \ourmethod\ consistently achieves substantial reductions in rollout tokens, yielding speedups of $1.66\times$ to $2.46\times$ compared to the corresponding vanilla baselines.
These efficiency gains are comparable to, and in some cases larger than, those observed on smaller models, indicating that speculative reuse remains effective even when scaling to a 14B parameter backbone.
In terms of reasoning performance, \ourmethod\ preserves or improves accuracy across both in-domain and OOD benchmarks.
For GRPO and DAPO, SPEC-RL slightly improves average accuracy while reducing rollout cost by more than half.
Under PPO, we observe notable gains on challenging math benchmarks such as AMC23 and AIME24, with overall performance remaining competitive despite the aggressive reduction in generation cost.
These results suggest that the reuse mechanism does not degrade learning signals at larger scale, and may even stabilize optimization by reducing redundant sampling.
Overall, the results on \textit{Qwen3-14B-Base} demonstrate that \ourmethod\ scales favorably to larger models, delivering consistent efficiency improvements without sacrificing reasoning performance.
This provides further evidence that speculative reuse is a practical and robust acceleration strategy for RLVR training across model sizes.

\subsection{Impact of Training Set Size on Acceleration}
\label{appendix:data-size}
\begin{figure}[!t]
    \centering
    \includegraphics[width=0.5\textwidth]{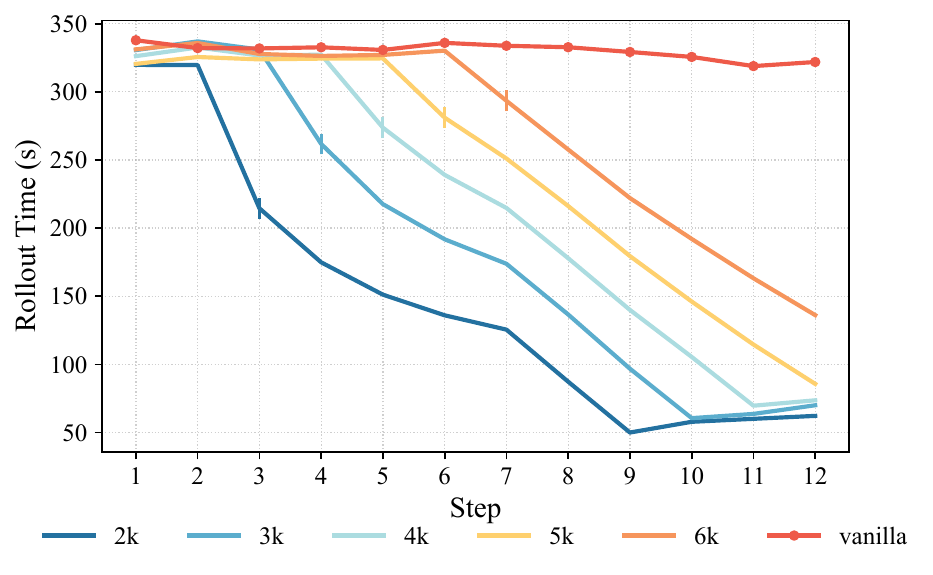}
    \caption{Rollout time under different training set sizes (2K–6K) with GRPO on Qwen-3-1.7B-Base. 
    Markers highlight the first reuse points at the start of epoch~2, when SPEC-RL begins to accelerate rollouts.}   
    \label{fig:cumulative-gen}
\end{figure}

Since \ourmethod\ accelerates training by reusing cached rollouts from the previous epoch, acceleration can only take effect starting from the second epoch. 
To study how dataset size influences this effect, we vary the training set size to 2K, 3K, 4K, 5K, and 6K samples, and train Qwen-3-1.7B-Base with GRPO. 
Figure~\ref{fig:cumulative-gen} reports the rollout time across training steps. 

We observe that smaller datasets lead to earlier reuse opportunities, since epochs finish more quickly and the second epoch arrives sooner. 
For example, with 2K samples, the rollout time drops sharply after step 3, whereas with 6K samples, the reduction is delayed until later steps. 
Across all settings, rollout time decreases steadily once reuse begins, with larger speedups achieved as training progresses. 
The markers in the figure denote the first reuse points (the first step of epoch~2), where SPEC-RL begins to take effect. 
This analysis confirms that the efficiency gains of \ourmethod\ depend not only on algorithm and model choice, but also on the dataset size, which determines how soon reuse can be activated during training.

\begin{table*}[!t]
\centering
\resizebox{\textwidth}{!}{%
\begin{tabular}{l r r r r r r r r r r r}
\toprule
\multirow{3}{*}{\textbf{Algorithm}} &
\multirow{3}{*}{\textbf{Step}} &
\multicolumn{2}{c}{\textbf{Rollout Efficiency}} &
\multicolumn{5}{c}{\textbf{Math Reasoning}} &
\multicolumn{2}{c}{\textbf{OOD}} &
\multirow{3}{*}{\textbf{AVG}} \\
\cmidrule(lr){3-4}\cmidrule(lr){5-9}\cmidrule(lr){10-11}
& & \textbf{Tokens (M)} & \textbf{Speedup} &
\textbf{AMC23} &
\textbf{AIME24} &
\textbf{\begin{tabular}[c]{@{}c@{}}MATH\\ 500\end{tabular}} &
\textbf{\begin{tabular}[c]{@{}c@{}}Minerva\\ Math\end{tabular}} &
\textbf{\begin{tabular}[c]{@{}c@{}}Olympiad\\ Bench\end{tabular}} &
\textbf{\begin{tabular}[c]{@{}c@{}}MMLU\\ STEM\end{tabular}} &
\textbf{IFEval} & \\
\midrule
Base Model & 0 & - & - & 21.9 & 2.5 & 45.0 & 12.5 & 16.7 & 39.3 & 17.9 & 22.3 \\
\hdashline
GRPO & 10  & 61.5  & 1.00$\times$ & 31.7 & 4.6 & 57.6 & 19.1 & 22.7 & 44.4 & 21.1 & 28.7 \\
\specrow \hspace{2mm}$\hookrightarrow$~+~\ourmethod & 10  & 50.5  & 1.16$\times$ & 33.1 & 5.7 & 57.8 & 19.1 & 24.9 & 44.9 & 21.3 & 29.5 \\
GRPO & 20  & 123.1 & 1.00$\times$ & 35.0 & 5.3 & 60.6 & 22.1 & 25.9 & 46.1 & 23.7 & 31.2 \\
\specrow \hspace{2mm}$\hookrightarrow$~+~\ourmethod & 20  & 84.3  & 1.33$\times$ & 34.5 & 6.9 & 62.2 & 20.6 & 28.0 & 46.4 & 20.9 & 31.4 \\
GRPO & 30  & 185.3 & 1.00$\times$ & 33.9 & 6.5 & 63.8 & 24.3 & 27.0 & 46.1 & 23.7 & 32.2 \\
\specrow \hspace{2mm}$\hookrightarrow$~+~\ourmethod & 30  & 112.2 & 1.47$\times$ & 36.1 & 6.7 & 61.4 & 26.8 & 28.7 & 49.0 & 25.1 & 33.4 \\
GRPO & 40  & 247.7 & 1.00$\times$ & 33.8 & 7.5 & 63.0 & 22.4 & 28.1 & 46.9 & 21.1 & 31.8 \\
\specrow \hspace{2mm}$\hookrightarrow$~+~\ourmethod & 40  & 136.8 & 1.57$\times$ & 39.4 & 8.6 & 65.8 & 24.3 & 29.5 & 49.8 & 26.1 & 34.8 \\
GRPO & 50  & 312.5 & 1.00$\times$ & 35.9 & 7.4 & 65.2 & 27.2 & 26.8 & 48.7 & 23.1 & 33.5 \\
\specrow \hspace{2mm}$\hookrightarrow$~+~\ourmethod & 50  & 171.4 & 1.58$\times$ & 41.6 & 9.6 & 66.2 & 28.7 & 30.8 & 52.9 & 26.2 & 36.6 \\
GRPO & 60  & 377.7 & 1.00$\times$ & 39.5 & 7.3 & 64.6 & 26.5 & 28.0 & 48.5 & 23.3 & 34.0 \\
\specrow \hspace{2mm}$\hookrightarrow$~+~\ourmethod & 60  & 206.9 & 1.58$\times$ & 39.5 & 9.5 & 67.8 & 26.8 & 32.0 & 53.5 & 25.5 & 36.4 \\
GRPO & 70  & 444.8 & 1.00$\times$ & 39.5 & 8.3 & 66.0 & 26.5 & 26.5 & 48.6 & 20.9 & 33.8 \\
\specrow \hspace{2mm}$\hookrightarrow$~+~\ourmethod & 70  & 246.5 & 1.56$\times$ & 42.7 & 10.5 & 70.0 & 28.3 & 31.3 & 53.8 & 27.9 & 37.8 \\
GRPO & 80  & 512.9 & 1.00$\times$ & 39.5 & 7.4 & 66.8 & 26.5 & 30.5 & 47.5 & 25.0 & 34.7 \\
\specrow \hspace{2mm}$\hookrightarrow$~+~\ourmethod & 80  & 283.2 & 1.55$\times$ & 44.8 & 10.7 & 68.4 & 26.5 & 31.7 & 55.5 & 25.0 & 37.5 \\
GRPO & 90  & 582.6 & 1.00$\times$ & 38.6 & 7.5 & 66.6 & 25.7 & 29.5 & 48.4 & 24.6 & 34.4 \\
\specrow \hspace{2mm}$\hookrightarrow$~+~\ourmethod & 90  & 324.8 & 1.53$\times$ & 44.2 & 11.6 & 70.4 & 27.9 & 31.9 & 55.4 & 27.0 & 38.3 \\
GRPO & 100 & 639.4 & 1.00$\times$ & 40.5 & 8.6 & 68.2 & 27.2 & 30.5 & 49.4 & 24.0 & 35.5 \\
\specrow \hspace{2mm}$\hookrightarrow$~+~\ourmethod & 100 & 354.0 & 1.54$\times$ & 41.2 & 12.9 & 72.2 & 27.2 & 32.1 & 57.4 & 27.7 & 38.7 \\
\bottomrule
\end{tabular}
}
\caption{Intermediate training results of Qwen-3-1.7B-Base on \textbf{SimpleRL-8K} with GRPO and \ourmethod.
We report rollout efficiency and accuracy every 10 training steps, with GRPO and its \ourmethod\ variant interleaved.}
\label{tab:grpo-1.7b-simplerl}
\end{table*}

\begin{table*}[!t]
\centering
\resizebox{\textwidth}{!}{%
\begin{tabular}{l r r r r r r r r r r r}
\toprule
\multirow{3}{*}{\textbf{Algorithm}} &
\multirow{3}{*}{\textbf{Step}} &
\multicolumn{2}{c}{\textbf{Rollout Efficiency}} &
\multicolumn{5}{c}{\textbf{Math Reasoning}} &
\multicolumn{2}{c}{\textbf{OOD}} &
\multirow{3}{*}{\textbf{AVG}} \\
\cmidrule(lr){3-4}\cmidrule(lr){5-9}\cmidrule(lr){10-11}
& & \textbf{Tokens (M)} & \textbf{Speedup} &
\textbf{AMC23} &
\textbf{\begin{tabular}[c]{@{}c@{}}AIME24\end{tabular}} &
\textbf{\begin{tabular}[c]{@{}c@{}}MATH\\ 500\end{tabular}} &
\textbf{\begin{tabular}[c]{@{}c@{}}Minerva\\ Math\end{tabular}} &
\textbf{\begin{tabular}[c]{@{}c@{}}Olympiad\\ Bench\end{tabular}} &
\textbf{\begin{tabular}[c]{@{}c@{}}MMLU\\ STEM\end{tabular}} &
\textbf{IFEval} & \\
\midrule
Base Model & 0 & - & - & 21.9 & 2.5 & 45.0 & 12.5 & 16.7 & 39.3 & 17.9 & 22.3 \\
\hdashline
GRPO & 10  & 65.8  & 1.00$\times$ & 27.8 & 4.3 & 54.0 & 18.8 & 22.8 & 43.9 & 20.3 & 27.4 \\
\specrow \hspace{2mm}$\hookrightarrow$~+~Random Reuse & 10  & 58.0  & 1.11$\times$ & 31.7 & 3.8 & 57.8 & 21.0 & 21.0 & 42.9 & 22.0 & 28.6 \\
GRPO & 20  & 127.4 & 1.00$\times$ & 31.2 & 6.1 & 57.8 & 25.4 & 24.7 & 45.5 & 20.3 & 30.1 \\
\specrow \hspace{2mm}$\hookrightarrow$~+~Random Reuse & 20  & 98.9  & 1.43$\times$ & 35.5 & 5.3 & 59.4 & 24.6 & 24.1 & 46.0 & 25.9 & 31.5 \\
GRPO & 30  & 187.7 & 1.00$\times$ & 33.8 & 5.9 & 60.4 & 22.4 & 25.0 & 47.9 & 20.9 & 30.9 \\
\specrow \hspace{2mm}$\hookrightarrow$~+~Random Reuse & 30  & 134.1 & 1.61$\times$ & 32.0 & 7.6 & 63.2 & 26.8 & 24.9 & 50.6 & 29.2 & 33.5 \\
GRPO & 40  & 248.3 & 1.00$\times$ & 36.4 & 6.6 & 65.0 & 23.5 & 24.7 & 50.7 & 21.3 & 32.6 \\
\specrow \hspace{2mm}$\hookrightarrow$~+~Random Reuse & 40  & 165.4 & 1.75$\times$ & 38.9 & 7.4 & 64.0 & 26.1 & 28.4 & 56.8 & 27.7 & 35.6 \\
GRPO & 50  & 309.1 & 1.00$\times$ & 36.6 & 5.9 & 64.2 & 25.7 & 24.6 & 53.7 & 25.1 & 33.7 \\
\specrow \hspace{2mm}$\hookrightarrow$~+~Random Reuse & 50  & 194.1 & 1.89$\times$ & 36.1 & 6.8 & 63.2 & 27.9 & 27.0 & 57.7 & 25.7 & 34.9 \\
GRPO & 60  & 370.4 & 1.00$\times$ & 38.0 & 6.7 & 63.6 & 28.3 & 26.7 & 56.0 & 24.0 & 34.8 \\
\specrow \hspace{2mm}$\hookrightarrow$~+~Random Reuse & 60  & 221.8 & 2.03$\times$ & 37.0 & 6.7 & 64.2 & 24.6 & 25.9 & 59.2 & 26.1 & 34.8 \\
GRPO & 70  & 431.9 & 1.00$\times$ & 35.3 & 8.0 & 61.8 & 26.5 & 28.1 & 55.5 & 26.2 & 34.5 \\
\specrow \hspace{2mm}$\hookrightarrow$~+~Random Reuse & 70  & 249.6 & 2.14$\times$ & 36.7 & 5.4 & 63.2 & 27.2 & 26.1 & 60.9 & 25.5 & 35.0 \\
GRPO & 80  & 493.5 & 1.00$\times$ & 36.1 & 7.2 & 64.4 & 24.3 & 26.4 & 59.4 & 25.0 & 34.7 \\
\specrow \hspace{2mm}$\hookrightarrow$~+~Random Reuse & 80  & 277.5 & 2.25$\times$ & 27.5 & 4.9 & 57.0 & 21.7 & 22.7 & 44.6 & 22.4 & 28.7 \\
GRPO & 90  & 554.8 & 1.00$\times$ & 35.0 & 7.8 & 64.4 & 26.5 & 25.5 & 60.7 & 24.4 & 34.9 \\
\specrow \hspace{2mm}$\hookrightarrow$~+~Random Reuse & 90  & 304.5 & 2.35$\times$ & 30.3 & 5.0 & 60.4 & 21.7 & 25.3 & 53.1 & 24.0 & 31.4 \\
\bottomrule
\end{tabular}
}
\caption{Intermediate training results of Qwen-3-1.7B-Base on DeepMath-6K with GRPO and \textbf{Random Reuse}.
We report rollout efficiency and accuracy every 10 training steps, with GRPO and its Random Reuse variant interleaved.}
\label{tab:grpo-1.7b-randomreuse}
\end{table*}

\begin{table*}[!t]
\centering
\resizebox{\textwidth}{!}{%
\begin{tabular}{l r r r r r r r r r r r r}
\toprule
\multirow{3}{*}{\textbf{Algorithm}} &
\multirow{3}{*}{\textbf{Step}} &
\multicolumn{2}{c}{\textbf{Rollout Efficiency}} &
\multicolumn{6}{c}{\textbf{Math Reasoning}} &
\multicolumn{1}{c}{\textbf{OOD}} &
\multirow{3}{*}{\textbf{AVG}} \\
\cmidrule(lr){3-4}\cmidrule(lr){5-10}\cmidrule(lr){11-11}
& &
\textbf{Tokens (M)} & \textbf{Speedup} &
\textbf{AMC23} &
\textbf{AIME24} &
\textbf{\begin{tabular}[c]{@{}c@{}}MATH\\ 500\end{tabular}} &
\textbf{\begin{tabular}[c]{@{}c@{}}Minerva\\ Math\end{tabular}} &
\textbf{\begin{tabular}[c]{@{}c@{}}Olympiad\\ Bench\end{tabular}} &
\textbf{\begin{tabular}[c]{@{}c@{}}MMLU\\ STEM\end{tabular}} &
\textbf{IFEval} & \\
\midrule
Base Model & 0 & - & - & 21.9 & 2.5 & 45.0 & 12.5 & 16.7 & 39.3 & 17.9 & 22.3 \\

\hdashline
GRPO & 10 & 65.8 & 1.00$\times$ & 27.8 & 4.3 & 54.0 & 18.8 & 22.8 & 43.9 & 20.3 & 27.4 \\
\specrow \hspace{2mm}$\hookrightarrow$~+~Delayed Reuse & 10 & 44.4 & 1.29$\times$ & 32.2 & 4.9 & 57.4 & 21.7 & 22.1 & 44.0 & 19.6 & 28.8 \\

GRPO & 20 & 127.4 & 1.00$\times$ & 31.2 & 6.1 & 57.8 & 25.4 & 24.7 & 45.5 & 20.3 & 30.1 \\
\specrow \hspace{2mm}$\hookrightarrow$~+~Delayed Reuse & 20 & 85.4 & 1.26$\times$ & 34.4 & 5.5 & 58.4 & 21.7 & 24.9 & 46.5 & 21.3 & 30.4 \\

GRPO & 30 & 187.7 & 1.00$\times$ & 33.8 & 5.9 & 60.4 & 22.4 & 25.0 & 47.9 & 20.9 & 30.9 \\
\specrow \hspace{2mm}$\hookrightarrow$~+~Delayed Reuse & 30 & 123.8 & 1.27$\times$ & 33.3 & 6.5 & 61.0 & 26.5 & 26.1 & 51.2 & 24.2 & 32.7 \\

GRPO & 40 & 248.3 & 1.00$\times$ & 36.4 & 6.6 & 65.0 & 23.5 & 24.7 & 50.7 & 21.3 & 32.6 \\
\specrow \hspace{2mm}$\hookrightarrow$~+~Delayed Reuse & 40 & 156.5 & 1.31$\times$ & 35.5 & 6.8 & 62.6 & 27.9 & 24.3 & 53.2 & 25.1 & 33.6 \\

GRPO & 50 & 309.1 & 1.00$\times$ & 36.6 & 5.9 & 64.2 & 25.7 & 24.6 & 53.7 & 25.1 & 33.7 \\
\specrow \hspace{2mm}$\hookrightarrow$~+~Delayed Reuse & 50 & 187.9 & 1.35$\times$ & 35.8 & 7.4 & 65.0 & 27.6 & 28.4 & 55.4 & 25.9 & 35.1 \\

GRPO & 60 & 370.4 & 1.00$\times$ & 38.0 & 6.7 & 63.6 & 28.3 & 26.7 & 56.0 & 24.0 & 34.8 \\
\specrow \hspace{2mm}$\hookrightarrow$~+~Delayed Reuse & 60 & 218.5 & 1.37$\times$ & 35.3 & 7.8 & 66.4 & 27.6 & 27.0 & 55.9 & 26.2 & 35.2 \\

GRPO & 70 & 431.9 & 1.00$\times$ & 35.3 & 8.0 & 61.8 & 26.5 & 28.1 & 55.5 & 26.2 & 34.5 \\
\specrow \hspace{2mm}$\hookrightarrow$~+~Delayed Reuse & 70 & 248.8 & 1.40$\times$ & 31.4 & 8.9 & 64.6 & 28.3 & 26.8 & 57.5 & 24.6 & 34.6 \\

GRPO & 80 & 493.5 & 1.00$\times$ & 36.1 & 7.2 & 64.4 & 24.3 & 26.4 & 59.4 & 25.0 & 34.7 \\
\specrow \hspace{2mm}$\hookrightarrow$~+~Delayed Reuse & 80 & 279.7 & 1.42$\times$ & 32.8 & 8.8 & 65.4 & 29.0 & 27.3 & 57.2 & 26.4 & 35.3 \\

GRPO & 90 & 554.8 & 1.00$\times$ & 35.0 & 7.8 & 64.4 & 26.5 & 25.5 & 60.7 & 24.4 & 34.9 \\
\specrow \hspace{2mm}$\hookrightarrow$~+~Delayed Reuse & 90 & 308.8 & 1.44$\times$ & 35.6 & 9.0 & 66.4 & 29.0 & 28.6 & 57.3 & 28.3 & 36.3 \\

\bottomrule
\end{tabular}
}
\caption{Interleaved comparison of GRPO and \textbf{Delayed Reuse}.
Results are reported every 10 training steps.}
\label{tab:grpo-delayed-reuse-interleave}
\end{table*}

\subsection{Random Reuse and Delayed Reuse Results}
\label{appendix:random-reuse}
For completeness, we also report the full training trajectories of the Random Reuse baseline on Qwen-3-1.7B-Base, trained with GRPO on DeepMath-6K. 

Table~\ref{tab:grpo-1.7b-randomreuse} interleaves results of GRPO and Random Reuse every 10 training steps, providing a step-wise view of rollout efficiency and accuracy. 
While Table~\ref{tab:specdec_vs_random} summarizes the overall comparison, these detailed results illustrate how Random Reuse accelerates rollouts but produces unstable performance over the course of training.

Table~\ref{tab:grpo-delayed-reuse-interleave} similarly reports the full training trajectories of the \emph{Delayed Reuse} variant, interleaving GRPO and Delayed Reuse every 10 training steps. 
This presentation provides a step-wise view of rollout efficiency and accuracy under delayed reuse scheduling, allowing direct comparison with the Random Reuse baseline. 

\subsection{Full Lenience Ablation Results}
\label{appendix:lenience-ablation}

For completeness, we provide the step-level results corresponding to the lenience ablation in Section~\ref{section:lenience-effect}. 
While Table~\ref{tab:ablation_lenience} reports only the final-step outcomes for comparison across different lenience values, we include detailed intermediate results every 10 training steps in Tables \ref{tab:grpo-1.7b-details-l1}, \ref{tab:grpo-1.7b-details-l02}, \ref{tab:grpo-1.7b-details-l05}, \ref{tab:grpo-1.7b-details-l08}, \ref{tab:grpo-1.7b-details-l20}, \ref{tab:grpo-1.7b-details-linf}. 
These tables document how rollout efficiency and accuracy evolve throughout training under various lenience settings ($\ell=1, e^{0.2}, e^{0.5}, e^{0.8}, e^{1.0}, e^{2.0},$ and $\ell \to \infty$), complementing the aggregated trends shown in Table~\ref{tab:ablation_lenience}.

\begin{table*}[!t]
\centering
\resizebox{\textwidth}{!}{%
\begin{tabular}{l r r r r r r r r r r r}
\toprule
\multirow{3}{*}{\textbf{Algorithm}} &
\multirow{3}{*}{\textbf{Step}} &
\multicolumn{2}{c}{\textbf{Rollout Efficiency}} &
\multicolumn{5}{c}{\textbf{Math Reasoning}} &
\multicolumn{2}{c}{\textbf{OOD}} &
\multirow{3}{*}{\textbf{AVG}} \\
\cmidrule(lr){3-4}\cmidrule(lr){5-9}\cmidrule(lr){10-11}
& & \textbf{Tokens (M)} & \textbf{Speedup} &
\textbf{AMC23} & \textbf{AIME24} &
\textbf{\begin{tabular}[c]{@{}c@{}}MATH\\500\end{tabular}} &
\textbf{\begin{tabular}[c]{@{}c@{}}Minerva\\Math\end{tabular}} &
\textbf{\begin{tabular}[c]{@{}c@{}}Olympiad\\Bench\end{tabular}} &
\textbf{\begin{tabular}[c]{@{}c@{}}MMLU\\STEM\end{tabular}} &
\textbf{IFEval} & \\
\midrule
Base Model & 0 & - & - & 21.9 & 2.5 & 45.0 & 12.5 & 16.7 & 39.3 & 17.9 & 22.3 \\

\hdashline
GRPO & 10 & 65.8 & 1.00$\times$ & 27.8 & 4.3 & 54.0 & 18.8 & 22.8 & 43.9 & 20.3 & 27.4 \\
\specrow \hspace{2mm}$\hookrightarrow$~+~\ourmethod\ $\ell=1$ & 10 & 63.5 & 1.03$\times$ & 28.6 & 5.0 & 55.2 & 20.2 & 21.2 & 44.5 & 18.9 & 27.7 \\

GRPO & 20 & 127.4 & 1.00$\times$ & 31.2 & 6.1 & 57.8 & 25.4 & 24.7 & 45.5 & 20.3 & 30.1 \\
\specrow \hspace{2mm}$\hookrightarrow$~+~\ourmethod\ $\ell=1$ & 20 & 118.3 & 1.05$\times$ & 34.7 & 5.2 & 61.6 & 25.4 & 22.4 & 45.1 & 22.9 & 31.0 \\

GRPO & 30 & 187.7 & 1.00$\times$ & 33.8 & 5.9 & 60.4 & 22.4 & 25.0 & 47.9 & 20.9 & 30.9 \\
\specrow \hspace{2mm}$\hookrightarrow$~+~\ourmethod\ $\ell=1$ & 30 & 168.0 & 1.07$\times$ & 34.1 & 6.1 & 61.2 & 25.0 & 25.0 & 49.9 & 23.7 & 32.1 \\

GRPO & 40 & 248.3 & 1.00$\times$ & 36.4 & 6.6 & 65.0 & 23.5 & 24.7 & 50.7 & 21.3 & 32.6 \\
\specrow \hspace{2mm}$\hookrightarrow$~+~\ourmethod\ $\ell=1$ & 40 & 213.5 & 1.11$\times$ & 36.2 & 6.5 & 63.0 & 27.2 & 24.9 & 51.7 & 22.4 & 33.1 \\

GRPO & 50 & 309.1 & 1.00$\times$ & 36.6 & 5.9 & 64.2 & 25.7 & 24.6 & 53.7 & 25.1 & 33.7 \\
\specrow \hspace{2mm}$\hookrightarrow$~+~\ourmethod\ $\ell=1$ & 50 & 257.0 & 1.14$\times$ & 35.0 & 6.9 & 63.2 & 29.0 & 24.7 & 54.0 & 24.4 & 33.9 \\

GRPO & 60 & 370.4 & 1.00$\times$ & 38.0 & 6.7 & 63.6 & 28.3 & 26.7 & 56.0 & 24.0 & 34.8 \\
\specrow \hspace{2mm}$\hookrightarrow$~+~\ourmethod\ $\ell=1$ & 60 & 298.5 & 1.17$\times$ & 36.6 & 7.5 & 64.2 & 26.8 & 26.7 & 55.6 & 23.7 & 34.4 \\

GRPO & 70 & 431.9 & 1.00$\times$ & 35.3 & 8.0 & 61.8 & 26.5 & 28.1 & 55.5 & 26.2 & 34.5 \\
\specrow \hspace{2mm}$\hookrightarrow$~+~\ourmethod\ $\ell=1$ & 70 & 339.2 & 1.19$\times$ & 36.4 & 6.6 & 62.6 & 28.7 & 28.0 & 58.0 & 26.8 & 35.3 \\

GRPO & 80 & 493.5 & 1.00$\times$ & 36.1 & 7.2 & 64.4 & 24.3 & 26.4 & 59.4 & 25.0 & 34.7 \\
\specrow \hspace{2mm}$\hookrightarrow$~+~\ourmethod\ $\ell=1$ & 80 & 379.7 & 1.20$\times$ & 37.0 & 7.4 & 67.4 & 22.8 & 28.0 & 60.2 & 23.7 & 35.2 \\

GRPO & 90 & 554.8 & 1.00$\times$ & 35.0 & 7.8 & 64.4 & 26.5 & 25.5 & 60.7 & 24.4 & 34.9 \\
\specrow \hspace{2mm}$\hookrightarrow$~+~\ourmethod\ $\ell=1$ & 90 & 419.1 & 1.22$\times$ & 37.8 & 8.0 & 63.8 & 28.7 & 26.5 & 59.6 & 25.9 & 35.8 \\

\bottomrule
\end{tabular}
}
\caption{Intermediate training results of Qwen-3-1.7B-Base on DeepMath-6K with GRPO and \ourmethod\ at lenience $\ell=1$. We report rollout efficiency and accuracy every 10 training steps, illustrating the progression of model performance during training.}
\label{tab:grpo-1.7b-details-l1}
\end{table*}

\begin{table*}[!t]
\centering
\resizebox{\textwidth}{!}{%
\begin{tabular}{l r r r r r r r r r r r}
\toprule
\multirow{3}{*}{\textbf{Algorithm}} &
\multirow{3}{*}{\textbf{Step}} &
\multicolumn{2}{c}{\textbf{Rollout Efficiency}} &
\multicolumn{5}{c}{\textbf{Math Reasoning}} &
\multicolumn{2}{c}{\textbf{OOD}} &
\multirow{3}{*}{\textbf{AVG}} \\
\cmidrule(lr){3-4}\cmidrule(lr){5-9}\cmidrule(lr){10-11}
& & \textbf{Tokens (M)} & \textbf{Speedup} &
\textbf{AMC23} & \textbf{AIME24} &
\textbf{\begin{tabular}[c]{@{}c@{}}MATH\\500\end{tabular}} &
\textbf{\begin{tabular}[c]{@{}c@{}}Minerva\\Math\end{tabular}} &
\textbf{\begin{tabular}[c]{@{}c@{}}Olympiad\\Bench\end{tabular}} &
\textbf{\begin{tabular}[c]{@{}c@{}}MMLU\\STEM\end{tabular}} &
\textbf{IFEval} & \\
\midrule
Base Model & 0 & - & - & 21.9 & 2.5 & 45.0 & 12.5 & 16.7 & 39.3 & 17.9 & 22.3 \\

\hdashline
GRPO & 10 & 65.8 & 1.00$\times$ & 27.8 & 4.3 & 54.0 & 18.8 & 22.8 & 43.9 & 20.3 & 27.4 \\
\specrow \hspace{2mm}$\hookrightarrow$~+~\ourmethod\ $\ell=e^{0.2}$ 
& 10 & 53.6 & 1.19$\times$ & 30.5 & 4.0 & 55.6 & 19.1 & 22.1 & 42.4 & 20.0 & 27.7 \\

GRPO & 20 & 127.4 & 1.00$\times$ & 31.2 & 6.1 & 57.8 & 25.4 & 24.7 & 45.5 & 20.3 & 30.1 \\
\specrow \hspace{2mm}$\hookrightarrow$~+~\ourmethod\ $\ell=e^{0.2}$ 
& 20 & 92.7 & 1.29$\times$ & 33.3 & 5.2 & 60.6 & 26.5 & 25.5 & 46.2 & 20.0 & 31.0 \\

GRPO & 30 & 187.7 & 1.00$\times$ & 33.8 & 5.9 & 60.4 & 22.4 & 25.0 & 47.9 & 20.9 & 30.9 \\
\specrow \hspace{2mm}$\hookrightarrow$~+~\ourmethod\ $\ell=e^{0.2}$ 
& 30 & 120.5 & 1.42$\times$ & 34.8 & 6.2 & 60.8 & 26.8 & 27.0 & 49.5 & 22.9 & 32.6 \\

GRPO & 40 & 248.3 & 1.00$\times$ & 36.4 & 6.6 & 65.0 & 23.5 & 24.7 & 50.7 & 21.3 & 32.6 \\
\specrow \hspace{2mm}$\hookrightarrow$~+~\ourmethod\ $\ell=e^{0.2}$ 
& 40 & 143.6 & 1.54$\times$ & 34.2 & 6.7 & 61.8 & 28.3 & 26.4 & 51.8 & 22.2 & 33.1 \\

GRPO & 50 & 309.1 & 1.00$\times$ & 36.6 & 5.9 & 64.2 & 25.7 & 24.6 & 53.7 & 25.1 & 33.7 \\
\specrow \hspace{2mm}$\hookrightarrow$~+~\ourmethod\ $\ell=e^{0.2}$ 
& 50 & 166.1 & 1.62$\times$ & 33.4 & 8.5 & 65.2 & 27.6 & 25.9 & 54.4 & 25.5 & 34.4 \\

GRPO & 60 & 370.4 & 1.00$\times$ & 38.0 & 6.7 & 63.6 & 28.3 & 26.7 & 56.0 & 24.0 & 34.8 \\
\specrow \hspace{2mm}$\hookrightarrow$~+~\ourmethod\ $\ell=e^{0.2}$ 
& 60 & 186.1 & 1.70$\times$ & 35.9 & 7.5 & 63.6 & 29.4 & 25.5 & 55.0 & 24.4 & 34.5 \\

GRPO & 70 & 431.9 & 1.00$\times$ & 35.3 & 8.0 & 61.8 & 26.5 & 28.1 & 55.5 & 26.2 & 34.5 \\
\specrow \hspace{2mm}$\hookrightarrow$~+~\ourmethod\ $\ell=e^{0.2}$ 
& 70 & 206.5 & 1.77$\times$ & 34.1 & 7.3 & 64.6 & 27.2 & 29.5 & 58.4 & 25.1 & 35.2 \\

GRPO & 80 & 493.5 & 1.00$\times$ & 36.1 & 7.2 & 64.4 & 24.3 & 26.4 & 59.4 & 25.0 & 34.7 \\
\specrow \hspace{2mm}$\hookrightarrow$~+~\ourmethod\ $\ell=e^{0.2}$ 
& 80 & 226.3 & 1.82$\times$ & 37.3 & 5.9 & 63.4 & 29.4 & 29.9 & 57.9 & 24.0 & 35.4 \\

GRPO & 90 & 554.8 & 1.00$\times$ & 35.0 & 7.8 & 64.4 & 26.5 & 25.5 & 60.7 & 24.4 & 34.9 \\
\specrow \hspace{2mm}$\hookrightarrow$~+~\ourmethod\ $\ell=e^{0.2}$ 
& 90 & 246.7 & 1.86$\times$ & 35.8 & 7.8 & 66.4 & 29.8 & 29.6 & 58.5 & 25.9 & 36.3 \\

\bottomrule
\end{tabular}
}
\caption{Intermediate training results of Qwen-3-1.7B-Base on DeepMath-6K with GRPO and \ourmethod\ at lenience $\ell=e^{0.2}$. We report rollout efficiency and accuracy every 10 training steps, illustrating the progression of model performance during training.}
\label{tab:grpo-1.7b-details-l02}
\end{table*}

\begin{table*}[!t]
\centering
\resizebox{\textwidth}{!}{%
\begin{tabular}{l r r r r r r r r r r r}
\toprule
\multirow{3}{*}{\textbf{Algorithm}} &
\multirow{3}{*}{\textbf{Step}} &
\multicolumn{2}{c}{\textbf{Rollout Efficiency}} &
\multicolumn{5}{c}{\textbf{Math Reasoning}} &
\multicolumn{2}{c}{\textbf{OOD}} &
\multirow{3}{*}{\textbf{AVG}} \\
\cmidrule(lr){3-4}\cmidrule(lr){5-9}\cmidrule(lr){10-11}
& & \textbf{Tokens (M)} & \textbf{Speedup} &
\textbf{AMC23} & \textbf{AIME24} &
\textbf{\begin{tabular}[c]{@{}c@{}}MATH\\ 500\end{tabular}} &
\textbf{\begin{tabular}[c]{@{}c@{}}Minerva\\ Math\end{tabular}} &
\textbf{\begin{tabular}[c]{@{}c@{}}Olympiad\\ Bench\end{tabular}} &
\textbf{\begin{tabular}[c]{@{}c@{}}MMLU\\ STEM\end{tabular}} &
\textbf{IFEval} & \\
\midrule
Base Model & 0 & - & - & 21.9 & 2.5 & 45.0 & 12.5 & 16.7 & 39.3 & 17.9 & 22.3 \\

\hdashline
GRPO & 10 & 65.8  & 1.00$\times$ & 27.8 & 4.3 & 54.0 & 18.8 & 22.8 & 43.9 & 20.3 & 27.4 \\
\specrow \hspace{2mm}$\hookrightarrow$~+~\ourmethod\ $\ell=e^{0.5}$ & 10 & 43.6  & 1.41$\times$ & 32.5 & 5.7 & 55.8 & 21.7 & 21.8 & 43.0 & 22.4 & 29.0 \\

GRPO & 20 & 127.4 & 1.00$\times$ & 31.2 & 6.1 & 57.8 & 25.4 & 24.7 & 45.5 & 20.3 & 30.1 \\
\specrow \hspace{2mm}$\hookrightarrow$~+~\ourmethod\ $\ell=e^{0.5}$ & 20 & 67.2  & 1.66$\times$ & 34.2 & 5.9 & 63.0 & 25.0 & 24.6 & 46.9 & 22.2 & 31.7 \\

GRPO & 30 & 187.7 & 1.00$\times$ & 33.8 & 5.9 & 60.4 & 22.4 & 25.0 & 47.9 & 20.9 & 30.9 \\
\specrow \hspace{2mm}$\hookrightarrow$~+~\ourmethod\ $\ell=e^{0.5}$ & 30 & 85.1  & 1.85$\times$ & 34.5 & 7.2 & 64.0 & 25.4 & 27.6 & 51.1 & 26.2 & 33.7 \\

GRPO & 40 & 248.3 & 1.00$\times$ & 36.4 & 6.6 & 65.0 & 23.5 & 24.7 & 50.7 & 21.3 & 32.6 \\
\specrow \hspace{2mm}$\hookrightarrow$~+~\ourmethod\ $\ell=e^{0.5}$ & 40 & 102.9 & 1.96$\times$ & 37.0 & 7.4 & 63.8 & 26.5 & 26.1 & 52.2 & 23.5 & 33.8 \\

GRPO & 50 & 309.1 & 1.00$\times$ & 36.6 & 5.9 & 64.2 & 25.7 & 24.6 & 53.7 & 25.1 & 33.7 \\
\specrow \hspace{2mm}$\hookrightarrow$~+~\ourmethod\ $\ell=e^{0.5}$ & 50 & 119.4 & 2.06$\times$ & 33.1 & 7.3 & 64.4 & 28.7 & 28.0 & 55.6 & 27.7 & 35.0 \\

GRPO & 60 & 370.4 & 1.00$\times$ & 38.0 & 6.7 & 63.6 & 28.3 & 26.7 & 56.0 & 24.0 & 34.8 \\
\specrow \hspace{2mm}$\hookrightarrow$~+~\ourmethod\ $\ell=e^{0.5}$ & 60 & 135.1 & 2.14$\times$ & 36.6 & 7.0 & 66.4 & 26.5 & 29.9 & 54.7 & 28.8 & 35.7 \\

GRPO & 70 & 431.9 & 1.00$\times$ & 35.3 & 8.0 & 61.8 & 26.5 & 28.1 & 55.5 & 26.2 & 34.5 \\
\specrow \hspace{2mm}$\hookrightarrow$~+~\ourmethod\ $\ell=e^{0.5}$ & 70 & 153.2 & 2.18$\times$ & 37.0 & 8.1 & 65.4 & 26.5 & 29.9 & 55.6 & 27.4 & 35.7 \\

GRPO & 80 & 493.5 & 1.00$\times$ & 36.1 & 7.2 & 64.4 & 24.3 & 26.4 & 59.4 & 25.0 & 34.7 \\
\specrow \hspace{2mm}$\hookrightarrow$~+~\ourmethod\ $\ell=e^{0.5}$ & 80 & 168.1 & 2.24$\times$ & 35.0 & 10.1 & 67.0 & 29.8 & 29.6 & 57.1 & 28.3 & 36.7 \\

GRPO & 90 & 554.8 & 1.00$\times$ & 35.0 & 7.8 & 64.4 & 26.5 & 25.5 & 60.7 & 24.4 & 34.9 \\
\specrow \hspace{2mm}$\hookrightarrow$~+~\ourmethod\ $\ell=e^{0.5}$ & 90 & 182.7 & 2.29$\times$ & 38.3 & 9.1 & 68.0 & 29.4 & 29.3 & 58.3 & 28.8 & 37.3 \\
\bottomrule
\end{tabular}
}
\caption{Intermediate training results of Qwen-3-1.7B-Base on DeepMath-6K with GRPO and \ourmethod\ at lenience $\ell=e^{0.5}$.
We report rollout efficiency and accuracy every 10 training steps, illustrating the progression of model performance during training.}
\label{tab:grpo-1.7b-details-l05}
\end{table*}

\begin{table*}[!t]
\centering
\resizebox{\textwidth}{!}{%
\begin{tabular}{l r r r r r r r r r r r}
\toprule
\multirow{3}{*}{\textbf{Algorithm}} &
\multirow{3}{*}{\textbf{Step}} &
\multicolumn{2}{c}{\textbf{Rollout Efficiency}} &
\multicolumn{5}{c}{\textbf{Math Reasoning}} &
\multicolumn{2}{c}{\textbf{OOD}} &
\multirow{3}{*}{\textbf{AVG}} \\
\cmidrule(lr){3-4}\cmidrule(lr){5-9}\cmidrule(lr){10-11}
& & \textbf{Tokens (M)} & \textbf{Speedup} &
\textbf{AMC23} & \textbf{AIME24} &
\textbf{\begin{tabular}[c]{@{}c@{}}MATH\\500\end{tabular}} &
\textbf{\begin{tabular}[c]{@{}c@{}}Minerva\\Math\end{tabular}} &
\textbf{\begin{tabular}[c]{@{}c@{}}Olympiad\\Bench\end{tabular}} &
\textbf{\begin{tabular}[c]{@{}c@{}}MMLU\\STEM\end{tabular}} &
\textbf{IFEval} & \\
\midrule
Base Model & 0 & - & - & 21.9 & 2.5 & 45.0 & 12.5 & 16.7 & 39.3 & 17.9 & 22.3 \\

\hdashline
GRPO & 10 & 65.8  & 1.00$\times$ & 27.8 & 4.3 & 54.0 & 18.8 & 22.8 & 43.9 & 20.3 & 27.4 \\
\specrow \hspace{2mm}$\hookrightarrow$~+~\ourmethod\ $\ell=e^{0.8}$ 
& 10 & 41.9 & 1.46$\times$ & 31.7 & 3.8 & 57.0 & 19.9 & 21.6 & 44.8 & 20.1 & 28.4 \\

GRPO & 20 & 127.4 & 1.00$\times$ & 31.2 & 6.1 & 57.8 & 25.4 & 24.7 & 45.5 & 20.3 & 30.1 \\
\specrow \hspace{2mm}$\hookrightarrow$~+~\ourmethod\ $\ell=e^{0.8}$ 
& 20 & 57.5 & 1.86$\times$ & 35.5 & 5.3 & 60.0 & 26.1 & 24.4 & 46.2 & 23.3 & 31.5 \\

GRPO & 30 & 187.7 & 1.00$\times$ & 33.8 & 5.9 & 60.4 & 22.4 & 25.0 & 47.9 & 20.9 & 30.9 \\
\specrow \hspace{2mm}$\hookrightarrow$~+~\ourmethod\ $\ell=e^{0.8}$ 
& 30 & 70.1 & 2.12$\times$ & 33.4 & 7.6 & 62.4 & 28.7 & 24.0 & 51.5 & 27.4 & 33.6 \\

GRPO & 40 & 248.3 & 1.00$\times$ & 36.4 & 6.6 & 65.0 & 23.5 & 24.7 & 50.7 & 21.3 & 32.6 \\
\specrow \hspace{2mm}$\hookrightarrow$~+~\ourmethod\ $\ell=e^{0.8}$ 
& 40 & 81.8 & 2.29$\times$ & 37.7 & 7.4 & 63.8 & 26.8 & 27.7 & 53.8 & 28.5 & 35.1 \\

GRPO & 50 & 309.1 & 1.00$\times$ & 36.6 & 5.9 & 64.2 & 25.7 & 24.6 & 53.7 & 25.1 & 33.7 \\
\specrow \hspace{2mm}$\hookrightarrow$~+~\ourmethod\ $\ell=e^{0.8}$ 
& 50 & 97.5 & 2.35$\times$ & 35.8 & 6.8 & 63.4 & 28.3 & 26.5 & 57.0 & 28.7 & 35.2 \\

GRPO & 60 & 370.4 & 1.00$\times$ & 38.0 & 6.7 & 63.6 & 28.3 & 26.7 & 56.0 & 24.0 & 34.8 \\
\specrow \hspace{2mm}$\hookrightarrow$~+~\ourmethod\ $\ell=e^{0.8}$ 
& 60 & 110.8 & 2.43$\times$ & 36.2 & 6.7 & 61.8 & 26.8 & 25.8 & 57.7 & 25.5 & 34.4 \\

GRPO & 70 & 431.9 & 1.00$\times$ & 35.3 & 8.0 & 61.8 & 26.5 & 28.1 & 55.5 & 26.2 & 34.5 \\
\specrow \hspace{2mm}$\hookrightarrow$~+~\ourmethod\ $\ell=e^{0.8}$ 
& 70 & 120.0 & 2.54$\times$ & 37.2 & 5.4 & 62.2 & 26.8 & 25.9 & 58.9 & 27.9 & 34.9 \\

GRPO & 80 & 493.5 & 1.00$\times$ & 36.1 & 7.2 & 64.4 & 24.3 & 26.4 & 59.4 & 25.0 & 34.7 \\
\specrow \hspace{2mm}$\hookrightarrow$~+~\ourmethod\ $\ell=e^{0.8}$ 
& 80 & 132.0 & 2.60$\times$ & 25.0 & 4.9 & 63.6 & 27.6 & 26.5 & 58.4 & 28.3 & 33.5 \\

GRPO & 90 & 554.8 & 1.00$\times$ & 35.0 & 7.8 & 64.4 & 26.5 & 25.5 & 60.7 & 24.4 & 34.9 \\
\specrow \hspace{2mm}$\hookrightarrow$~+~\ourmethod\ $\ell=e^{0.8}$ 
& 90 & 144.8 & 2.64$\times$ & 28.6 & 5.0 & 63.6 & 27.2 & 25.0 & 61.7 & 26.2 & 33.9 \\

\bottomrule
\end{tabular}
}
\caption{Intermediate training results of Qwen-3-1.7B-Base on DeepMath-6K with GRPO and \ourmethod\ at lenience $\ell=e^{0.8}$.
We report rollout efficiency and accuracy every 10 training steps, illustrating the progression of model performance during training.}
\label{tab:grpo-1.7b-details-l08}
\end{table*}

\begin{table*}[!t]
\centering
\resizebox{\textwidth}{!}{%
\begin{tabular}{l r r r r r r r r r r r}
\toprule
\multirow{3}{*}{\textbf{Algorithm}} &
\multirow{3}{*}{\textbf{Step}} &
\multicolumn{2}{c}{\textbf{Rollout Efficiency}} &
\multicolumn{5}{c}{\textbf{Math Reasoning}} &
\multicolumn{2}{c}{\textbf{OOD}} &
\multirow{3}{*}{\textbf{AVG}} \\
\cmidrule(lr){3-4}\cmidrule(lr){5-9}\cmidrule(lr){10-11}
& & \textbf{Tokens (M)} & \textbf{Speedup} &
\textbf{AMC23} & \textbf{AIME24} &
\textbf{\begin{tabular}[c]{@{}c@{}}MATH\\ 500\end{tabular}} &
\textbf{\begin{tabular}[c]{@{}c@{}}Minerva\\ Math\end{tabular}} &
\textbf{\begin{tabular}[c]{@{}c@{}}Olympiad\\ Bench\end{tabular}} &
\textbf{\begin{tabular}[c]{@{}c@{}}MMLU\\ STEM\end{tabular}} &
\textbf{IFEval} & \\
\midrule
Base Model & 0 & - & - & 21.9 & 2.5 & 45.0 & 12.5 & 16.7 & 39.3 & 17.9 & 22.3 \\

\hdashline
GRPO & 10 & 65.8 & 1.00$\times$ & 27.8 & 4.3 & 54.0 & 18.8 & 22.8 & 43.9 & 20.3 & 27.4 \\
\specrow \hspace{2mm}$\hookrightarrow$~+~\ourmethod\ $\ell=e^{2.0}$ & 10 & 40.4 & 1.50$\times$ & 32.8 & 4.3 & 55.4 & 21.7 & 21.8 & 44.0 & 19.6 & 28.5 \\

GRPO & 20 & 127.4 & 1.00$\times$ & 31.2 & 6.1 & 57.8 & 25.4 & 24.7 & 45.5 & 20.3 & 30.1 \\
\specrow \hspace{2mm}$\hookrightarrow$~+~\ourmethod\ $\ell=e^{2.0}$ & 20 & 43.9 & 2.24$\times$ & 34.8 & 4.8 & 60.4 & 28.3 & 23.7 & 44.3 & 25.0 & 31.6 \\

GRPO & 30 & 187.7 & 1.00$\times$ & 33.8 & 5.9 & 60.4 & 22.4 & 25.0 & 47.9 & 20.9 & 30.9 \\
\specrow \hspace{2mm}$\hookrightarrow$~+~\ourmethod\ $\ell=e^{2.0}$ & 30 & 51.2 & 2.59$\times$ & 36.2 & 7.3 & 61.0 & 26.1 & 25.6 & 49.8 & 29.2 & 33.6 \\

GRPO & 40 & 248.3 & 1.00$\times$ & 36.4 & 6.6 & 65.0 & 23.5 & 24.7 & 50.7 & 21.3 & 32.6 \\
\specrow \hspace{2mm}$\hookrightarrow$~+~\ourmethod\ $\ell=e^{2.0}$ & 40 & 54.5 & 2.93$\times$ & 35.9 & 7.4 & 64.0 & 28.7 & 26.4 & 52.5 & 31.4 & 35.2 \\

GRPO & 50 & 309.1 & 1.00$\times$ & 36.6 & 5.9 & 64.2 & 25.7 & 24.6 & 53.7 & 25.1 & 33.7 \\
\specrow \hspace{2mm}$\hookrightarrow$~+~\ourmethod\ $\ell=e^{2.0}$ & 50 & 60.1 & 3.14$\times$ & 26.6 & 4.0 & 61.6 & 25.7 & 24.1 & 52.4 & 29.2 & 31.9 \\

GRPO & 60 & 370.4 & 1.00$\times$ & 38.0 & 6.7 & 63.6 & 28.3 & 26.7 & 56.0 & 24.0 & 34.8 \\
\specrow \hspace{2mm}$\hookrightarrow$~+~\ourmethod\ $\ell=e^{2.0}$ & 60 & 67.2 & 3.24$\times$ & 29.4 & 6.4 & 57.8 & 25.0 & 24.3 & 54.6 & 31.4 & 32.7 \\

GRPO & 70 & 431.9 & 1.00$\times$ & 35.3 & 8.0 & 61.8 & 26.5 & 28.1 & 55.5 & 26.2 & 34.5 \\
\specrow \hspace{2mm}$\hookrightarrow$~+~\ourmethod\ $\ell=e^{2.0}$ & 70 & 76.8 & 3.28$\times$ & 29.5 & 6.0 & 57.4 & 21.7 & 23.7 & 52.5 & 28.7 & 31.4 \\

GRPO & 80 & 493.5 & 1.00$\times$ & 36.1 & 7.2 & 64.4 & 24.3 & 26.4 & 59.4 & 25.0 & 34.7 \\
\specrow \hspace{2mm}$\hookrightarrow$~+~\ourmethod\ $\ell=e^{2.0}$ & 80 & 90.5 & 3.24$\times$ & 27.8 & 4.9 & 57.4 & 19.5 & 23.4 & 54.8 & 30.3 & 31.2 \\

GRPO & 90 & 554.8 & 1.00$\times$ & 35.0 & 7.8 & 64.4 & 26.5 & 25.5 & 60.7 & 24.4 & 34.9 \\
\specrow \hspace{2mm}$\hookrightarrow$~+~\ourmethod\ $\ell=e^{2.0}$ & 90 & 114.4 & 3.05$\times$ & 26.2 & 5.2 & 55.0 & 21.0 & 21.9 & 53.5 & 29.0 & 30.3 \\

\bottomrule
\end{tabular}
}
\caption{Intermediate training results of Qwen-3-1.7B-Base on DeepMath-6K with GRPO and \ourmethod\ at lenience $\ell=e^{2.0}$.
We report rollout efficiency and accuracy every 10 training steps, illustrating the progression of model performance during training.}
\label{tab:grpo-1.7b-details-l20}
\end{table*}

\begin{table*}[!t]
\centering
\resizebox{\textwidth}{!}{%
\begin{tabular}{l r r r r r r r r r r r}
\toprule
\multirow{3}{*}{\textbf{Algorithm}} &
\multirow{3}{*}{\textbf{Step}} &
\multicolumn{2}{c}{\textbf{Rollout Efficiency}} &
\multicolumn{5}{c}{\textbf{Math Reasoning}} &
\multicolumn{2}{c}{\textbf{OOD}} &
\multirow{3}{*}{\textbf{AVG}} \\
\cmidrule(lr){3-4}\cmidrule(lr){5-9}\cmidrule(lr){10-11}
& & \textbf{Tokens (M)} & \textbf{Speedup} &
\textbf{AMC23} & \textbf{AIME24} &
\textbf{\begin{tabular}[c]{@{}c@{}}MATH\\ 500\end{tabular}} &
\textbf{\begin{tabular}[c]{@{}c@{}}Minerva\\ Math\end{tabular}} &
\textbf{\begin{tabular}[c]{@{}c@{}}Olympiad\\ Bench\end{tabular}} &
\textbf{\begin{tabular}[c]{@{}c@{}}MMLU\\ STEM\end{tabular}} &
\textbf{IFEval} & \\
\midrule
Base Model & 0 & - & - & 21.9 & 2.5 & 45.0 & 12.5 & 16.7 & 39.3 & 17.9 & 22.3 \\

\hdashline
GRPO & 10 & 65.8 & 1.00$\times$ & 27.8 & 4.3 & 54.0 & 18.8 & 22.8 & 43.9 & 20.3 & 27.4 \\
\specrow \hspace{2mm}$\hookrightarrow$~+~\ourmethod\ $\ell=\infty$ & 10 & 40.0 & 1.75$\times$ & 29.2 & 3.5 & 53.8 & 16.9 & 21.6 & 42.5 & 20.0 & 26.8 \\

GRPO & 20 & 127.4 & 1.00$\times$ & 31.2 & 6.1 & 57.8 & 25.4 & 24.7 & 45.5 & 20.3 & 30.1 \\
\specrow \hspace{2mm}$\hookrightarrow$~+~\ourmethod\ $\ell=\infty$ & 20 & 40.0 & 3.39$\times$ & 31.6 & 4.8 & 56.4 & 21.0 & 22.7 & 44.4 & 21.4 & 28.9 \\

GRPO & 30 & 187.7 & 1.00$\times$ & 33.8 & 5.9 & 60.4 & 22.4 & 25.0 & 47.9 & 20.9 & 30.9 \\
\specrow \hspace{2mm}$\hookrightarrow$~+~\ourmethod\ $\ell=\infty$ & 30 & 40.0 & 5.03$\times$ & 30.2 & 3.8 & 53.2 & 21.7 & 21.3 & 43.0 & 21.4 & 27.8 \\

GRPO & 40 & 248.3 & 1.00$\times$ & 36.4 & 6.6 & 65.0 & 23.5 & 24.7 & 50.7 & 21.3 & 32.6 \\
\specrow \hspace{2mm}$\hookrightarrow$~+~\ourmethod\ $\ell=\infty$ & 40 & 40.0 & 6.65$\times$ & 33.0 & 4.9 & 57.0 & 17.3 & 22.4 & 43.3 & 22.6 & 28.6 \\

GRPO & 50 & 309.1 & 1.00$\times$ & 36.6 & 5.9 & 64.2 & 25.7 & 24.6 & 53.7 & 25.1 & 33.7 \\
\specrow \hspace{2mm}$\hookrightarrow$~+~\ourmethod\ $\ell=\infty$ & 50 & 40.0 & 8.29$\times$ & 34.1 & 4.3 & 57.0 & 23.2 & 22.5 & 43.3 & 22.7 & 29.6 \\

GRPO & 60 & 370.4 & 1.00$\times$ & 38.0 & 6.7 & 63.6 & 28.3 & 26.7 & 56.0 & 24.0 & 34.8 \\
\specrow \hspace{2mm}$\hookrightarrow$~+~\ourmethod\ $\ell=\infty$ & 60 & 40.0 & 9.93$\times$ & 31.9 & 3.8 & 58.8 & 23.5 & 22.8 & 42.6 & 23.7 & 29.6 \\

GRPO & 70 & 431.9 & 1.00$\times$ & 35.3 & 8.0 & 61.8 & 26.5 & 28.1 & 55.5 & 26.2 & 34.5 \\
\specrow \hspace{2mm}$\hookrightarrow$~+~\ourmethod\ $\ell=\infty$ & 70 & 40.0 & 11.56$\times$ & 33.1 & 4.5 & 58.6 & 23.2 & 22.4 & 44.0 & 20.9 & 29.5 \\

GRPO & 80 & 493.5 & 1.00$\times$ & 36.1 & 7.2 & 64.4 & 24.3 & 26.4 & 59.4 & 25.0 & 34.7 \\
\specrow \hspace{2mm}$\hookrightarrow$~+~\ourmethod\ $\ell=\infty$ & 80 & 40.0 & 13.21$\times$ & 31.1 & 4.0 & 60.2 & 23.9 & 24.3 & 42.8 & 23.3 & 29.9 \\

GRPO & 90 & 554.8 & 1.00$\times$ & 35.0 & 7.8 & 64.4 & 26.5 & 25.5 & 60.7 & 24.4 & 34.9 \\
\specrow \hspace{2mm}$\hookrightarrow$~+~\ourmethod\ $\ell=\infty$ & 90 & 40.0 & 14.86$\times$ & 30.2 & 4.4 & 60.4 & 19.9 & 23.7 & 44.1 & 22.0 & 29.2 \\
\bottomrule
\end{tabular}
}
\caption{Intermediate training results of Qwen-3-1.7B-Base on DeepMath-6K with GRPO and \ourmethod\ at lenience $\ell=\infty$.
We report rollout efficiency and accuracy every 10 training steps, illustrating the progression of model performance during training.}
\label{tab:grpo-1.7b-details-linf}
\end{table*}

\subsection{Generality Across Datasets}
\label{appendix:extra-datasets}
To examine whether the gains of \ourmethod\ depend on a specific training corpus, we conduct experiments on two distinct datasets: DeepMath-6K and SimpleRL-8K. 
Results in Table~\ref{tab:ablation_datasets} show that \ourmethod\ consistently improves rollout efficiency across both settings. 
For example, on Qwen-3-1.7B-Base with GRPO, rollout tokens drop from 554.8M to 182.7M on DeepMath-6K and from 639.4M to 354.0M on SimpleRL-8K. 
Accuracy remains comparable or slightly improved, confirming that the efficiency benefits of \ourmethod\ are robust to the choice of dataset.
Intermediate performance on SimpleRL-8K is reported in Table~\ref{tab:grpo-1.7b-simplerl}, while the detailed results for DeepMath-6K can be found in Table~\ref{tab:grpo-1.7b-details}.
These results suggest that the efficiency improvements of \ourmethod\ do not rely on a particular training distribution.

\clearpage
\twocolumn[\section{Full Tables and Case Studies}]

\subsection{Full Performance over Training Steps}
\label{appendix:details-steps-algo-models}
To provide a more complete view of model behavior and enhance the robustness of our method, we also report performance trajectories throughout training.

\begin{figure*}[!t]
    \centering
    \includegraphics[width=1.0\textwidth]{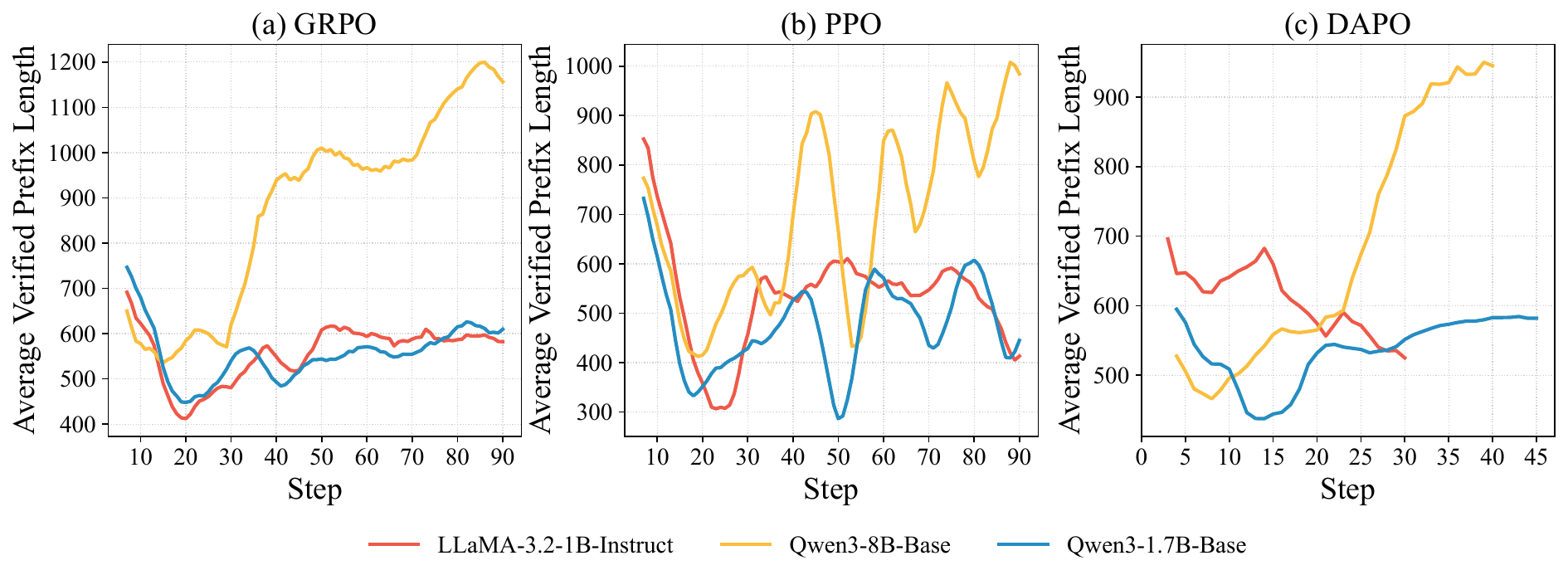}
    \caption{Average verified prefix length trajectories of \ourmethod\ under three RL algorithms: (a) GRPO, (b) PPO, and (c) DAPO. The y-axis reports the average length of the verified speculative prefix per training step, and the x-axis is the training step. colors denote model backbones: red: LLaMA-3.2-1B-Instruct, yellow: Qwen3-8B-Base, blue: Qwen3-1.7B-Base. }
    \label{fig:main_avg_prefix_len}
\end{figure*}

\begin{figure*}[!t]
    \centering
    \includegraphics[width=1.0\textwidth]{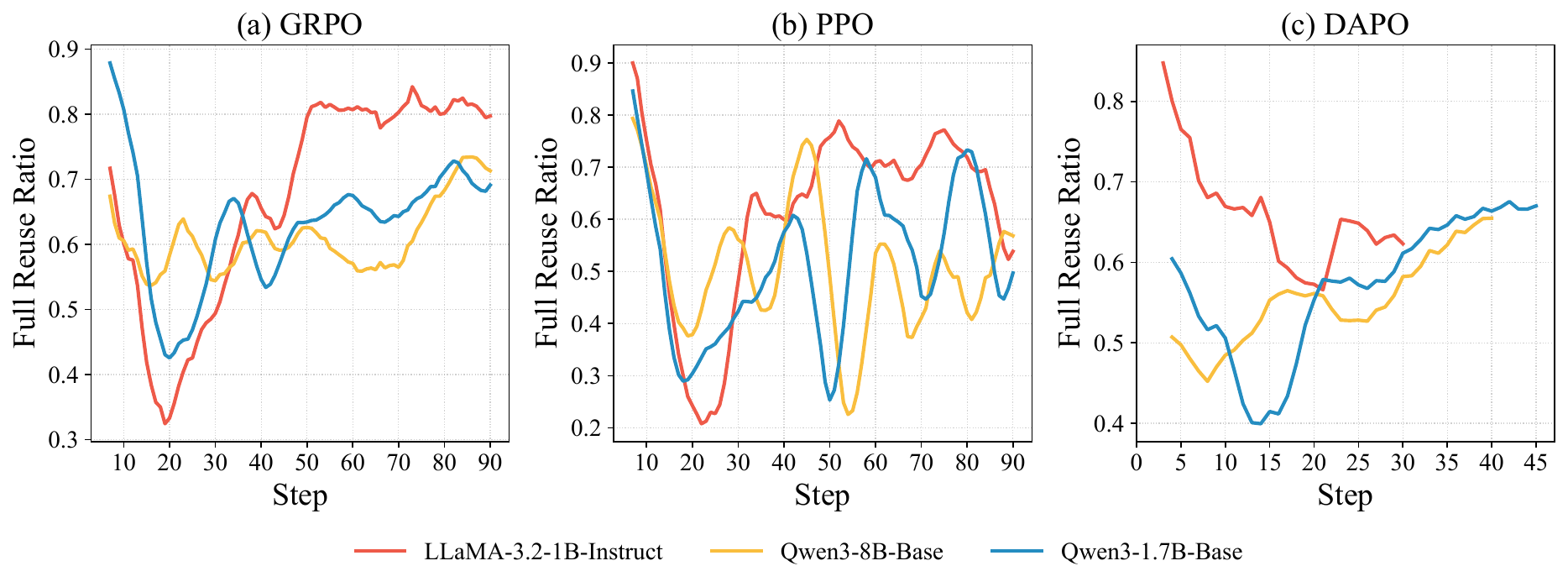}
    \caption{Full reuse ratio trajectories of \ourmethod\ under three RL algorithms: (a) GRPO, (b) PPO, and (c) DAPO. The y-axis reports the fraction of rollouts per step that are fully reused, and the x-axis the training step. colors denote model backbones: red: LLaMA-3.2-1B-Instruct, yellow: Qwen3-8B-Base, blue: Qwen3-1.7B-Base. Across settings, \ourmethod\ quickly stabilizes at a high full reuse ratio, indicating effective speculative reuse during training.}
    \label{fig:main_full_reuse_ratio}
\end{figure*}

\begin{figure*}[!t]
    \centering
    \includegraphics[width=1.0\textwidth]{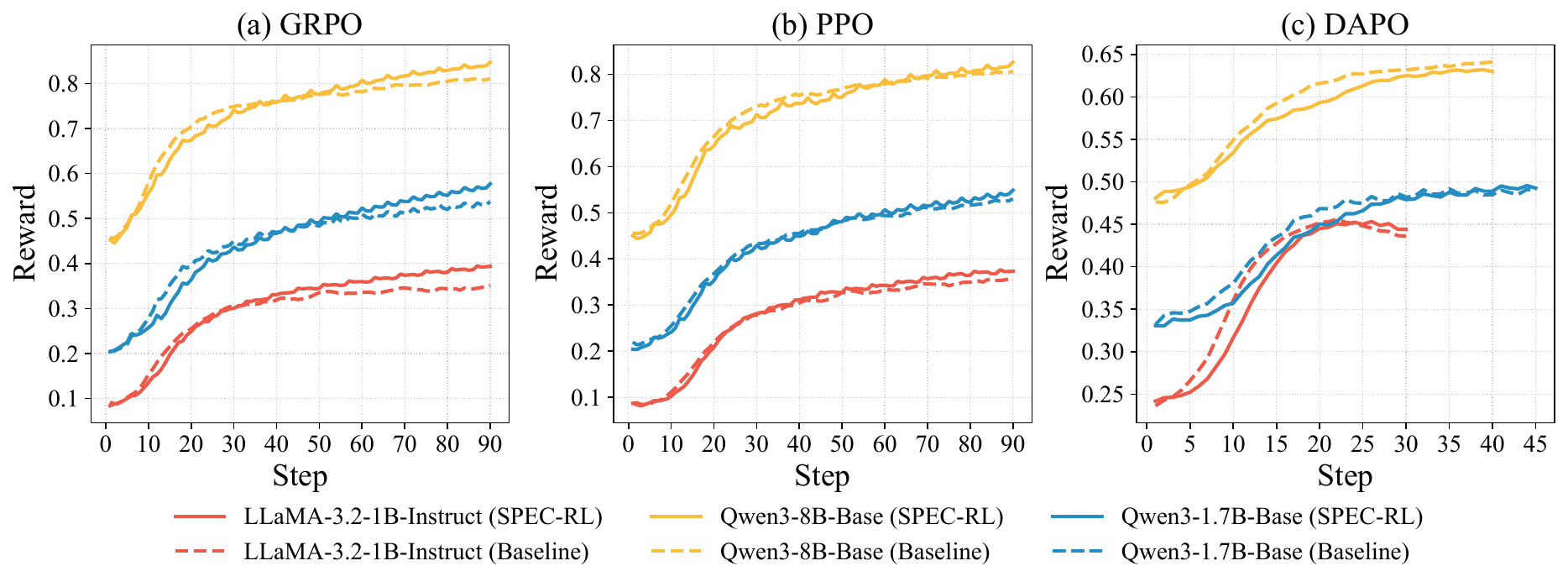}
    \caption{Training reward trajectories of \ourmethod\ versus baseline under three RL algorithms: (a) GRPO, (b) PPO, and (c) DAPO. The y-axis reports reward, and the x-axis the training step. colors denote model backbones: red: LLaMA-3.2-1B-Instruct, yellow: Qwen3-8B-Base, blue: Qwen3-1.7B-Base, while solid lines indicate \ourmethod\ and dashed lines the corresponding vanilla baselines. \ourmethod\ matches or exceeds baseline rewards under different algorithms across all backbones.}
    \label{fig:main_reward}
\end{figure*}

\begin{figure*}[!t]
    \centering
    \includegraphics[width=1.0\textwidth]{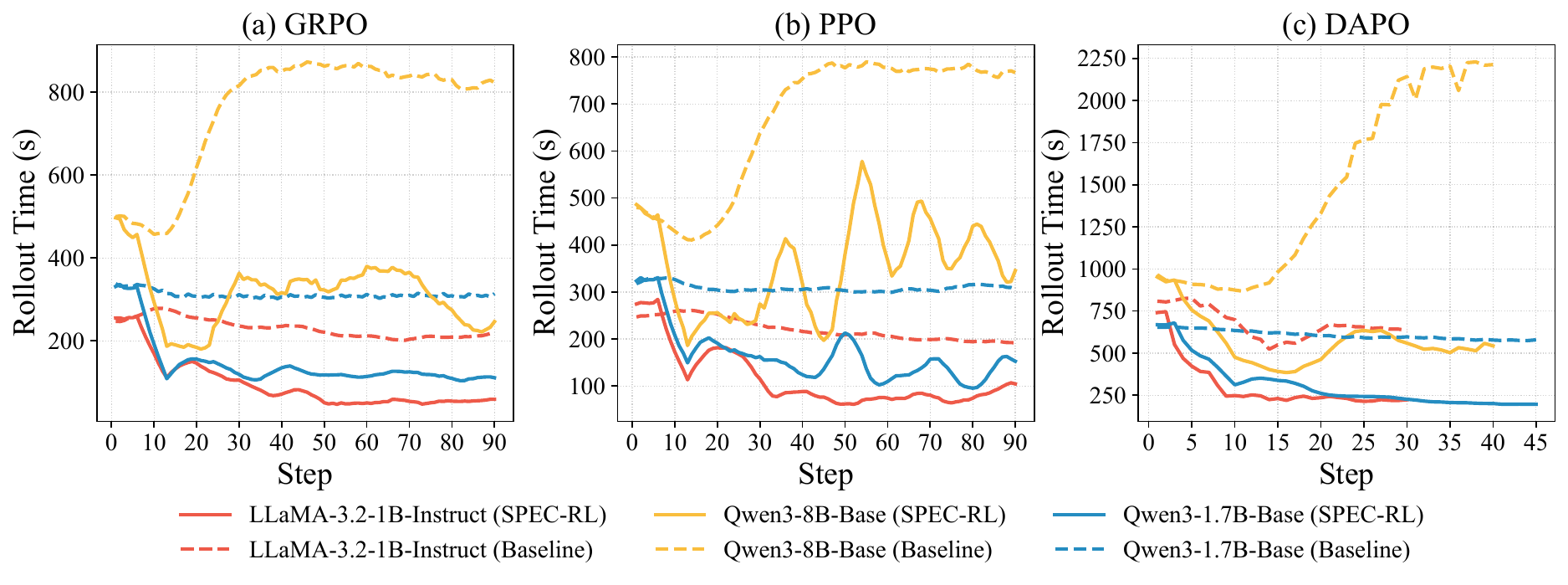}
    \caption{Training rollout time of \ourmethod\ versus baseline under three RL algorithms: (a) GRPO, (b) PPO, and (c) DAPO. The y-axis reports rollout time (seconds) and the x-axis the training step. colors denote model backbones: red: LLaMA-3.2-1B-Instruct, yellow: Qwen3-8B-Base, blue: Qwen3-1.7B-Base, while solid lines indicate \ourmethod\ and dashed lines the corresponding vanilla baselines. Across algorithms and models, \ourmethod\ yields consistently lower rollout time than the baselines.}
    \label{fig:main_rollout_time}
\end{figure*}

\paragraph{Training Dynamics and Efficiency Across Different RL Algorithms.}
We present the efficiency of our method across RL algorithms in Figures~\ref{fig:main_avg_prefix_len},~\ref{fig:main_full_reuse_ratio}, and compare rewards and rollout time against baselines in Figures~\ref{fig:main_reward},~\ref{fig:main_rollout_time}.
Across all three algorithms, SPEC-RL substantially reduces rollout time while preserving learning quality: rewards match or exceed the vanilla baselines under PPO and GRPO, and are largely on par under DAPO (with a minor late-stage gap on Qwen3-8B). The efficiency gains align with stronger speculative reuse signals: the full reuse ratio quickly rises and stabilizes around 0.6–0.85 after early transients, and the average verified prefix length remains large (hundreds to 1.2k tokens) and generally increases over training—most prominently on Qwen3-8B for GRPO/DAPO. Together, these curves indicate that SPEC-RL learns to reuse long, verified prefixes, trading decoding for reuse, which yields lower per-step generation cost without compromising reward progress.

For each setting, results are shown every 10 steps, comparing the vanilla algorithm with its \ourmethod\ variant, as shown in Tables~\ref{tab:grpo-1.7b-details}, \ref{tab:grpo-8b-details}, \ref{tab:grpo-1b-details}, \ref{tab:grpo-14b-details}, \ref{tab:ppo-1.7b-details}, \ref{tab:ppo-8b-details}, \ref{tab:ppo-1b-details}, \ref{tab:ppo-14b-details}, \ref{tab:dapo-1.7b-details}, \ref{tab:dapo-8b-details}, \ref{tab:dapo-1b-details}, \ref{tab:dapo-14b-details}.
This step-wise view complements the main results by illustrating how rollout efficiency and accuracy evolve consistently during training, rather than only at the final checkpoint.

\subsection{Case Studies}
\label{appendix:case-study}

\begin{figure*}[!t]
    \centering
    \includegraphics[width=1.0\textwidth]{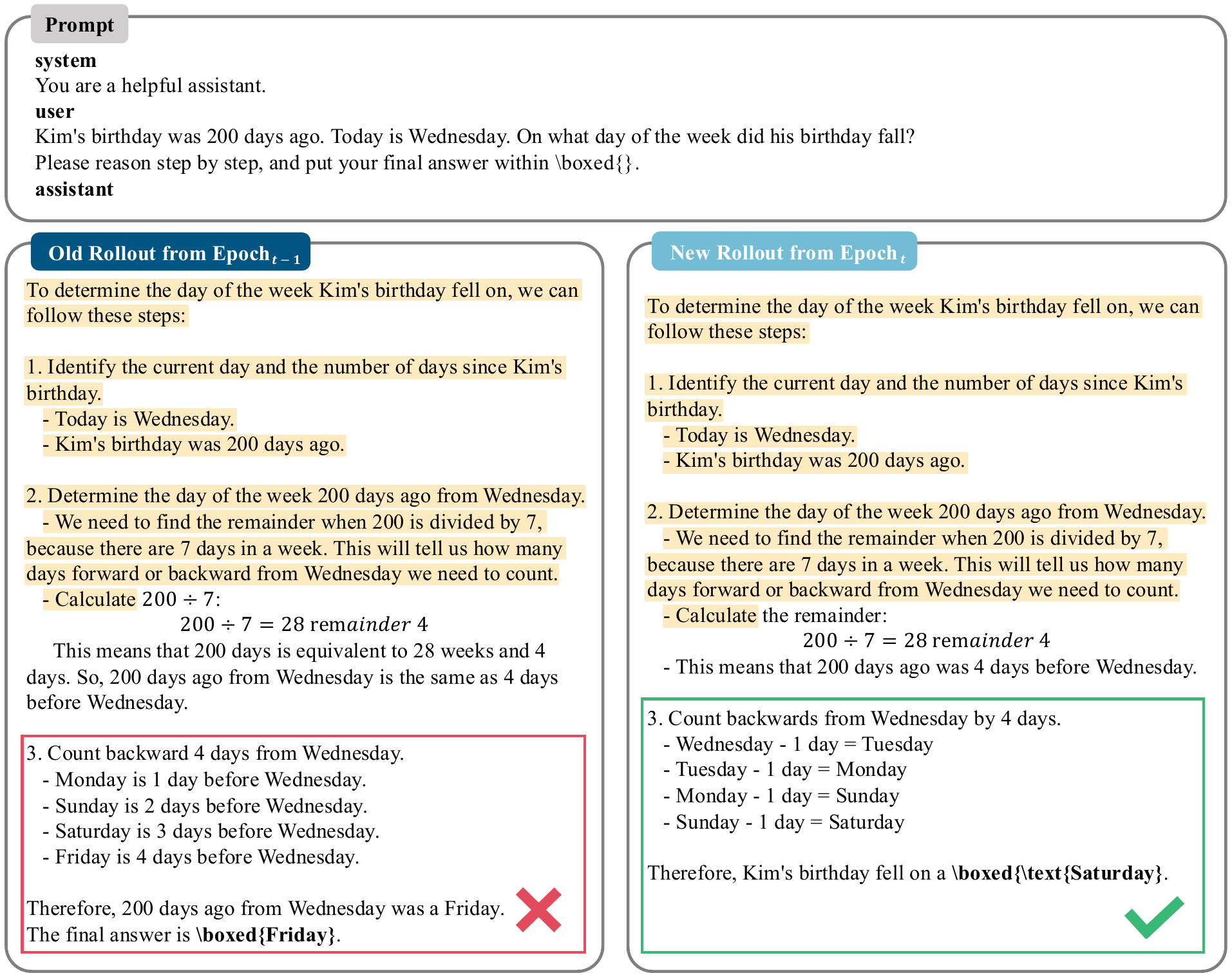}
    \caption{Case study comparing rollouts from previous and current training steps. The prompt denotes the model input. The old rollout and new rollout are generated by the respective model from corresponding epochs. Tokens highlighted in yellow indicate the verified speculative prefix. The red box marks incorrect reasoning steps, whereas the green box highlights correct reasoning steps.}
    \label{fig:case-study-spec-1}
\end{figure*}

\begin{figure*}[!t]
    \centering
    \includegraphics[width=1.0\textwidth]{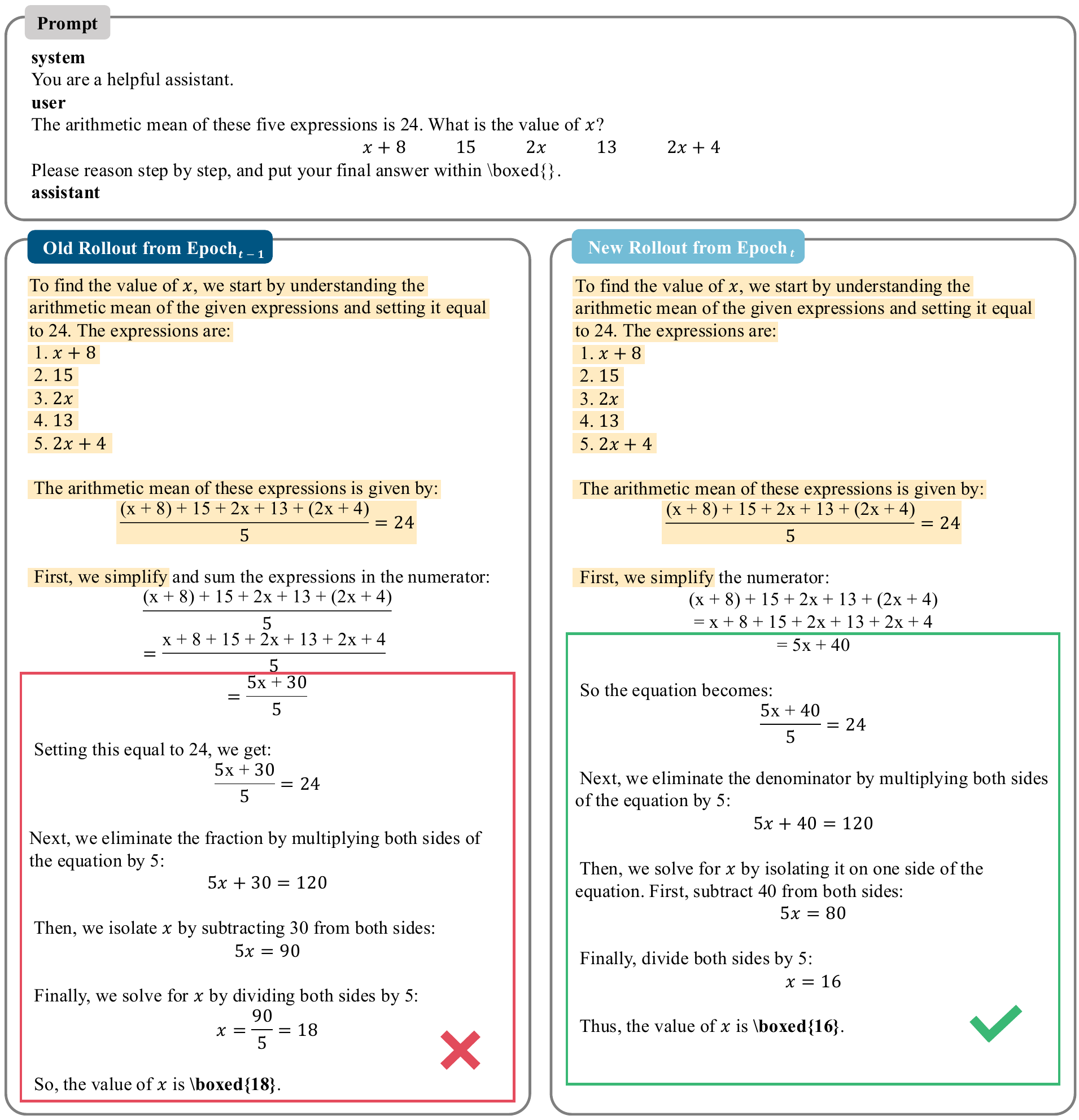}
    \caption{Case study comparing rollouts from previous and current training steps. The prompt denotes the model input. The old rollout and new rollout are generated by the respective model from corresponding epochs. Tokens highlighted in yellow indicate the verified prefix. The red box marks incorrect reasoning steps, whereas the green box highlights correct reasoning steps.}
    \label{fig:case-study-spec-2}
\end{figure*}

\begin{figure*}[!t]
    \centering
    \includegraphics[width=1.0\textwidth]{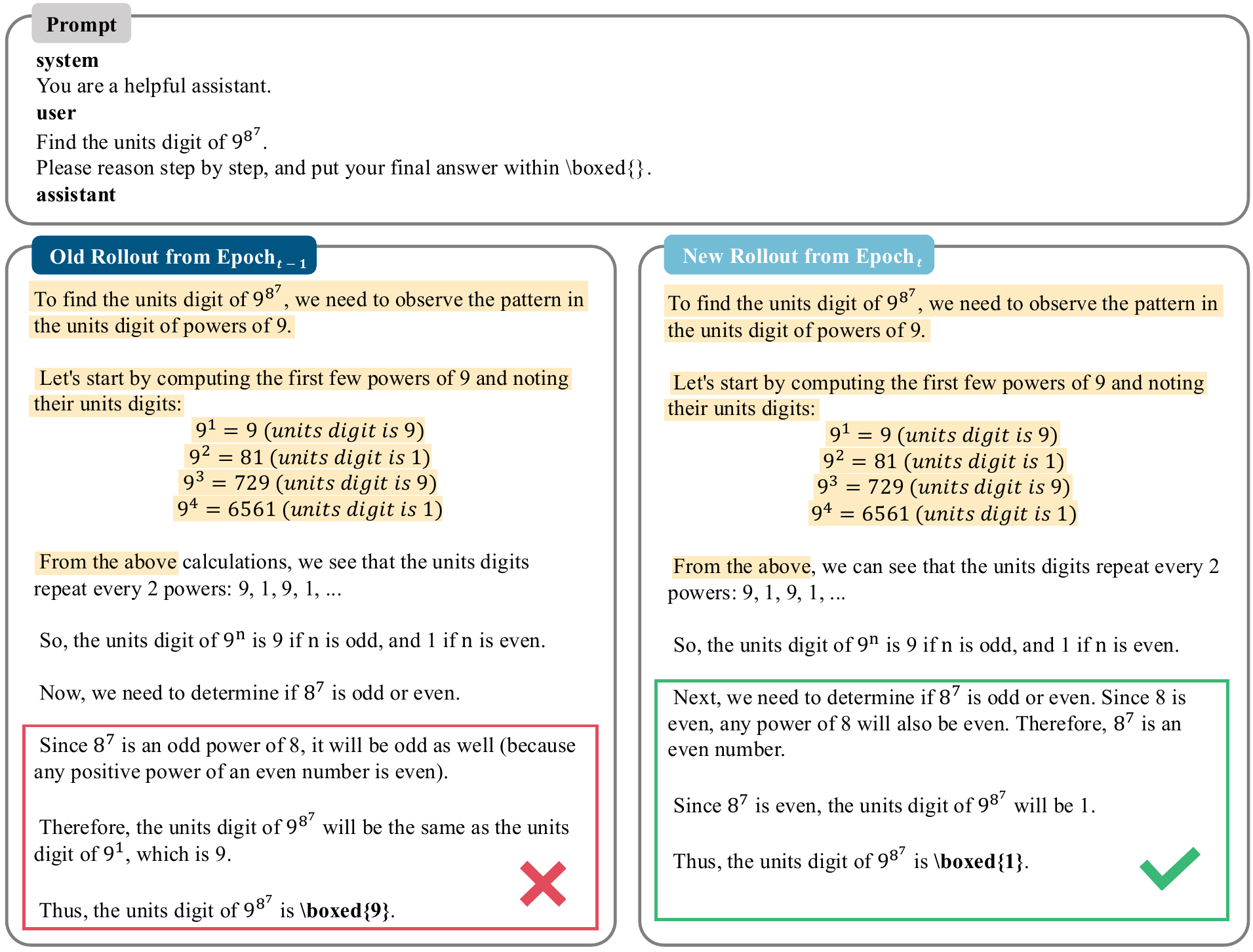}
    \caption{Case study comparing rollouts from previous and current training steps. The prompt denotes the model input. The old rollout and new rollout are generated by the respective model from corresponding epochs. Tokens highlighted in yellow indicate the verified speculative prefix. The red box marks incorrect reasoning steps, whereas the green box highlights correct reasoning steps.}
    \label{fig:case-study-spec-3}
\end{figure*}

\begin{figure*}[!t]
    \centering
    \includegraphics[width=1.0\textwidth]{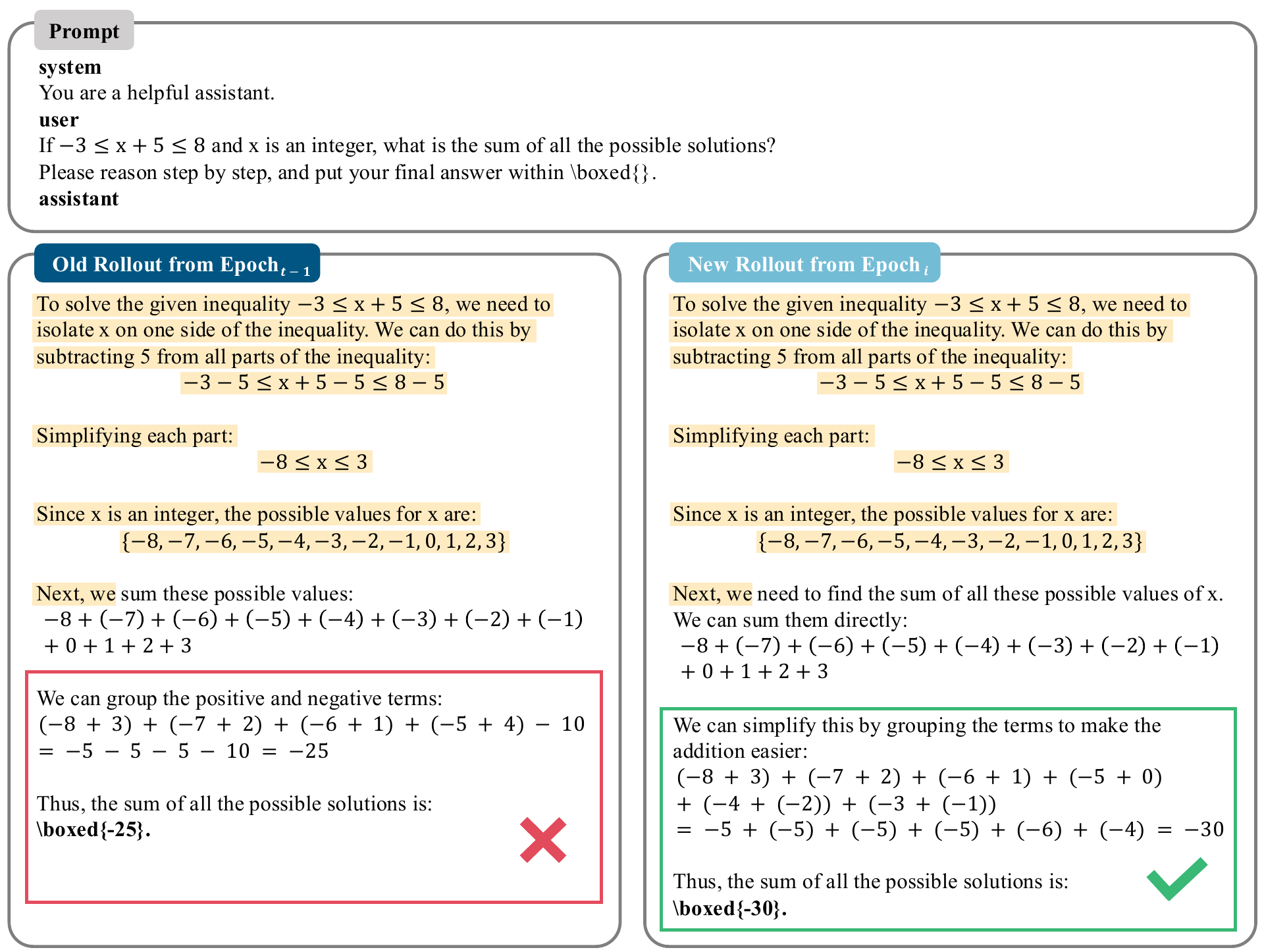}
    \caption{Case study comparing rollouts from previous and current training steps. The prompt denotes the model input. The old rollout and new rollout are generated by the respective model from corresponding epochs. Tokens highlighted in yellow indicate the verified speculative prefix. The red box marks incorrect reasoning steps, whereas the green box highlights correct reasoning steps.}
    \label{fig:case-study-spec-4}
\end{figure*}

To provide a more intuitive understanding of how \ourmethod\ operates during training, we present several case studies comparing cached rollouts from previous epochs with newly generated rollouts under the current policy. 
These examples highlight how speculative prefixes are verified and reused, and how continuation is triggered once a rejection occurs. 
They also illustrate typical scenarios where \ourmethod\ improves efficiency by avoiding redundant generation, while still correcting erroneous reasoning steps when necessary. 
Representative cases are shown in Figures~\ref{fig:case-study-spec-1}, \ref{fig:case-study-spec-2}, \ref{fig:case-study-spec-3}, and \ref{fig:case-study-spec-4}.

\clearpage

\clearpage

\begin{table*}[t]
\centering
\resizebox{\textwidth}{!}{%
\begin{tabular}{l r r r r r r r r r r r}
\toprule
\multirow{3}{*}{\textbf{Algorithm}} &
\multirow{3}{*}{\textbf{Step}} &
\multicolumn{2}{c}{\textbf{Rollout Efficiency}} &
\multicolumn{5}{c}{\textbf{Math Reasoning}} &
\multicolumn{2}{c}{\textbf{OOD}} &
\multirow{3}{*}{\textbf{AVG}} \\
\cmidrule(lr){3-4}\cmidrule(lr){5-9}\cmidrule(lr){10-11}
& & \textbf{Tokens (M)} & \textbf{Speedup} &
\textbf{AMC23} & \textbf{AIME24} &
\textbf{\begin{tabular}[c]{@{}c@{}}MATH\\ 500\end{tabular}} &
\textbf{\begin{tabular}[c]{@{}c@{}}Minerva\\ Math\end{tabular}} &
\textbf{\begin{tabular}[c]{@{}c@{}}Olympiad\\ Bench\end{tabular}} &
\textbf{\begin{tabular}[c]{@{}c@{}}MMLU\\ STEM\end{tabular}} &
\textbf{IFEval} & \\
\midrule
Base Model & 0 & - & - & 21.9 & 2.5 & 45.0 & 12.5 & 16.7 & 39.3 & 17.9 & 22.3 \\

\hdashline
GRPO & 10 & 65.8 & 1.00$\times$ & 27.8 & 4.3 & 54.0 & 18.8 & 22.8 & 43.9 & 20.3 & 27.4 \\
\specrow \hspace{2mm}$\hookrightarrow$~+~\ourmethod & 10 & 43.6 & 1.41$\times$ & 32.5 & 5.7 & 55.8 & 21.7 & 21.8 & 43.0 & 22.4 & 29.0 \\

GRPO & 20 & 127.4 & 1.00$\times$ & 31.2 & 6.1 & 57.8 & 25.4 & 24.7 & 45.5 & 20.3 & 30.1 \\
\specrow \hspace{2mm}$\hookrightarrow$~+~\ourmethod & 20 & 67.2 & 1.66$\times$ & 34.2 & 5.9 & 63.0 & 25.0 & 24.6 & 46.9 & 22.2 & 31.7 \\

GRPO & 30 & 187.7 & 1.00$\times$ & 33.8 & 5.9 & 60.4 & 22.4 & 25.0 & 47.9 & 20.9 & 30.9 \\
\specrow \hspace{2mm}$\hookrightarrow$~+~\ourmethod & 30 & 85.1 & 1.85$\times$ & 34.5 & 7.2 & 64.0 & 25.4 & 27.6 & 51.1 & 26.2 & 33.7 \\

GRPO & 40 & 248.3 & 1.00$\times$ & 36.4 & 6.6 & 65.0 & 23.5 & 24.7 & 50.7 & 21.3 & 32.6 \\
\specrow \hspace{2mm}$\hookrightarrow$~+~\ourmethod & 40 & 102.9 & 1.96$\times$ & 37.0 & 7.4 & 63.8 & 26.5 & 26.1 & 52.2 & 23.5 & 33.8 \\

GRPO & 50 & 309.1 & 1.00$\times$ & 36.6 & 5.9 & 64.2 & 25.7 & 24.6 & 53.7 & 25.1 & 33.7 \\
\specrow \hspace{2mm}$\hookrightarrow$~+~\ourmethod & 50 & 119.4 & 2.06$\times$ & 33.1 & 7.3 & 64.4 & 28.7 & 28.0 & 55.6 & 27.7 & 35.0 \\

GRPO & 60 & 370.4 & 1.00$\times$ & 38.0 & 6.7 & 63.6 & 28.3 & 26.7 & 56.0 & 24.0 & 34.8 \\
\specrow \hspace{2mm}$\hookrightarrow$~+~\ourmethod & 60 & 135.1 & 2.14$\times$ & 36.6 & 7.0 & 66.4 & 26.5 & 29.9 & 54.7 & 28.8 & 35.7 \\

GRPO & 70 & 431.9 & 1.00$\times$ & 35.3 & 8.0 & 61.8 & 26.5 & 28.1 & 55.5 & 26.2 & 34.5 \\
\specrow \hspace{2mm}$\hookrightarrow$~+~\ourmethod & 70 & 153.2 & 2.18$\times$ & 37.0 & 8.1 & 65.4 & 26.5 & 29.9 & 55.6 & 27.4 & 35.7 \\

GRPO & 80 & 493.5 & 1.00$\times$ & 36.1 & 7.2 & 64.4 & 24.3 & 26.4 & 59.4 & 25.0 & 34.7 \\
\specrow \hspace{2mm}$\hookrightarrow$~+~\ourmethod & 80 & 168.1 & 2.24$\times$ & 35.0 & 10.1 & 67.0 & 29.8 & 29.6 & 57.1 & 28.3 & 36.7 \\

GRPO & 90 & 554.8 & 1.00$\times$ & 35.0 & 7.8 & 64.4 & 26.5 & 25.5 & 60.7 & 24.4 & 34.9 \\
\specrow \hspace{2mm}$\hookrightarrow$~+~\ourmethod & 90 & 182.7 & 2.29$\times$ & 38.3 & 9.1 & 68.0 & 29.4 & 29.3 & 58.3 & 28.8 & 37.3 \\

\bottomrule
\end{tabular}
}
\caption{Intermediate training results of \textbf{Qwen-3-1.7B-Base} on DeepMath-6K with \textbf{GRPO} and \ourmethod.
We report rollout efficiency and accuracy every 10 training steps, with GRPO and its \ourmethod\ variant interleaved.}
\label{tab:grpo-1.7b-details}
\end{table*}

\begin{table*}[t]
\centering
\resizebox{\textwidth}{!}{%
\begin{tabular}{l r r r r r r r r r r r}
\toprule
\multirow{3}{*}{\textbf{Algorithm}} &
\multirow{3}{*}{\textbf{Step}} &
\multicolumn{2}{c}{\textbf{Rollout Efficiency}} &
\multicolumn{5}{c}{\textbf{Math Reasoning}} &
\multicolumn{2}{c}{\textbf{OOD}} &
\multirow{3}{*}{\textbf{AVG}} \\
\cmidrule(lr){3-4}\cmidrule(lr){5-9}\cmidrule(lr){10-11}
& & \textbf{Tokens (M)} & \textbf{Speedup} &
\textbf{AMC23} & \textbf{AIME24} &
\textbf{\begin{tabular}[c]{@{}c@{}}MATH\\ 500\end{tabular}} &
\textbf{\begin{tabular}[c]{@{}c@{}}Minerva\\ Math\end{tabular}} &
\textbf{\begin{tabular}[c]{@{}c@{}}Olympiad\\ Bench\end{tabular}} &
\textbf{\begin{tabular}[c]{@{}c@{}}MMLU\\ STEM\end{tabular}} &
\textbf{IFEval} & \\
\midrule
Base Model & 0 & - & - & 40.2 & 11.5 & 67.4 & 27.2 & 34.1 & 60.4 & 29.9 & 38.7 \\

\hdashline
GRPO & 10 & 71.9 & 1.00$\times$ & 56.2 & 16.0 & 80.0 & 32.7 & 44.3 & 64.7 & 34.4 & 46.9 \\
\specrow \hspace{2mm}$\hookrightarrow$~+~\ourmethod & 10 & 53.0 & 1.36$\times$ & 56.2 & 17.1 & 80.0 & 37.1 & 43.9 & 64.2 & 37.2 & 48.0 \\

GRPO & 20 & 158.2 & 1.00$\times$ & 64.7 & 20.8 & 82.8 & 40.4 & 49.3 & 77.3 & 39.0 & 53.5 \\
\specrow \hspace{2mm}$\hookrightarrow$~+~\ourmethod & 20 & 76.4 & 1.96$\times$ & 63.4 & 22.3 & 83.6 & 42.6 & 48.6 & 72.2 & 43.4 & 53.7 \\

GRPO & 30 & 278.1 & 1.00$\times$ & 65.8 & 23.5 & 84.2 & 39.7 & 48.9 & 80.4 & 35.7 & 54.0 \\
\specrow \hspace{2mm}$\hookrightarrow$~+~\ourmethod & 30 & 116.7 & 2.18$\times$ & 66.6 & 24.4 & 85.0 & 43.0 & 49.5 & 80.4 & 47.9 & 56.7 \\

GRPO & 40 & 404.2 & 1.00$\times$ & 66.2 & 24.6 & 85.2 & 40.8 & 50.2 & 82.0 & 37.9 & 55.3 \\
\specrow \hspace{2mm}$\hookrightarrow$~+~\ourmethod & 40 & 156.4 & 2.31$\times$ & 68.9 & 25.1 & 84.2 & 44.5 & 49.0 & 83.3 & 46.8 & 57.4 \\

GRPO & 50 & 532.0 & 1.00$\times$ & 68.0 & 23.5 & 85.4 & 42.6 & 49.5 & 82.8 & 40.1 & 56.0 \\
\specrow \hspace{2mm}$\hookrightarrow$~+~\ourmethod & 50 & 194.4 & 2.36$\times$ & 72.2 & 25.3 & 84.8 & 44.1 & 52.3 & 83.2 & 45.7 & 58.2 \\

GRPO & 60 & 659.3 & 1.00$\times$ & 70.6 & 24.4 & 84.8 & 44.1 & 51.4 & 83.0 & 38.8 & 56.7 \\
\specrow \hspace{2mm}$\hookrightarrow$~+~\ourmethod & 60 & 235.7 & 2.36$\times$ & 70.6 & 27.7 & 85.4 & 43.0 & 51.1 & 84.4 & 44.9 & 58.2 \\

GRPO & 70 & 785.6 & 1.00$\times$ & 70.3 & 24.3 & 84.8 & 43.4 & 51.3 & 84.3 & 34.8 & 56.2 \\
\specrow \hspace{2mm}$\hookrightarrow$~+~\ourmethod & 70 & 279.2 & 2.36$\times$ & 70.9 & 27.1 & 87.0 & 43.8 & 51.7 & 84.7 & 47.5 & 59.0 \\

GRPO & 80 & 910.2 & 1.00$\times$ & 70.6 & 25.3 & 85.8 & 43.4 & 50.2 & 84.7 & 40.1 & 57.2 \\
\specrow \hspace{2mm}$\hookrightarrow$~+~\ourmethod & 80 & 311.1 & 2.42$\times$ & 73.8 & 25.4 & 87.4 & 43.4 & 52.1 & 85.2 & 48.2 & 59.4 \\

GRPO & 90 & 1033.1 & 1.00$\times$ & 69.2 & 24.2 & 86.4 & 43.8 & 53.0 & 84.6 & 41.2 & 57.5 \\
\specrow \hspace{2mm}$\hookrightarrow$~+~\ourmethod & 90 & 336.6 & 2.51$\times$ & 72.8 & 26.9 & 87.8 & 44.1 & 51.0 & 84.5 & 47.7 & 59.3 \\

\bottomrule
\end{tabular}
}
\caption{Intermediate training results of \textbf{Qwen-3-8B-Base} on DeepMath-6K with \textbf{GRPO} and \ourmethod.
We report rollout efficiency and accuracy every 10 training steps, with GRPO and its \ourmethod\ variant interleaved.}
\label{tab:grpo-8b-details}
\end{table*}

\begin{table*}[t]
\centering
\resizebox{\textwidth}{!}{%
\begin{tabular}{l r r r r r r r r r r r}
\toprule
\multirow{3}{*}{\textbf{Algorithm}} &
\multirow{3}{*}{\textbf{Step}} &
\multicolumn{2}{c}{\textbf{Rollout Efficiency}} &
\multicolumn{5}{c}{\textbf{Math Reasoning}} &
\multicolumn{2}{c}{\textbf{OOD}} &
\multirow{3}{*}{\textbf{AVG}} \\
\cmidrule(lr){3-4}\cmidrule(lr){5-9}\cmidrule(lr){10-11}
& & \textbf{Tokens (M)} & \textbf{Speedup} &
\textbf{AMC23} & \textbf{AIME24} &
\textbf{\begin{tabular}[c]{@{}c@{}}MATH\\500\end{tabular}} &
\textbf{\begin{tabular}[c]{@{}c@{}}Minerva\\Math\end{tabular}} &
\textbf{\begin{tabular}[c]{@{}c@{}}Olympiad\\Bench\end{tabular}} &
\textbf{\begin{tabular}[c]{@{}c@{}}MMLU\\STEM\end{tabular}} &
\textbf{IFEval} & \\
\midrule
Base Model & 0  & -     & -     & 7.5 & 0.6 & 14.2 & 4.0 & 2.8 & 32.6 & 37.0 & 14.1 \\

\hdashline
GRPO & 10 & 72.1  & 1.00$\times$ & 5.9 & 0.4 & 14.2 & 2.9 & 3.4 & 31.7 & 38.8 & 13.9 \\
\specrow \hspace{2mm}$\hookrightarrow$~+~\ourmethod & 10 & 47.8  & 1.38$\times$ & 6.9 & 0.1 & 12.8 & 2.6 & 3.4 & 32.8 & 38.6 & 13.9 \\

GRPO & 20 & 141.2 & 1.00$\times$ & 6.4 & 1.0 & 15.6 & 2.6 & 3.9 & 33.5 & 39.7 & 14.7 \\
\specrow \hspace{2mm}$\hookrightarrow$~+~\ourmethod & 20 & 78.5  & 1.57$\times$ & 6.1 & 0.7 & 18.0 & 3.7 & 3.9 & 35.0 & 38.6 & 15.1 \\

GRPO & 30 & 204.5 & 1.00$\times$ & 9.2 & 0.8 & 17.4 & 2.9 & 4.4 & 35.1 & 38.3 & 15.4 \\
\specrow \hspace{2mm}$\hookrightarrow$~+~\ourmethod & 30 & 100.7 & 1.73$\times$ & 6.9 & 0.9 & 18.4 & 2.6 & 4.9 & 33.4 & 40.5 & 15.4 \\

GRPO & 40 & 266.8 & 1.00$\times$ & 7.3 & 0.5 & 15.4 & 3.3 & 4.4 & 33.2 & 41.2 & 15.0 \\
\specrow \hspace{2mm}$\hookrightarrow$~+~\ourmethod & 40 & 115.0 & 1.94$\times$ & 9.1 & 2.0 & 18.6 & 4.4 & 5.5 & 34.2 & 38.8 & 16.1 \\

GRPO & 50 & 326.2 & 1.00$\times$ & 9.7 & 0.8 & 17.6 & 3.7 & 5.3 & 34.3 & 38.1 & 15.6 \\
\specrow \hspace{2mm}$\hookrightarrow$~+~\ourmethod & 50 & 126.3 & 2.12$\times$ & 8.0 & 1.0 & 20.2 & 4.4 & 4.7 & 36.0 & 39.6 & 16.3 \\

GRPO & 60 & 382.9 & 1.00$\times$ & 9.1 & 0.6 & 17.8 & 3.3 & 5.2 & 34.0 & 40.5 & 15.8 \\
\specrow \hspace{2mm}$\hookrightarrow$~+~\ourmethod & 60 & 134.7 & 2.29$\times$ & 9.4 & 1.2 & 19.0 & 4.4 & 5.5 & 35.6 & 37.9 & 16.1 \\

GRPO & 70 & 438.3 & 1.00$\times$ & 7.7 & 1.1 & 17.6 & 4.8 & 5.5 & 34.6 & 37.3 & 15.5 \\
\specrow \hspace{2mm}$\hookrightarrow$~+~\ourmethod & 70 & 143.6 & 2.41$\times$ & 9.4 & 1.2 & 19.8 & 5.1 & 5.8 & 36.1 & 37.2 & 16.4 \\

GRPO & 80 & 495.3 & 1.00$\times$ & 7.0 & 0.8 & 17.6 & 4.0 & 3.9 & 33.3 & 37.2 & 14.8 \\
\specrow \hspace{2mm}$\hookrightarrow$~+~\ourmethod & 80 & 152.8 & 2.52$\times$ & 7.2 & 1.0 & 19.4 & 2.9 & 3.9 & 35.6 & 42.7 & 16.1 \\

GRPO & 90 & 553.9 & 1.00$\times$ & 7.5 & 1.4 & 19.2 & 3.3 & 4.9 & 33.1 & 37.0 & 15.2 \\
\specrow \hspace{2mm}$\hookrightarrow$~+~\ourmethod & 90 & 162.5 & 2.60$\times$ & 8.8 & 1.7 & 19.4 & 1.8 & 5.0 & 34.5 & 37.2 & 15.5 \\
\bottomrule
\end{tabular}
}
\caption{Intermediate training results of \textbf{LLaMA-3.2-1B-Instruct} on DeepMath-6K with \textbf{GRPO} and \ourmethod.
We report rollout efficiency and accuracy every 10 training steps, with GRPO and its \ourmethod\ variant interleaved.}
\label{tab:grpo-1b-details}
\end{table*}

\begin{table*}[t]
\centering
\resizebox{\textwidth}{!}{%
\begin{tabular}{l r r r r r r r r r r r}
\toprule
\multirow{3}{*}{\textbf{Algorithm}} &
\multirow{3}{*}{\textbf{Step}} &
\multicolumn{2}{c}{\textbf{Rollout Efficiency}} &
\multicolumn{5}{c}{\textbf{Math Reasoning}} &
\multicolumn{2}{c}{\textbf{OOD}} &
\multirow{3}{*}{\textbf{AVG}} \\
\cmidrule(lr){3-4}\cmidrule(lr){5-9}\cmidrule(lr){10-11}
& & \textbf{Tokens (M)} & \textbf{Speedup} &
\textbf{AMC23} & \textbf{AIME24} &
\textbf{\begin{tabular}[c]{@{}c@{}}MATH\\ 500\end{tabular}} &
\textbf{\begin{tabular}[c]{@{}c@{}}Minerva\\ Math\end{tabular}} &
\textbf{\begin{tabular}[c]{@{}c@{}}Olympiad\\ Bench\end{tabular}} &
\textbf{\begin{tabular}[c]{@{}c@{}}MMLU\\ STEM\end{tabular}} &
\textbf{IFEval} & \\
\midrule
Base Model & 0 & - & - & 45.2 & 9.1 & 68.6 & 28.3 & 33.5 & 68.3 & 43.6 & 42.4 \\

\hdashline
GRPO & 10 & 64.8 & 1.00$\times$ & 57.3 & 13.3 & 80.2 & 38.2 & 42.8 & 68.3 & 45.5 & 49.4 \\
\specrow \hspace{2mm}$\hookrightarrow$~+~SPEC-RL & 10 & 50.5 & 1.21$\times$ & 58.3 & 13.6 & 80.4 & 40.8 & 44.3 & 69.8 & 45.7 & 50.4 \\

GRPO & 20 & 126.9 & 1.00$\times$ & 59.4 & 14.0 & 80.0 & 41.5 & 45.6 & 69.1 & 47.3 & 51.0 \\
\specrow \hspace{2mm}$\hookrightarrow$~+~SPEC-RL & 20 & 75.9 & 1.50$\times$ & 62.3 & 16.2 & 81.8 & 43.8 & 44.6 & 69.7 & 49.4 & 52.5 \\

GRPO & 30 & 189.7 & 1.00$\times$ & 60.3 & 15.7 & 82.8 & 41.2 & 46.1 & 72.1 & 50.5 & 52.7 \\
\specrow \hspace{2mm}$\hookrightarrow$~+~SPEC-RL & 30 & 99.6 & 1.66$\times$ & 63.1 & 17.1 & 83.8 & 43.4 & 45.5 & 71.1 & 51.0 & 53.6 \\

GRPO & 40 & 253.8 & 1.00$\times$ & 62.7 & 15.9 & 82.6 & 39.0 & 45.5 & 74.3 & 46.0 & 52.3 \\
\specrow \hspace{2mm}$\hookrightarrow$~+~SPEC-RL & 40 & 122.7 & 1.78$\times$ & 64.7 & 18.1 & 84.0 & 44.5 & 47.7 & 72.8 & 53.6 & 55.1 \\

GRPO & 50 & 319.1 & 1.00$\times$ & 63.0 & 16.4 & 84.0 & 41.2 & 46.4 & 76.3 & 45.8 & 53.3 \\
\specrow \hspace{2mm}$\hookrightarrow$~+~SPEC-RL & 50 & 144.4 & 1.88$\times$ & 64.8 & 20.0 & 84.0 & 44.9 & 48.6 & 77.0 & 51.0 & 55.8 \\

GRPO & 60 & 385.3 & 1.00$\times$ & 61.9 & 17.9 & 85.0 & 42.6 & 48.0 & 79.0 & 48.4 & 54.7 \\
\specrow \hspace{2mm}$\hookrightarrow$~+~SPEC-RL & 60 & 172.8 & 1.89$\times$ & 68.1 & 22.5 & 85.6 & 44.9 & 49.3 & 79.0 & 51.0 & 57.2 \\

GRPO & 70 & 452.2 & 1.00$\times$ & 65.8 & 20.1 & 84.4 & 44.5 & 47.1 & 81.4 & 49.4 & 56.1 \\
\specrow \hspace{2mm}$\hookrightarrow$~+~SPEC-RL & 70 & 193.7 & 1.97$\times$ & 66.2 & 21.8 & 85.0 & 40.8 & 49.2 & 82.0 & 53.0 & 56.9 \\

GRPO & 80 & 519.4 & 1.00$\times$ & 65.0 & 19.6 & 83.2 & 43.8 & 49.0 & 83.3 & 49.2 & 56.2 \\
\specrow \hspace{2mm}$\hookrightarrow$~+~SPEC-RL & 80 & 212.9 & 2.04$\times$ & 68.4 & 26.1 & 87.6 & 43.8 & 51.4 & 83.5 & 51.0 & 58.8 \\

GRPO & 90 & 587.6 & 1.00$\times$ & 66.2 & 21.1 & 87.2 & 43.4 & 49.5 & 84.7 & 49.7 & 57.4 \\
\specrow \hspace{2mm}$\hookrightarrow$~+~SPEC-RL & 90 & 234.4 & 2.09$\times$ & 67.8 & 26.2 & 86.8 & 46.3 & 50.7 & 84.9 & 54.9 & 59.7 \\

\bottomrule
\end{tabular}
}
\caption{Intermediate training results of \textbf{Qwen-3-14B-Base} on DeepMath-6K with \textbf{GRPO} and SPEC-RL.
We report rollout efficiency and accuracy every 10 training steps, with GRPO and its SPEC-RL variant interleaved.}
\label{tab:grpo-14b-details}
\end{table*}

\begin{table*}[t]
\centering
\resizebox{\textwidth}{!}{%
\begin{tabular}{l r r r r r r r r r r r}
\toprule
\multirow{3}{*}{\textbf{Algorithm}} &
\multirow{3}{*}{\textbf{Step}} &
\multicolumn{2}{c}{\textbf{Rollout Efficiency}} &
\multicolumn{5}{c}{\textbf{Math Reasoning}} &
\multicolumn{2}{c}{\textbf{OOD}} &
\multirow{3}{*}{\textbf{AVG}} \\
\cmidrule(lr){3-4}\cmidrule(lr){5-9}\cmidrule(lr){10-11}
& & \textbf{Tokens (M)} & \textbf{Speedup} &
\textbf{AMC23} & \textbf{AIME24} &
\textbf{\begin{tabular}[c]{@{}c@{}}MATH\\500\end{tabular}} &
\textbf{\begin{tabular}[c]{@{}c@{}}Minerva\\Math\end{tabular}} &
\textbf{\begin{tabular}[c]{@{}c@{}}Olympiad\\Bench\end{tabular}} &
\textbf{\begin{tabular}[c]{@{}c@{}}MMLU\\STEM\end{tabular}} &
\textbf{IFEval} & \\
\midrule
Base Model & 0 & - & - & 21.9 & 2.5 & 45.0 & 12.5 & 16.7 & 39.3 & 17.9 & 22.3 \\

\hdashline
PPO & 10 & 66.6 & 1.00$\times$ & 27.8 & 3.4 & 54.8 & 19.9 & 21.3 & 42.6 & 19.2 & 27.0 \\
\specrow \hspace{2mm}$\hookrightarrow$~+~\ourmethod & 10 & 46.5 & 1.34$\times$ & 30.6 & 4.5 & 56.8 & 17.6 & 23.3 & 42.9 & 19.4 & 27.9 \\

PPO & 20 & 129.2 & 1.00$\times$ & 30.8 & 4.9 & 60.4 & 22.8 & 25.0 & 46.6 & 21.8 & 30.3 \\
\specrow \hspace{2mm}$\hookrightarrow$~+~\ourmethod & 20 & 80.0 & 1.44$\times$ & 32.3 & 5.5 & 58.0 & 23.2 & 23.4 & 46.9 & 20.9 & 30.0 \\

PPO & 30 & 191.5 & 1.00$\times$ & 32.3 & 5.3 & 59.4 & 22.8 & 26.5 & 47.8 & 19.6 & 30.5 \\
\specrow \hspace{2mm}$\hookrightarrow$~+~\ourmethod & 30 & 106.2 & 1.56$\times$ & 34.1 & 5.6 & 62.6 & 23.5 & 25.6 & 49.8 & 22.7 & 32.0 \\

PPO & 40 & 253.6 & 1.00$\times$ & 32.7 & 6.4 & 61.4 & 23.5 & 25.3 & 50.7 & 22.7 & 31.8 \\
\specrow \hspace{2mm}$\hookrightarrow$~+~\ourmethod & 40 & 126.2 & 1.69$\times$ & 33.1 & 6.6 & 63.2 & 22.4 & 27.3 & 51.3 & 23.7 & 32.5 \\

PPO & 50 & 315.7 & 1.00$\times$ & 34.7 & 6.8 & 61.8 & 26.8 & 25.6 & 51.5 & 21.8 & 32.7 \\
\specrow \hspace{2mm}$\hookrightarrow$~+~\ourmethod & 50 & 157.3 & 1.68$\times$ & 37.3 & 5.9 & 64.4 & 26.1 & 25.9 & 54.2 & 26.6 & 34.3 \\

PPO & 60 & 377.7 & 1.00$\times$ & 35.0 & 7.3 & 63.8 & 29.4 & 26.8 & 53.9 & 23.3 & 34.2 \\
\specrow \hspace{2mm}$\hookrightarrow$~+~\ourmethod & 60 & 172.0 & 1.79$\times$ & 35.5 & 6.7 & 64.2 & 23.5 & 27.0 & 53.8 & 24.4 & 33.6 \\

PPO & 70 & 440.0 & 1.00$\times$ & 35.3 & 6.5 & 60.6 & 25.7 & 26.7 & 55.0 & 22.9 & 33.2 \\
\specrow \hspace{2mm}$\hookrightarrow$~+~\ourmethod & 70 & 194.7 & 1.83$\times$ & 35.8 & 6.7 & 65.8 & 27.6 & 26.8 & 55.1 & 25.9 & 34.8 \\

PPO & 80 & 503.0 & 1.00$\times$ & 36.6 & 6.9 & 63.8 & 25.4 & 29.3 & 58.6 & 23.7 & 34.9 \\
\specrow \hspace{2mm}$\hookrightarrow$~+~\ourmethod & 80 & 207.2 & 1.93$\times$ & 38.8 & 6.2 & 63.6 & 29.8 & 28.0 & 54.3 & 25.7 & 35.2 \\

PPO & 90 & 565.1 & 1.00$\times$ & 35.3 & 8.3 & 63.0 & 26.8 & 25.3 & 59.4 & 25.5 & 34.8 \\
\specrow \hspace{2mm}$\hookrightarrow$~+~\ourmethod & 90 & 230.8 & 1.94$\times$ & 36.9 & 7.7 & 64.8 & 25.4 & 25.9 & 58.6 & 25.9 & 35.0 \\

\bottomrule
\end{tabular}
}
\caption{Intermediate training results of \textbf{Qwen-3-1.7B-Base} on DeepMath-6K with \textbf{PPO} and \ourmethod.
We report rollout efficiency and accuracy every 10 training steps, with GRPO and its \ourmethod\ variant interleaved.}
\label{tab:ppo-1.7b-details}
\end{table*}

\begin{table*}[t]
\centering
\resizebox{\textwidth}{!}{%
\begin{tabular}{l r r r r r r r r r r r}
\toprule
\multirow{3}{*}{\textbf{Algorithm}} &
\multirow{3}{*}{\textbf{Step}} &
\multicolumn{2}{c}{\textbf{Rollout Efficiency}} &
\multicolumn{5}{c}{\textbf{Math Reasoning}} &
\multicolumn{2}{c}{\textbf{OOD}} &
\multirow{3}{*}{\textbf{AVG}} \\
\cmidrule(lr){3-4}\cmidrule(lr){5-9}\cmidrule(lr){10-11}
& & \textbf{Tokens (M)} & \textbf{Speedup} &
\textbf{AMC23} & \textbf{AIME24} &
\textbf{\begin{tabular}[c]{@{}c@{}}MATH\\ 500\end{tabular}} &
\textbf{\begin{tabular}[c]{@{}c@{}}Minerva\\ Math\end{tabular}} &
\textbf{\begin{tabular}[c]{@{}c@{}}Olympiad\\ Bench\end{tabular}} &
\textbf{\begin{tabular}[c]{@{}c@{}}MMLU\\ STEM\end{tabular}} &
\textbf{IFEval} & \\
\midrule
Base Model & 0 & - & - & 40.2 & 11.5 & 67.4 & 27.2 & 34.1 & 60.4 & 29.9 & 38.7 \\

\hdashline
PPO & 10  & 73.4  & 1.00$\times$ & 54.7 & 13.9 & 75.4 & 33.5 & 43.4 & 63.2 & 32.2 & 45.2 \\
\specrow \hspace{2mm}$\hookrightarrow$~+~\ourmethod & 10  & 51.9  & 1.30$\times$ & 51.4 & 13.8 & 79.6 & 32.0 & 42.4 & 62.6 & 36.0 & 45.4 \\

PPO & 20  & 144.8 & 1.00$\times$ & 59.7 & 18.8 & 81.0 & 39.3 & 45.3 & 67.0 & 35.1 & 49.5 \\
\specrow \hspace{2mm}$\hookrightarrow$~+~\ourmethod & 20  & 85.6  & 1.48$\times$ & 60.8 & 19.7 & 82.0 & 39.7 & 45.5 & 67.4 & 37.9 & 50.4 \\

PPO & 30  & 241.0 & 1.00$\times$ & 66.4 & 23.0 & 82.8 & 38.6 & 46.7 & 77.6 & 37.3 & 53.2 \\
\specrow \hspace{2mm}$\hookrightarrow$~+~\ourmethod & 30  & 115.6 & 1.75$\times$ & 64.4 & 22.0 & 82.2 & 40.4 & 48.1 & 74.6 & 42.7 & 53.5 \\

PPO & 40  & 359.0 & 1.00$\times$ & 66.2 & 23.0 & 83.6 & 41.9 & 49.3 & 79.5 & 36.6 & 54.3 \\
\specrow \hspace{2mm}$\hookrightarrow$~+~\ourmethod & 40  & 159.4 & 1.87$\times$ & 66.6 & 23.6 & 84.0 & 39.3 & 49.5 & 77.6 & 43.3 & 54.8 \\

PPO & 50  & 484.2 & 1.00$\times$ & 67.8 & 24.0 & 86.0 & 41.2 & 51.6 & 82.1 & 39.7 & 56.1 \\
\specrow \hspace{2mm}$\hookrightarrow$~+~\ourmethod & 50  & 197.9 & 1.98$\times$ & 69.1 & 23.8 & 84.2 & 42.6 & 49.3 & 81.9 & 41.8 & 56.1 \\

PPO & 60  & 609.9 & 1.00$\times$ & 69.5 & 23.6 & 85.4 & 42.6 & 49.9 & 82.9 & 42.0 & 56.6 \\
\specrow \hspace{2mm}$\hookrightarrow$~+~\ourmethod & 60  & 251.2 & 1.95$\times$ & 68.8 & 25.7 & 84.6 & 43.8 & 52.0 & 81.7 & 43.6 & 57.2 \\

PPO & 70  & 735.0 & 1.00$\times$ & 68.4 & 24.8 & 84.4 & 44.1 & 51.1 & 83.6 & 42.9 & 57.0 \\
\specrow \hspace{2mm}$\hookrightarrow$~+~\ourmethod & 70  & 307.9 & 1.91$\times$ & 66.9 & 25.0 & 84.6 & 42.6 & 50.8 & 84.0 & 44.7 & 56.9 \\

PPO & 80  & 859.9 & 1.00$\times$ & 69.5 & 26.5 & 85.6 & 42.3 & 51.9 & 83.3 & 40.9 & 57.1 \\
\specrow \hspace{2mm}$\hookrightarrow$~+~\ourmethod & 80  & 358.1 & 1.90$\times$ & 69.2 & 24.7 & 83.4 & 44.1 & 50.2 & 84.4 & 43.1 & 57.0 \\

PPO & 90  & 984.0 & 1.00$\times$ & 68.9 & 26.4 & 85.8 & 43.0 & 51.6 & 83.8 & 41.6 & 57.3 \\
\specrow \hspace{2mm}$\hookrightarrow$~+~\ourmethod & 90  & 400.1 & 1.94$\times$ & 70.5 & 25.1 & 85.2 & 43.4 & 50.8 & 84.4 & 41.0 & 57.2 \\

\bottomrule
\end{tabular}
}
\caption{Intermediate training results of \textbf{Qwen-3-8B-Base} on DeepMath-6K with \textbf{PPO} and \ourmethod.
We report rollout efficiency and accuracy every 10 training steps, with PPO and its \ourmethod\ variant interleaved.}
\label{tab:ppo-8b-details}
\end{table*}

\begin{table*}[t]
\centering
\resizebox{\textwidth}{!}{%
\begin{tabular}{l r r r r r r r r r r r}
\toprule
\multirow{3}{*}{\textbf{Algorithm}} &
\multirow{3}{*}{\textbf{Step}} &
\multicolumn{2}{c}{\textbf{Rollout Efficiency}} &
\multicolumn{5}{c}{\textbf{Math Reasoning}} &
\multicolumn{2}{c}{\textbf{OOD}} &
\multirow{3}{*}{\textbf{AVG}} \\
\cmidrule(lr){3-4}\cmidrule(lr){5-9}\cmidrule(lr){10-11}
& & \textbf{Tokens (M)} & \textbf{Speedup} &
\textbf{AMC23} & \textbf{AIME24} &
\textbf{\begin{tabular}[c]{@{}c@{}}MATH\\ 500\end{tabular}} &
\textbf{\begin{tabular}[c]{@{}c@{}}Minerva\\ Math\end{tabular}} &
\textbf{\begin{tabular}[c]{@{}c@{}}Olympiad\\ Bench\end{tabular}} &
\textbf{\begin{tabular}[c]{@{}c@{}}MMLU\\ STEM\end{tabular}} &
\textbf{IFEval} & \\
\midrule
Base Model & 0 & - & - & 7.5 & 0.6 & 14.2 & 4.0 & 2.8 & 32.6 & 37.0 & 14.1 \\

\hdashline
PPO & 10  & 65.5  & 1.00$\times$ & 5.5 & 0.8 & 11.4 & 4.0 & 3.6 & 33.0 & 42.0 & 14.3 \\
\specrow \hspace{2mm}$\hookrightarrow$~+~\ourmethod & 10  & 50.2  & 1.26$\times$ & 4.4 & 0.9 & 14.0 & 2.6 & 3.7 & 32.5 & 41.0 & 14.2 \\

PPO & 20  & 131.1 & 1.00$\times$ & 8.1 & 1.0 & 16.2 & 3.7 & 4.7 & 33.9 & 35.9 & 14.8 \\
\specrow \hspace{2mm}$\hookrightarrow$~+~\ourmethod & 20  & 89.0  & 1.35$\times$ & 8.0 & 0.6 & 16.2 & 2.9 & 5.0 & 34.1 & 41.2 & 15.4 \\

PPO & 30  & 192.0 & 1.00$\times$ & 6.7 & 0.7 & 17.2 & 3.7 & 4.1 & 35.3 & 38.6 & 15.2 \\
\specrow \hspace{2mm}$\hookrightarrow$~+~\ourmethod & 30  & 118.3 & 1.47$\times$ & 8.9 & 0.6 & 17.4 & 4.4 & 6.4 & 34.2 & 38.8 & 15.8 \\

PPO & 40  & 250.3 & 1.00$\times$ & 8.3 & 1.0 & 19.6 & 2.9 & 4.4 & 35.1 & 39.0 & 15.8 \\
\specrow \hspace{2mm}$\hookrightarrow$~+~\ourmethod & 40  & 134.8 & 1.63$\times$ & 8.8 & 1.0 & 18.6 & 3.3 & 6.1 & 33.2 & 39.9 & 15.8 \\

PPO & 50  & 306.8 & 1.00$\times$ & 7.5 & 1.0 & 19.2 & 4.8 & 4.6 & 32.3 & 40.1 & 15.6 \\
\specrow \hspace{2mm}$\hookrightarrow$~+~\ourmethod & 50  & 147.2 & 1.78$\times$ & 9.1 & 1.5 & 19.4 & 5.1 & 5.0 & 35.3 & 40.3 & 16.5 \\

PPO & 60  & 361.6 & 1.00$\times$ & 9.4 & 1.0 & 18.4 & 5.1 & 4.3 & 35.1 & 41.6 & 16.4 \\
\specrow \hspace{2mm}$\hookrightarrow$~+~\ourmethod & 60  & 160.5 & 1.89$\times$ & 8.4 & 1.5 & 19.2 & 4.8 & 5.5 & 34.4 & 39.9 & 16.2 \\

PPO & 70  & 415.4 & 1.00$\times$ & 7.5 & 1.2 & 17.8 & 3.7 & 6.1 & 35.1 & 40.1 & 15.9 \\
\specrow \hspace{2mm}$\hookrightarrow$~+~\ourmethod & 70  & 175.8 & 1.95$\times$ & 10.0 & 0.9 & 19.6 & 3.3 & 5.5 & 34.9 & 39.9 & 16.3 \\

PPO & 80  & 469.0 & 1.00$\times$ & 9.5 & 1.0 & 19.6 & 4.8 & 4.1 & 34.8 & 41.6 & 16.5 \\
\specrow \hspace{2mm}$\hookrightarrow$~+~\ourmethod & 80  & 188.8 & 2.02$\times$ & 9.7 & 1.4 & 19.0 & 5.5 & 6.4 & 36.1 & 40.3 & 16.9 \\

PPO & 90  & 521.5 & 1.00$\times$ & 7.8 & 0.8 & 20.8 & 4.0 & 6.4 & 34.3 & 42.7 & 16.7 \\
\specrow \hspace{2mm}$\hookrightarrow$~+~\ourmethod & 90  & 210.6 & 2.01$\times$ & 9.2 & 1.4 & 20.2 & 5.5 & 5.0 & 35.3 & 40.7 & 16.8 \\

\bottomrule
\end{tabular}
}
\caption{Intermediate training results of \textbf{LLaMA-3.2-1B-Instruct} on DeepMath-6K with \textbf{PPO} and \ourmethod.
We report rollout efficiency and accuracy every 10 training steps, with PPO and its \ourmethod\ variant interleaved.}
\label{tab:ppo-1b-details}
\end{table*}

\begin{table*}[t]
\centering
\resizebox{\textwidth}{!}{%
\begin{tabular}{l r r r r r r r r r r r}
\toprule
\multirow{3}{*}{\textbf{Algorithm}} &
\multirow{3}{*}{\textbf{Step}} &
\multicolumn{2}{c}{\textbf{Rollout Efficiency}} &
\multicolumn{5}{c}{\textbf{Math Reasoning}} &
\multicolumn{2}{c}{\textbf{OOD}} &
\multirow{3}{*}{\textbf{AVG}} \\
\cmidrule(lr){3-4}\cmidrule(lr){5-9}\cmidrule(lr){10-11}
& & \textbf{Tokens (M)} & \textbf{Speedup} &
\textbf{AMC23} & \textbf{AIME24} &
\textbf{\begin{tabular}[c]{@{}c@{}}MATH\\ 500\end{tabular}} &
\textbf{\begin{tabular}[c]{@{}c@{}}Minerva\\ Math\end{tabular}} &
\textbf{\begin{tabular}[c]{@{}c@{}}Olympiad\\ Bench\end{tabular}} &
\textbf{\begin{tabular}[c]{@{}c@{}}MMLU\\ STEM\end{tabular}} &
\textbf{IFEval} & \\
\midrule
Base Model & 0 & - & - & 45.2 & 9.1 & 68.6 & 28.3 & 33.5 & 68.3 & 43.6 & 42.4 \\

\hdashline
PPO & 10 & 65.2 & 1.00$\times$ & 55.3 & 14.1 & 78.6 & 38.2 & 42.2 & 70.6 & 41.4 & 48.6 \\
\specrow \hspace{2mm}$\hookrightarrow$~+~\ourmethod & 10 & 50.0 & 1.28$\times$ & 55.9 & 13.1 & 78.2 & 36.0 & 41.9 & 70.0 & 42.7 & 48.3 \\

PPO & 20 & 126.8 & 1.00$\times$ & 60.5 & 15.3 & 79.0 & 40.4 & 45.9 & 69.6 & 44.2 & 50.7 \\
\specrow \hspace{2mm}$\hookrightarrow$~+~\ourmethod & 20 & 77.1 & 1.58$\times$ & 60.0 & 14.2 & 80.0 & 38.6 & 44.7 & 70.2 & 47.3 & 50.7 \\

PPO & 30 & 188.0 & 1.00$\times$ & 58.4 & 15.1 & 79.8 & 40.4 & 45.3 & 72.2 & 47.0 & 51.2 \\
\specrow \hspace{2mm}$\hookrightarrow$~+~\ourmethod & 30 & 97.9 & 1.78$\times$ & 61.1 & 17.4 & 82.0 & 39.7 & 45.5 & 72.4 & 48.6 & 52.4 \\

PPO & 40 & 249.6 & 1.00$\times$ & 63.0 & 17.6 & 83.6 & 40.4 & 45.6 & 73.0 & 47.9 & 53.0 \\
\specrow \hspace{2mm}$\hookrightarrow$~+~\ourmethod & 40 & 126.4 & 1.81$\times$ & 69.2 & 23.5 & 86.2 & 43.0 & 49.5 & 74.8 & 47.7 & 56.3 \\

PPO & 50 & 312.8 & 1.00$\times$ & 63.6 & 17.4 & 84.4 & 41.9 & 48.0 & 77.1 & 47.7 & 54.3 \\
\specrow \hspace{2mm}$\hookrightarrow$~+~\ourmethod & 50 & 164.0 & 1.75$\times$ & 66.2 & 24.4 & 84.0 & 43.4 & 51.3 & 76.9 & 47.5 & 56.2 \\

PPO & 60 & 378.2 & 1.00$\times$ & 66.6 & 18.2 & 84.8 & 44.1 & 47.6 & 81.2 & 48.4 & 55.8 \\
\specrow \hspace{2mm}$\hookrightarrow$~+~\ourmethod & 60 & 197.0 & 1.75$\times$ & 69.4 & 23.5 & 85.8 & 47.1 & 49.5 & 78.4 & 45.5 & 57.0 \\

PPO & 70 & 445.0 & 1.00$\times$ & 64.5 & 18.2 & 85.6 & 41.2 & 48.6 & 84.1 & 47.5 & 55.7 \\
\specrow \hspace{2mm}$\hookrightarrow$~+~\ourmethod & 70 & 236.1 & 1.71$\times$ & 71.6 & 27.0 & 87.0 & 43.4 & 53.3 & 80.6 & 46.2 & 58.4 \\

PPO & 80 & 511.7 & 1.00$\times$ & 65.3 & 18.1 & 87.0 & 44.9 & 51.1 & 85.0 & 48.8 & 57.2 \\
\specrow \hspace{2mm}$\hookrightarrow$~+~\ourmethod & 80 & 272.4 & 1.69$\times$ & 72.8 & 27.9 & 88.6 & 44.5 & 52.3 & 83.6 & 48.1 & 59.7 \\

PPO & 90 & 579.2 & 1.00$\times$ & 67.5 & 22.1 & 86.6 & 41.9 & 50.2 & 87.1 & 49.5 & 57.8 \\
\specrow \hspace{2mm}$\hookrightarrow$~+~\ourmethod & 90 & 310.1 & 1.66$\times$ & 75.5 & 29.1 & 86.2 & 41.9 & 53.6 & 82.3 & 47.1 & 59.4 \\

\bottomrule
\end{tabular}
}
\caption{Intermediate training results of \textbf{Qwen-3-14B-Base} on DeepMath-6K with \textbf{PPO} and \ourmethod.
We report rollout efficiency and accuracy every 10 training steps, with PPO and its \ourmethod\ variant interleaved.}
\label{tab:ppo-14b-details}
\end{table*}

\begin{table*}[t]
\centering
\resizebox{\textwidth}{!}{%
\begin{tabular}{l r r r r r r r r r r r r}
\toprule
\multirow{3}{*}{\textbf{Algorithm}} &
\multirow{3}{*}{\textbf{Step}} &
\multirow{3}{*}{\textbf{Gen-Step}} &
\multicolumn{2}{c}{\textbf{Rollout Efficiency}} &
\multicolumn{5}{c}{\textbf{Math Reasoning}} &
\multicolumn{2}{c}{\textbf{OOD}} &
\multirow{3}{*}{\textbf{AVG}} \\
\cmidrule(lr){4-5}\cmidrule(lr){6-10}\cmidrule(lr){11-12}
& & & \textbf{Tokens (M)} & \textbf{Speedup} &
\textbf{AMC23} & \textbf{AIME24} &
\textbf{\begin{tabular}[c]{@{}c@{}}MATH\\500\end{tabular}} &
\textbf{\begin{tabular}[c]{@{}c@{}}Minerva\\Math\end{tabular}} &
\textbf{\begin{tabular}[c]{@{}c@{}}Olympiad\\Bench\end{tabular}} &
\textbf{\begin{tabular}[c]{@{}c@{}}MMLU\\STEM\end{tabular}} &
\textbf{IFEval} & \\
\midrule
Base Model & 0 & 0 & - & - & 21.9 & 2.5 & 45.0 & 12.5 & 16.7 & 39.3 & 17.9 & 22.3 \\

\hdashline
DAPO & 5 & 10 & 65.2 & 1.00$\times$ & 28.6 & 3.4 & 50.0 & 17.6 & 20.0 & 41.5 & 20.1 & 25.9 \\
\specrow \hspace{2mm}$\hookrightarrow$~+~\ourmethod & 5 & 10 & 46.7 & 1.25$\times$ & 24.2 & 3.1 & 53.4 & 20.2 & 21.0 & 43.0 & 17.7 & 26.1 \\

DAPO & 10 & 20 & 127.4 & 1.00$\times$ & 27.3 & 4.1 & 58.4 & 19.1 & 22.4 & 42.9 & 19.6 & 27.7 \\
\specrow \hspace{2mm}$\hookrightarrow$~+~\ourmethod & 10 & 20 & 70.6 & 1.52$\times$ & 31.4 & 5.0 & 56.8 & 16.5 & 20.7 & 44.1 & 21.6 & 28.0 \\

DAPO & 15 & 30 & 187.7 & 1.00$\times$ & 29.5 & 4.8 & 56.4 & 19.9 & 22.7 & 45.8 & 20.7 & 28.5 \\
\specrow \hspace{2mm}$\hookrightarrow$~+~\ourmethod & 15 & 30 & 95.5 & 1.60$\times$ & 32.2 & 4.5 & 58.4 & 23.9 & 24.9 & 45.0 & 26.1 & 30.7 \\

DAPO & 20 & 40 & 247.6 & 1.00$\times$ & 34.1 & 4.5 & 54.8 & 23.2 & 21.8 & 46.8 & 23.3 & 29.8 \\
\specrow \hspace{2mm}$\hookrightarrow$~+~\ourmethod & 20 & 40 & 109.5 & 1.74$\times$ & 35.0 & 5.1 & 57.2 & 25.0 & 24.9 & 47.7 & 24.6 & 31.4 \\

DAPO & 25 & 50 & 307.5 & 1.00$\times$ & 32.0 & 5.0 & 59.4 & 20.6 & 25.9 & 47.2 & 19.4 & 29.9 \\
\specrow \hspace{2mm}$\hookrightarrow$~+~\ourmethod & 25 & 50 & 124.9 & 1.84$\times$ & 33.9 & 5.0 & 59.0 & 22.8 & 23.0 & 49.2 & 25.7 & 31.2 \\

DAPO & 30 & 60 & 367.0 & 1.00$\times$ & 31.2 & 5.4 & 60.6 & 25.0 & 24.3 & 48.0 & 22.2 & 31.0 \\
\specrow \hspace{2mm}$\hookrightarrow$~+~\ourmethod & 30 & 60 & 137.7 & 1.94$\times$ & 30.9 & 6.1 & 60.4 & 24.6 & 25.0 & 50.4 & 26.2 & 31.9 \\

DAPO & 35 & 70 & 425.6 & 1.00$\times$ & 34.1 & 5.0 & 59.8 & 27.9 & 24.3 & 49.7 & 22.7 & 31.9 \\
\specrow \hspace{2mm}$\hookrightarrow$~+~\ourmethod & 35 & 70 & 149.1 & 2.02$\times$ & 33.9 & 5.6 & 62.2 & 25.0 & 25.3 & 51.6 & 25.0 & 32.7 \\

DAPO & 40 & 80 & 484.6 & 1.00$\times$ & 35.5 & 4.9 & 61.6 & 24.6 & 25.0 & 50.8 & 22.7 & 32.2 \\
\specrow \hspace{2mm}$\hookrightarrow$~+~\ourmethod & 40 & 80 & 160.2 & 2.10$\times$ & 34.7 & 5.3 & 60.2 & 25.4 & 26.5 & 53.7 & 27.4 & 33.3 \\

DAPO & 45 & 90 & 543.1 & 1.00$\times$ & 33.1 & 4.9 & 60.8 & 24.6 & 23.0 & 52.2 & 24.8 & 31.9 \\
\specrow \hspace{2mm}$\hookrightarrow$~+~\ourmethod & 45 & 90 & 171.6 & 2.17$\times$ & 33.1 & 4.9 & 60.0 & 25.7 & 25.5 & 53.5 & 27.0 & 32.8 \\
\bottomrule
\end{tabular}
}
\caption{Intermediate training results of \textbf{Qwen-3-1.7B-Base} on DeepMath-6K with \textbf{DAPO} and \ourmethod.
Since DAPO adopts \textit{Dynamic Sampling}, one training step may correspond to multiple generation steps; thus we additionally report the \textbf{Gen-Step} column to indicate how many rollout batches the model has consumed.}
\label{tab:dapo-1.7b-details}
\end{table*}

\begin{table*}[t]
\centering
\resizebox{\textwidth}{!}{%
\begin{tabular}{l r r r r r r r r r r r r}
\toprule
\multirow{3}{*}{\textbf{Algorithm}} &
\multirow{3}{*}{\textbf{Step}} &
\multirow{3}{*}{\textbf{Gen-Step}} &
\multicolumn{2}{c}{\textbf{Rollout Efficiency}} &
\multicolumn{5}{c}{\textbf{Math Reasoning}} &
\multicolumn{2}{c}{\textbf{OOD}} &
\multirow{3}{*}{\textbf{AVG}} \\
\cmidrule(lr){4-5}\cmidrule(lr){6-10}\cmidrule(lr){11-12}
& & & \textbf{Tokens (M)} & \textbf{Speedup} &
\textbf{AMC23} & \textbf{AIME24} &
\textbf{\begin{tabular}[c]{@{}c@{}}MATH\\500\end{tabular}} &
\textbf{\begin{tabular}[c]{@{}c@{}}Minerva\\Math\end{tabular}} &
\textbf{\begin{tabular}[c]{@{}c@{}}Olympiad\\Bench\end{tabular}} &
\textbf{\begin{tabular}[c]{@{}c@{}}MMLU\\STEM\end{tabular}} &
\textbf{IFEval} & \\
\midrule

Base Model & 0 & 0 & - & - &
40.2 & 11.5 & 67.4 & 27.2 & 34.1 & 60.4 & 29.9 & 38.7 \\

\hdashline
DAPO & 5 & 10 & 75.0 & 1.00$\times$ &
49.7 & 12.9 & 73.6 & 27.9 & 39.1 & 60.5 & 32.0 & 42.2 \\

\specrow \hspace{2mm}$\hookrightarrow$~+~\ourmethod & 5 & 10 & 59.0 & 1.20$\times$ &
51.1 & 13.3 & 75.0 & 32.7 & 38.8 & 63.3 & 34.0 & 44.0 \\

DAPO & 10 & 20 & 148.8 & 1.00$\times$ &
58.0 & 17.1 & 78.6 & 36.8 & 43.4 & 64.0 & 36.0 & 47.7 \\

\specrow \hspace{2mm}$\hookrightarrow$~+~\ourmethod & 10 & 20 & 90.3 & 1.45$\times$ &
56.1 & 14.5 & 79.0 & 37.1 & 40.7 & 63.0 & 34.2 & 46.4 \\

DAPO & 15 & 30 & 235.5 & 1.00$\times$ &
61.4 & 19.0 & 80.6 & 39.7 & 47.4 & 70.6 & 38.1 & 51.0 \\

\specrow \hspace{2mm}$\hookrightarrow$~+~\ourmethod & 15 & 30 & 116.8 & 1.73$\times$ &
61.4 & 21.8 & 81.8 & 42.6 & 47.9 & 69.6 & 39.0 & 52.0 \\

DAPO & 20 & 40 & 354.9 & 1.00$\times$ &
65.9 & 23.0 & 84.6 & 41.2 & 46.1 & 77.5 & 36.4 & 53.5 \\

\specrow \hspace{2mm}$\hookrightarrow$~+~\ourmethod & 20 & 40 & 152.2 & 2.00$\times$ &
63.4 & 24.7 & 83.8 & 39.0 & 49.8 & 75.4 & 38.1 & 53.5 \\

DAPO & 25 & 51 & 509.1 & 1.00$\times$ &
64.7 & 22.2 & 83.4 & 39.3 & 49.6 & 79.8 & 38.8 & 54.0 \\

\specrow \hspace{2mm}$\hookrightarrow$~+~\ourmethod & 25 & 50 & 199.8 & 2.19$\times$ &
64.4 & 22.9 & 85.4 & 41.5 & 47.9 & 78.7 & 42.1 & 54.7 \\

DAPO & 30 & 63 & 685.0 & 1.00$\times$ &
63.9 & 25.4 & 83.8 & 44.5 & 48.9 & 81.1 & 39.2 & 55.3 \\

\specrow \hspace{2mm}$\hookrightarrow$~+~\ourmethod & 30 & 60 & 239.9 & 2.48$\times$ &
67.8 & 24.5 & 84.0 & 39.7 & 48.7 & 80.2 & 40.3 & 55.0 \\

DAPO & 35 & 75 & 867.6 & 1.00$\times$ &
68.8 & 22.7 & 82.8 & 40.8 & 49.5 & 81.9 & 38.1 & 54.9 \\

\specrow \hspace{2mm}$\hookrightarrow$~+~\ourmethod & 35 & 70 & 278.9 & 2.73$\times$ &
68.6 & 25.5 & 84.8 & 40.4 & 49.9 & 80.8 & 44.4 & 56.3 \\

DAPO & 40 & 87 & 1052.2 & 1.00$\times$ &
65.2 & 24.0 & 84.8 & 40.1 & 48.6 & 82.4 & 39.6 & 55.0 \\

\specrow \hspace{2mm}$\hookrightarrow$~+~\ourmethod & 40 & 82 & 326.2 & 2.88$\times$ &
69.7 & 26.4 & 84.4 & 43.8 & 50.4 & 82.2 & 44.4 & 57.3 \\

\bottomrule
\end{tabular}
}
\caption{Intermediate training results of \textbf{Qwen-3-8B-Base} on DeepMath-6K with \textbf{DAPO} and \ourmethod.
Since DAPO adopts \textit{Dynamic Sampling}, one training step may correspond to multiple generation steps; thus we additionally report the \textbf{Gen-Step} column to indicate how many rollout batches the model has consumed.}
\label{tab:dapo-8b-details}
\end{table*}

\begin{table*}[t]
\centering
\resizebox{\textwidth}{!}{%
\begin{tabular}{l r r r r r r r r r r r r}
\toprule
\multirow{3}{*}{\textbf{Algorithm}} &
\multirow{3}{*}{\textbf{Step}} &
\multirow{3}{*}{\textbf{Gen-Step}} &
\multicolumn{2}{c}{\textbf{Rollout Efficiency}} &
\multicolumn{5}{c}{\textbf{Math Reasoning}} &
\multicolumn{2}{c}{\textbf{OOD}} &
\multirow{3}{*}{\textbf{AVG}} \\
\cmidrule(lr){4-5}\cmidrule(lr){6-10}\cmidrule(lr){11-12}
& & & \textbf{Tokens (M)} & \textbf{Speedup} &
\textbf{AMC23} & \textbf{AIME24} &
\textbf{\begin{tabular}[c]{@{}c@{}}MATH\\500\end{tabular}} &
\textbf{\begin{tabular}[c]{@{}c@{}}Minerva\\Math\end{tabular}} &
\textbf{\begin{tabular}[c]{@{}c@{}}Olympiad\\Bench\end{tabular}} &
\textbf{\begin{tabular}[c]{@{}c@{}}MMLU\\STEM\end{tabular}} &
\textbf{IFEval} & \\
\midrule

Base Model & 0 & 0 & - & - & 7.5 & 0.6 & 14.2 & 4.0 & 2.8 & 32.6 & 37.0 & 14.1 \\

\hdashline

DAPO & 5 & 15 & 105.6 & 1.00$\times$ & 5.3 & 0.5 & 14.4 & 2.9 & 3.3 & 32.6 & 38.8 & 14.0 \\
\specrow \hspace{2mm}$\hookrightarrow$~+~\ourmethod & 5 & 15 & 52.4 & 1.95$\times$ & 5.8 & 1.1 & 14.4 & 2.6 & 3.0 & 32.7 & 38.3 & 14.0 \\

DAPO & 10 & 27 & 179.8 & 1.00$\times$ & 6.1 & 0.6 & 14.6 & 2.6 & 4.1 & 34.8 & 40.3 & 14.7 \\
\specrow \hspace{2mm}$\hookrightarrow$~+~\ourmethod & 10 & 28 & 69.2 & 2.16$\times$ & 7.2 & 1.1 & 16.6 & 2.6 & 4.9 & 33.7 & 39.2 & 15.0 \\

DAPO & 15 & 38 & 239.8 & 1.00$\times$ & 8.1 & 0.9 & 18.4 & 4.4 & 4.4 & 33.9 & 37.3 & 15.3 \\
\specrow \hspace{2mm}$\hookrightarrow$~+~\ourmethod & 15 & 39 & 79.2 & 2.19$\times$ & 9.8 & 0.9 & 16.4 & 6.2 & 3.4 & 33.9 & 37.5 & 15.4 \\

DAPO & 20 & 53 & 322.1 & 1.00$\times$ & 8.4 & 0.6 & 18.8 & 2.9 & 5.6 & 34.6 & 38.8 & 15.7 \\
\specrow \hspace{2mm}$\hookrightarrow$~+~\ourmethod & 20 & 53 & 92.4 & 2.31$\times$ & 11.1 & 1.7 & 19.6 & 5.9 & 5.5 & 34.2 & 38.1 & 16.6 \\

DAPO & 25 & 68 & 402.9 & 1.00$\times$ & 10.0 & 1.0 & 19.8 & 3.3 & 4.6 & 34.5 & 38.3 & 15.9 \\
\specrow \hspace{2mm}$\hookrightarrow$~+~\ourmethod & 25 & 68 & 105.3 & 2.43$\times$ & 10.9 & 2.1 & 19.8 & 4.0 & 6.1 & 35.5 & 35.5 & 16.3 \\

DAPO & 30 & 83 & 482.6 & 1.00$\times$ & 9.2 & 0.9 & 19.2 & 4.0 & 5.5 & 33.0 & 38.6 & 15.8 \\
\specrow \hspace{2mm}$\hookrightarrow$~+~\ourmethod & 30 & 83 & 123.1 & 2.48$\times$ & 11.6 & 2.0 & 20.2 & 4.0 & 5.5 & 35.5 & 38.4 & 16.7 \\

\bottomrule
\end{tabular}
}
\caption{Intermediate training results of \textbf{LLaMA-3.2-1B-Instruct} on DeepMath-6K with \textbf{DAPO} and \ourmethod.
Since DAPO adopts \textit{Dynamic Sampling}, one training step may correspond to multiple generation steps; thus we additionally report the \textbf{Gen-Step} column to indicate how many rollout batches the model has consumed.}
\label{tab:dapo-1b-details}
\end{table*}

\begin{table*}[t]
\centering
\resizebox{\textwidth}{!}{%
\begin{tabular}{l r r r r r r r r r r r r}
\toprule
\multirow{3}{*}{\textbf{Algorithm}} &
\multirow{3}{*}{\textbf{Step}} &
\multirow{3}{*}{\textbf{Gen-Step}} &
\multicolumn{2}{c}{\textbf{Rollout Efficiency}} &
\multicolumn{5}{c}{\textbf{Math Reasoning}} &
\multicolumn{2}{c}{\textbf{OOD}} &
\multirow{3}{*}{\textbf{AVG}} \\
\cmidrule(lr){4-5}\cmidrule(lr){6-10}\cmidrule(lr){11-12}
& & & \textbf{Tokens (M)} & \textbf{Speedup} &
\textbf{AMC23} & \textbf{AIME24} &
\textbf{\begin{tabular}[c]{@{}c@{}}MATH\\500\end{tabular}} &
\textbf{\begin{tabular}[c]{@{}c@{}}Minerva\\Math\end{tabular}} &
\textbf{\begin{tabular}[c]{@{}c@{}}Olympiad\\Bench\end{tabular}} &
\textbf{\begin{tabular}[c]{@{}c@{}}MMLU\\STEM\end{tabular}} &
\textbf{IFEval} & \\
\midrule

Base Model & 0 & 0 & - & - &
45.2 & 9.1 & 68.6 & 28.3 & 33.5 & 68.3 & 43.6 & 42.4 \\

\hdashline

DAPO & 5 & 10 & 68.2 & 1.00$\times$ &
55.0 & 13.0 & 77.4 & 39.0 & 40.4 & 69.0 & 44.9 & 48.4 \\
\specrow \hspace{2mm}$\hookrightarrow$~+~SPEC-RL & 5 & 10 & 51.6 & 1.28$\times$ &
55.9 & 13.5 & 76.8 & 37.1 & 39.3 & 69.5 & 41.6 & 47.7 \\

DAPO & 10 & 20 & 136.1 & 1.00$\times$ &
58.6 & 13.1 & 81.2 & 38.2 & 43.7 & 68.8 & 44.4 & 49.7 \\
\specrow \hspace{2mm}$\hookrightarrow$~+~SPEC-RL & 10 & 20 & 82.6 & 1.52$\times$ &
59.1 & 14.2 & 79.4 & 39.0 & 44.0 & 67.3 & 46.0 & 49.9 \\

DAPO & 15 & 31 & 212.0 & 1.00$\times$ &
58.3 & 14.1 & 81.0 & 37.9 & 44.9 & 69.6 & 46.2 & 50.3 \\
\specrow \hspace{2mm}$\hookrightarrow$~+~SPEC-RL & 15 & 30 & 109.0 & 1.78$\times$ &
62.5 & 14.7 & 81.4 & 42.3 & 45.2 & 67.9 & 50.1 & 52.0 \\

DAPO & 20 & 44 & 303.9 & 1.00$\times$ &
59.4 & 14.4 & 81.6 & 38.2 & 44.3 & 69.9 & 47.3 & 50.7 \\
\specrow \hspace{2mm}$\hookrightarrow$~+~SPEC-RL & 20 & 40 & 129.4 & 2.11$\times$ &
62.3 & 15.6 & 81.2 & 40.8 & 42.7 & 69.4 & 50.6 & 51.8 \\

DAPO & 25 & 59 & 412.7 & 1.00$\times$ &
60.3 & 15.0 & 82.0 & 40.8 & 44.6 & 71.6 & 45.5 & 51.4 \\
\specrow \hspace{2mm}$\hookrightarrow$~+~SPEC-RL & 25 & 53 & 157.1 & 2.33$\times$ &
61.1 & 16.0 & 82.8 & 42.3 & 45.3 & 70.0 & 48.4 & 52.3 \\

DAPO & 30 & 74 & 525.3 & 1.00$\times$ &
60.0 & 16.0 & 84.0 & 42.6 & 44.9 & 72.9 & 48.1 & 52.6 \\
\specrow \hspace{2mm}$\hookrightarrow$~+~SPEC-RL & 30 & 68 & 199.8 & 2.40$\times$ &
64.4 & 19.3 & 83.6 & 43.4 & 48.1 & 70.7 & 50.3 & 54.3 \\

DAPO & 35 & 89 & 641.1 & 1.00$\times$ &
62.7 & 16.4 & 84.6 & 41.9 & 45.2 & 72.9 & 49.9 & 53.4 \\
\specrow \hspace{2mm}$\hookrightarrow$~+~SPEC-RL & 35 & 83 & 242.8 & 2.46$\times$ &
65.6 & 19.9 & 85.2 & 42.6 & 48.4 & 72.4 & 51.4 & 55.1 \\

\bottomrule
\end{tabular}
}
\caption{Intermediate training results of \textbf{Qwen-3-14B-Base} on DeepMath-6K with \textbf{DAPO} and \textbf{SPEC-RL}.
Since DAPO adopts \textit{Dynamic Sampling}, one training step may correspond to multiple generation steps; thus we additionally report the \textbf{Gen-Step} column to indicate how many rollout batches the model has consumed.}
\label{tab:dapo-14b-details}
\end{table*}

\clearpage

\end{document}